%% file: paper.tex
\documentclass[]{article}

\usepackage[]{algorithm2e}
\usepackage{amsmath}
\usepackage{amssymb}
\usepackage{accents}
\usepackage{authblk}
\usepackage{booktabs}
\usepackage{caption}
\usepackage{color}
\usepackage{float}
\usepackage[margin=1.5in]{geometry}
\usepackage{graphicx}
\usepackage[utf8]{inputenc}
\usepackage{makecell}
\usepackage{siunitx}
\usepackage{subfig}
\usepackage[hidelinks]{hyperref}

\DeclareMathOperator*{\argmin}{arg\,min}

\newcommand{\sgn}{\operatorname{sgn}}
\DontPrintSemicolon

\newlength{\dhatheight}
\newcommand{\doublehat}[1]{%
    \settoheight{\dhatheight}{\ensuremath{\hat{#1}}}%
    \addtolength{\dhatheight}{-0.35ex}%
    \hat{\vphantom{\rule{1pt}{\dhatheight}}%
    \smash{\hat{#1}}}}

\newcommand*\mycaption[2]{\caption[#1]{\textbf{#1} #2}}

\newcommand*{\ea}{{et al.}\@\xspace}
\newcommand{\ie}[1][ ]{{i.\thinspace e\@.}#1}
\newcommand{\eg}[1][ ]{{e.\thinspace g\@.}#1}
\newcommand{\comment}[2][blue]{}

\newcommand{\refsec}[1]{Sec.~\ref{sec:#1}}
\newcommand{\reffig}[1]{Fig.~\ref{fig:#1}}
\newcommand{\refeq}[1]{Eq.~\ref{eq:#1}}
\newcommand{\reftab}[1]{Table~\ref{tab:#1}}

\title{Hierarchical Policy Design for Sample-Efficient Learning of Robot Table Tennis Through Self-Play}

\author[1,2]{Reza~Mahjourian}
\author[1]{Risto~Miikkulainen}
\author[2]{Nevena~Lazic}
\author[2]{Sergey~Levine}
\author[2]{Navdeep~Jaitly}
\affil[1]{University of Texas at Austin}
\affil[2]{Google Brain}

\date{}

\begin{document}

\maketitle

\begin{abstract}
\input{abstractbody}
\end{abstract}


\input{paper-body}

\section*{Acknowledgments}

Special thanks to Erwin Coumans for help with PyBullet and VR integration, Torsten Kröger and Kurt Konolige for help and discussions on the Reflexxes library, and the Google Brain team for discussions and support.

\bibliographystyle{plain}  
\bibliography{paper}        

\end{document}

%% file: abstractbody.tex

Training robots with physical bodies requires developing new methods and action representations that allow the learning agents to explore the space of policies efficiently.  This work studies sample-efficient learning of complex policies in the context of robot table tennis.  It incorporates learning into a hierarchical control framework using a model-free strategy layer (which requires complex reasoning about opponents that is difficult to do in a model-based way), model-based prediction of external objects (which are difficult to control directly with analytic control methods, but governed by learnable and relatively simple laws of physics), and analytic controllers for the robot itself.  Human demonstrations are used to train dynamics models, which together with the analytic controller allow any robot that is physically capable to play table tennis without training episodes.  Using only about 7,000 demonstrated trajectories, a striking policy can hit ball targets with about 20\,cm error.  Self-play is used to train cooperative and adversarial strategies on top of model-based striking skills trained from human demonstrations.  After only about 24,000 strikes in self-play the agent learns to best exploit the human dynamics models for longer cooperative games.  Further experiments demonstrate that more flexible variants of the policy can discover new strikes not demonstrated by humans and achieve higher performance at the expense of lower sample-efficiency.  Experiments are carried out in a virtual reality environment using sensory observations that are obtainable in the real world.  The high sample-efficiency demonstrated in the evaluations show that the proposed method is suitable for learning directly on physical robots without transfer of models or policies from simulation.\footnote{Supplementary material available at \url{https://sites.google.com/view/robottabletennis}}

%% file: paper-body.tex

\section{Introduction}
\label{sec:intro}


%

From ancient mythologies depicting \emph{artificial people} to the modern science fiction writings of Karel Čapek and Isaac Asimov, there seems to be a clear image of what robots ought to be able to do.  They are expected to operate in the world like human beings, to understand the world as humans do, and to be able to act in it with comparable dexterity and agility.

Just as today most households can have personal computers in the form of desktops, tablets, and phones, one can imagine a future where households can use the assistance of humanoid robots.  Rather than being pre-programmed to do specific jobs like communicating with people, helping with kitchen work, or taking care of pets, these robots would be able to learn new skills by observing and interacting with humans.  They can collectively share what they learn in different environments and use each other's knowledge to best approach a new task.  They already know their bodies well and are aware of their physical abilities.  They are also aware of how the world and the common objects in it work.  They just need to learn how to adapt to a new environment and a new task.  If they need to learn a new skill by trying it out, they can do so efficiently.  They can learn a lot from a few attempts and use reasoning and generalization to infer the best approach to complete the task without having to try it for thousands of times.

This article takes a step in that direction by building a robotic table-tennis agent that learns the dynamics of the game by observing human players, and learns to improve over the strategy demonstrated by humans using very few training episodes where the agent plays against itself in a self-play setup.



\subsection{Motivation}
\label{intro:motivation}

The rate of progress in creation of intelligent robots seems to have been slower than other areas of artificial intelligence, like machine learning.  That is because intelligent robotics requires not only human-like cognition, but also human-like movement and manipulation in the world.  As of now, the most successful applications of robotics remain in the industrial domains, where the focus is on precision and repeatability.  In those environment, the expected robot motion is known beforehand and there is no need to deviate from it.  However, the general usability of robots depends on their ability to execute complex actions that require making multiple decisions through time.

Deep learning and reinforcement learning have been successful in solving interesting problems like object detection, playing Atari games, and playing board games like chess and Go.  These advances have made it possible to approach human-level perception and cognition abilities.  While perception problems can be learned in data centers using millions of data samples and training episodes, learning general robotic skills requires interacting with physical robot bodies and environments, which cannot be parallelized.  Therefore, learning robotic agents need to be very efficient in how they use training samples.


This article explores sample-efficient learning of complex robotic skills in the context of table tennis.  Playing robot table-tennis games is a challenging task, as it requires understanding the physics of the robot and the game objects, planning to make contact with the ball, and reasoning about the opponent's behavior.

\begin{figure}[htb!]
  \centering
  \includegraphics[width=0.6\columnwidth]{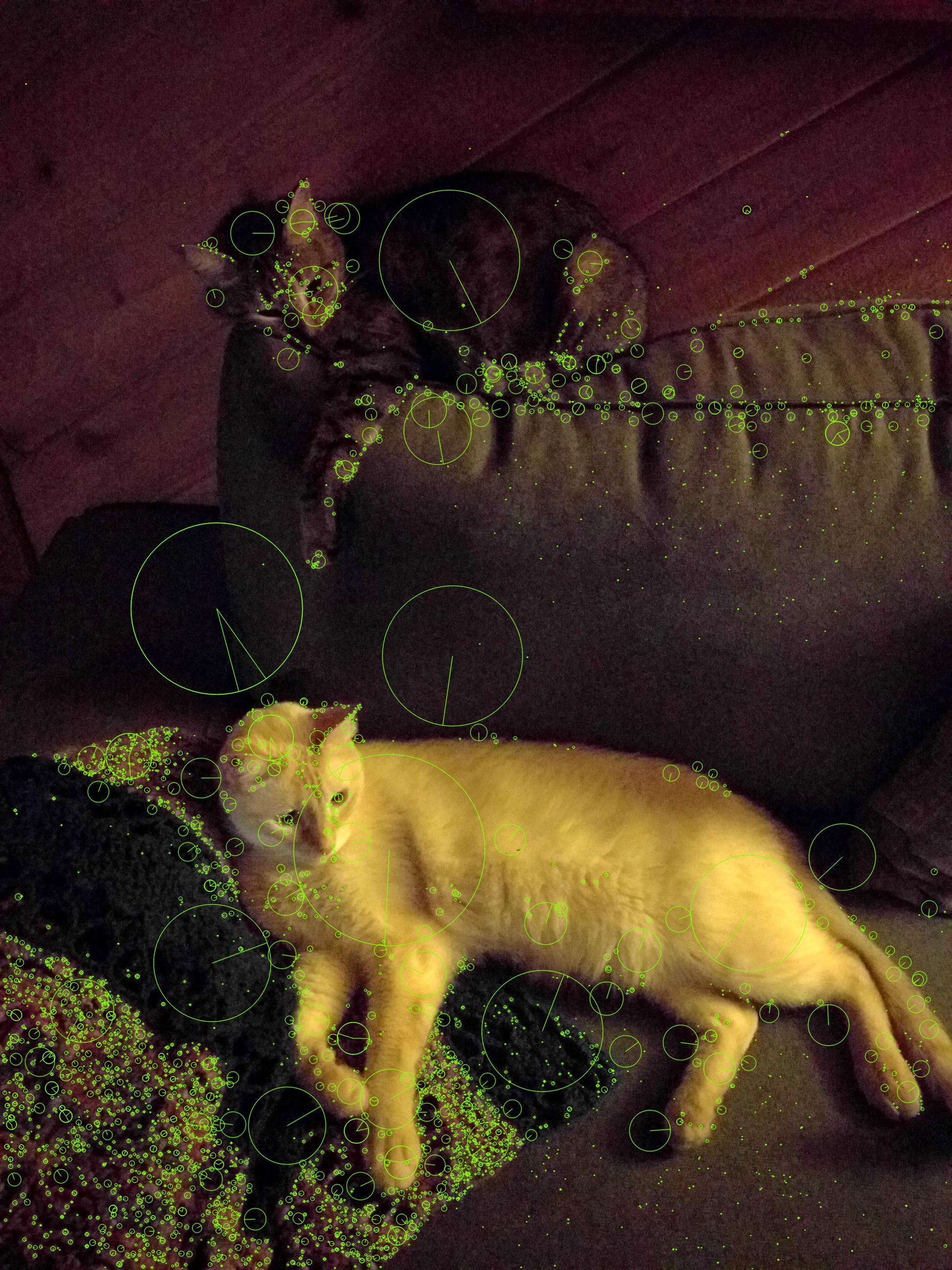}
  \mycaption{Example SIFT Keypoints Detected in an Image.}{SIFT keypoints can be used for object classification.  The keypoints are extracted from images and individually compared to a database of existing keypoints extracted from other images.  A matching algorithm can find candidate matching features based on the distance between feature vectors.  Application of deep learning to object classification has resulted in discovering convolutional feature maps that are more effective than SIFT features.}
  \label{fig:intro:sift}
\end{figure}

\begin{figure}[htb!]
  \centering
  \includegraphics[width=0.9\columnwidth]{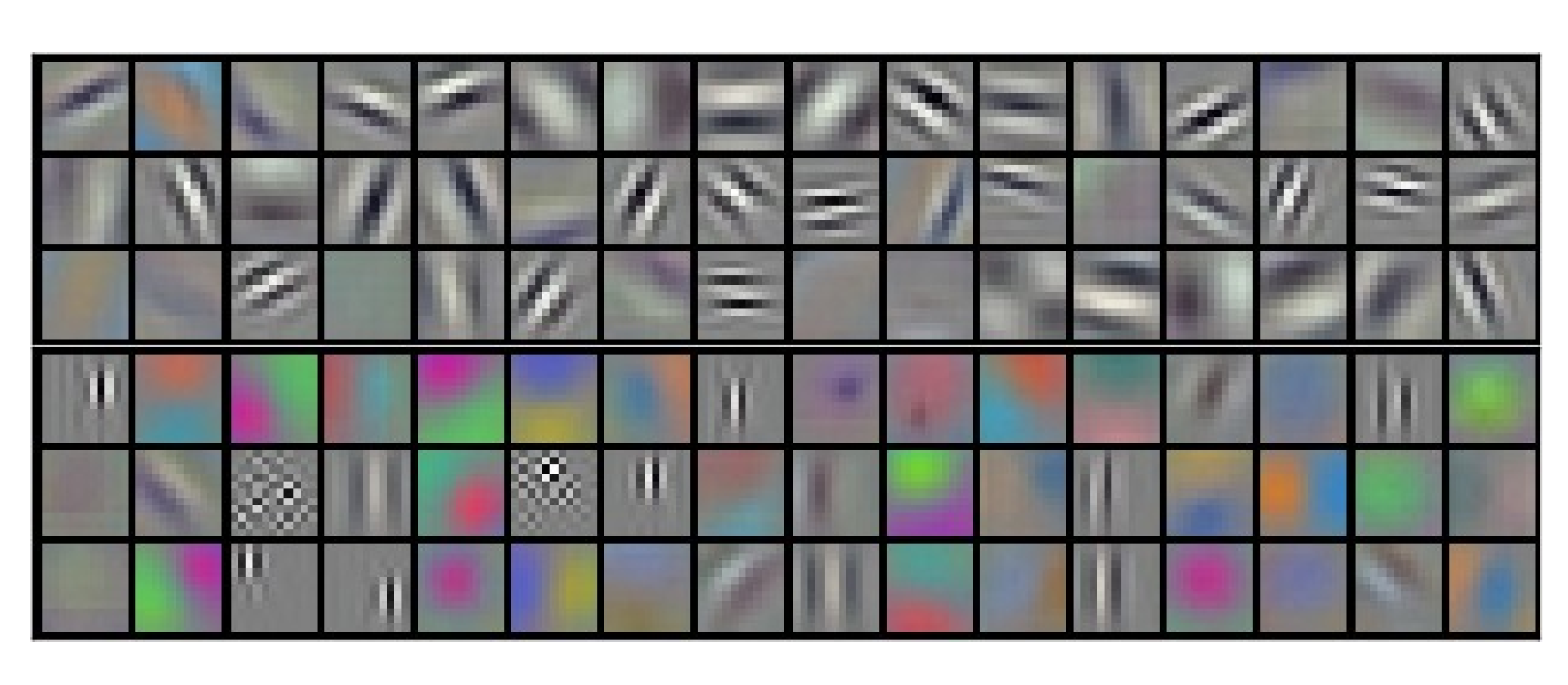}
  \mycaption{Convolutional Filters Learned in AlexNet~\cite{krizhevsky2012imagenet} for Image Classification.}{The image shows 96 convolutional kernels of size 11x11x3 learned by the first convolutional layer in AlexNet.  Deep learning is able to discover suitable features for the task of image classification.  These learned features perform better than the engineered SIFT features.  This example highlights the potential of learning algorithms to discover novel and effective solutions without a need for engineering.}
  \label{fig:intro:filters}
\end{figure}

There have been many examples where application of deep learning to a problem has resulted in developing a superior approach with improved performance.  For example, object classification and object detection tasks used to rely mainly on engineered SIFT features~\cite{lowe1999object}, an example of which is shown in \reffig{intro:sift}.  However, AlexNet~\cite{krizhevsky2012imagenet} demonstrated end-to-end learning of object classification on ImageNet~\cite{deng2009imagenet} using convolutional neural networks.  \reffig{intro:filters} visualizes the convolutional filters learned in AlexNet in the first layer of the neural network.  These filters can be regarded as the learned equivalents to SIFT image features.  In the object classification domain, using neural networks to solve the task end-to-end allowed it to discover a suitable representation for image features that outperformed engineered features.

As another example, for the tasks of speech recognition and language translation, end-to-end learning has replaced the pipelines based on human-designed acoustic models, language models, and vocabularies with neural networks that outperform the old approaches.  In the classic pipelines, the vocabularies shared between the nodes were engineered and fixed.  The components in the pipeline were restricted in choosing their outputs from the hand-designed vocabularies.  In the end-to-end approach, the network is free to learn and use an internal embedding for the speech data and the language data.  This added freedom allowed deep learning to discover intermediate representations and features that are more suitable for solving the task.

Similarly, Mnih \ea~\cite{mnih2013playing} applied deep reinforcement learning to playing Atari games and demonstrated the ability of deep learning to discover a value network that can map raw pixels in the game to an expectation of future rewards.

These successes suggest that there is similar opportunity for applying deep learning to discover novel intelligent robotic behaviors.  In the domain of table tennis, there is the potential for learning to discover:

\begin{enumerate}
\item \textbf{Better Strikes}: Can the robot swing the paddle in new ways beyond what humans have tried in table tennis?  In sports, one can observe leaps where a player tries a new technique and then very quickly it is adopted by other players.  For example, in the early nineties handball players started using spinshots that would hit the floor past the goalkeeper and turn to go inside the goal.  Can reinforcement learning discover new striking motions for hitting the table-tennis ball?
\item \textbf{Better Game Strategy}: There are established human strategies for playing adversarial table-tennis games.  Can reinforcement learning discover new overall game-play strategies that are more effective in defeating a human opponent?
\end{enumerate}

Discovering better motions and better strategies to solving tasks are relevant to household robots as well.  This article aims to utilize the ability of learning to discover such behaviors and demonstrate them in the domain of table tennis, and therefore show that such learning can be useful for general robotic tasks as well.


\subsection{Challenges}

General learning algorithms typically require millions of samples or training episodes to learn a task.  Collecting samples for learning robotic tasks is costly, since each sample can cause wear and tear on the robot.  The process is also time-consuming since interactions in the real world need to happen in real time and cannot be sped up by faster compute.  In addition, robotic environments are often fragile and one cannot depend on agents learning automatically in unsupervised environments.  Often, things break or objects get displaced requiring operator intervention to restore the setup to continue the training.

Moreover, there is usually an outer loop around the learning algorithms.  Applying reinforcement learning is typically a trial-and-error process.  The researchers usually develop new methods in an iterative manner by trying different approaches and hyperparameters.  For every new instance of the problem, the learning algorithm is typically run from scratch.  Therefore, in order to be feasible, advanced methods for learning general-purpose robotic tasks have to be able to use samples more efficiently than what is currently possible with deep learning and RL methods.

The end-to-end learning approach based on producing and consuming more and more data is not suitable for robotics.  It is possible to bring some scale to learning robotic tasks using parallel hardware setups like arm farms.  However, end-to-end learning methods often need so much data that this amount of parallelism is not enough to overcome the physical limitations that come with learning in the real world.  An increase in the number of hardware setups also increases the expected frequency of hardware failures, which increases the need for human supervision.

Learning end-to-end policies poses another challenge, which is identifying the source of bugs or inefficiencies in one component of the implementation.  In an end-to-end setup, the impact of a new change can only be observed by how it affects the overall performance of the system.  Often learning algorithms are able to mask bugs by continuing to operate at a slightly reduced capacity or precision, thereby making it difficult to trace the root source of a problem after a few stages of development.

Some applications of deep learning to robotics can avoid some of the physical limitations by focusing on the perception part of the problem and ignoring learning motor skills.  For example, object grasping can be approached as a regression problem, where the agent maps the input image to a grasp position and angle, which is then executed using a canned motion.  However, when learning robotic skills it is very desirable for the learning algorithms to also discover novel motor skills.  Learning algorithms may be able to discover new ways of handling objects that are more suitable for robots, and more effective with fewer degrees of freedom typically present in robot bodies.

A common approach to learning robotic tasks is sim2real: learning in simulation and then transferring the policy to work in the real world.  With this method, learning can be done in the simulator.  However, this approach requires solving a secondary problem, which is making the learned policy work with the real sensory observations and the control dynamics of the physical robot.  Depending on the task, this transfer might not be any easier than learning the main problem.

Achieving breakthroughs in robotic learning most likely depends on discovering new learning approaches and new intermediate state and action representations that allow learning algorithms to spend a limited experimentation budget more strategically.  Such intermediate state and action representations should be general enough to sufficiently capture the state of all policies.  Yet, at the same time, they should be high-level enough to allow the learning agent to efficiently explore the space of all policies without having to try every combination.  This article takes a step in that direction by presenting an approach that achieves high sample-efficiency in learning the complex game of robotic table-tennis.


\subsection{Approach}

The approach presented in this article offers a solution to the challenges discussed in the previous section by developing a learning solution that can discover \textbf{general} robotic behaviors for table tennis, yet is \textbf{sample-efficient} enough that it can be deployed in the \textbf{real world} without relying on transfer learning from simulators.

The approach incorporates learning into a hierarchical control framework for a robot playing table tennis by using a model-free strategy layer (which requires complex reasoning about opponents that is difficult to do in a model-based way), model-based prediction of external objects (which are difficult to control directly with analytic control methods, but governed by learnable and relatively simple laws of physics), and analytic controllers for the robot itself.

The approach can be summarized around eight design decisions reviewed below.  Using a \emph{virtual reality environment} and collecting \emph{human demonstrations} in this environment make the approach more \textbf{realistic}.  Also, working with \emph{low-dimensional state} instead of raw vision increases the chances that the sample-efficiency achieved in simulation would be reproducible in the real world. Introducing \emph{rich high-level action representations} based on landing targets for the ball and target motion-states for the paddle, and \emph{learning game-play strategies with self-play} enables the approach to discover \textbf{general} striking motions and versatile strategies for playing table-tennis.  The division of tasks in a \emph{hierarchical policy}, employing \emph{model-based learning} for the striking skills, training the models over \emph{low-dimensional state} and \emph{high-level action representations}, and developing an \emph{analytic robot controller} for executing high-level paddle-motion targets makes the method \textbf{sample-efficient}.  The following subsections provide an overview of the main components of the approach.

\subsubsection{Virtual Reality Learning Environment}

In order to establish whether the approach can handle the complexity of real-world sensors, the method is developed in a Virtual Reality (VR) environment which allows for capturing the same sensory observations that would be available in a real-world table-tennis environment.  Using VR instead of using plain simulation helps make sure the learning environment is realistic enough that the results would transfer to the real-world.

Although the method in this article can be combined with sim2real approaches by using the models and policies learned in simulation as a starting point for training real-world models and policies, the emphasis in this article is on developing an approach that would be sample-efficient enough to be able to learn the task from scratch in the real world.  So, the method is developed in a VR environment where the elements in the observation and action spaces have counterparts in the real world.  The next section describes how using low-dimensional state can increase similarities to real-world setups.  It also outlines how the low-dimensional state can be obtained from physical sensors in the real world.


\subsubsection{Using Low-Dimensional State}

To make the environment more realistic, and to increase sample-efficiency, the approach uses low-dimensional state instead of raw vision.  The observation and action spaces in the VR environment are chosen such that they have parallels in the real world.  More specifically, the state space of the learning agents is limited to the low-dimensional ball-motion state and paddle-motion state.   In the real world, ball-motion state can be obtained from ball-tracking algorithms, and paddle-motion state can be obtained with identical or similar sensors to what is used in the VR environment.  Similarly, the action space of the striking policies is defined by paddle-motion targets, which can be tracked and controlled precisely on physical robots.

\textbf{Ball-motion state} includes the position and velocity of the ball.  In the VR environment, ball-motion state is available from the underlying simulator.  In the real world, a ball tracker~\cite{seo1997ball, chen2012ball} can provide the position and velocity of the ball.  Ball trackers usually track the ball velocity as well, since estimates on the current velocity of the ball can speed up the detection algorithm by limiting the search to a small region in the image and improve its accuracy by ruling out false positives.  Detecting and tracking the location of a ball in a camera image is a relatively simple computer vision task.  Ball tracking can be done with classic computer vision algorithms and does not require learning.  The ball has a fixed geometry and a simple appearance in the camera images.  A blob detection algorithm can identify the ball in the image.  Given detections from two or more cameras and the camera intrinsics, the 3D location of the ball can be estimated.  An advantage of using classic vision algorithms over using deep neural networks is the higher computational speed, which is critical in a high-speed game like table tennis.

\begin{figure}[htb!]
  \centering
  \subfloat[A Vive tracker.]{{\includegraphics[width=0.45\columnwidth]{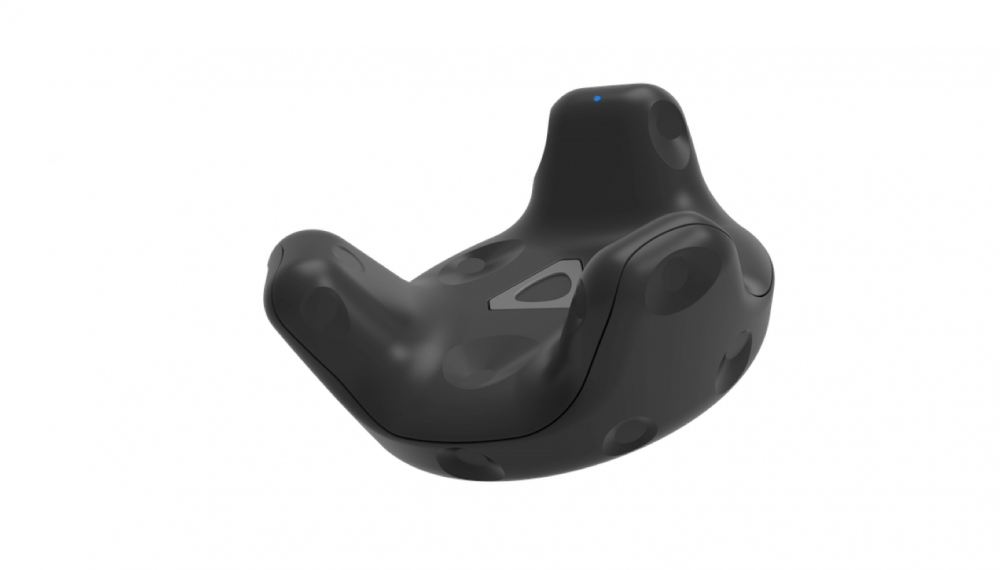} }}
  \qquad
  \subfloat[A Vive tracker attached to a paddle.]{{\includegraphics[width=0.45\columnwidth]{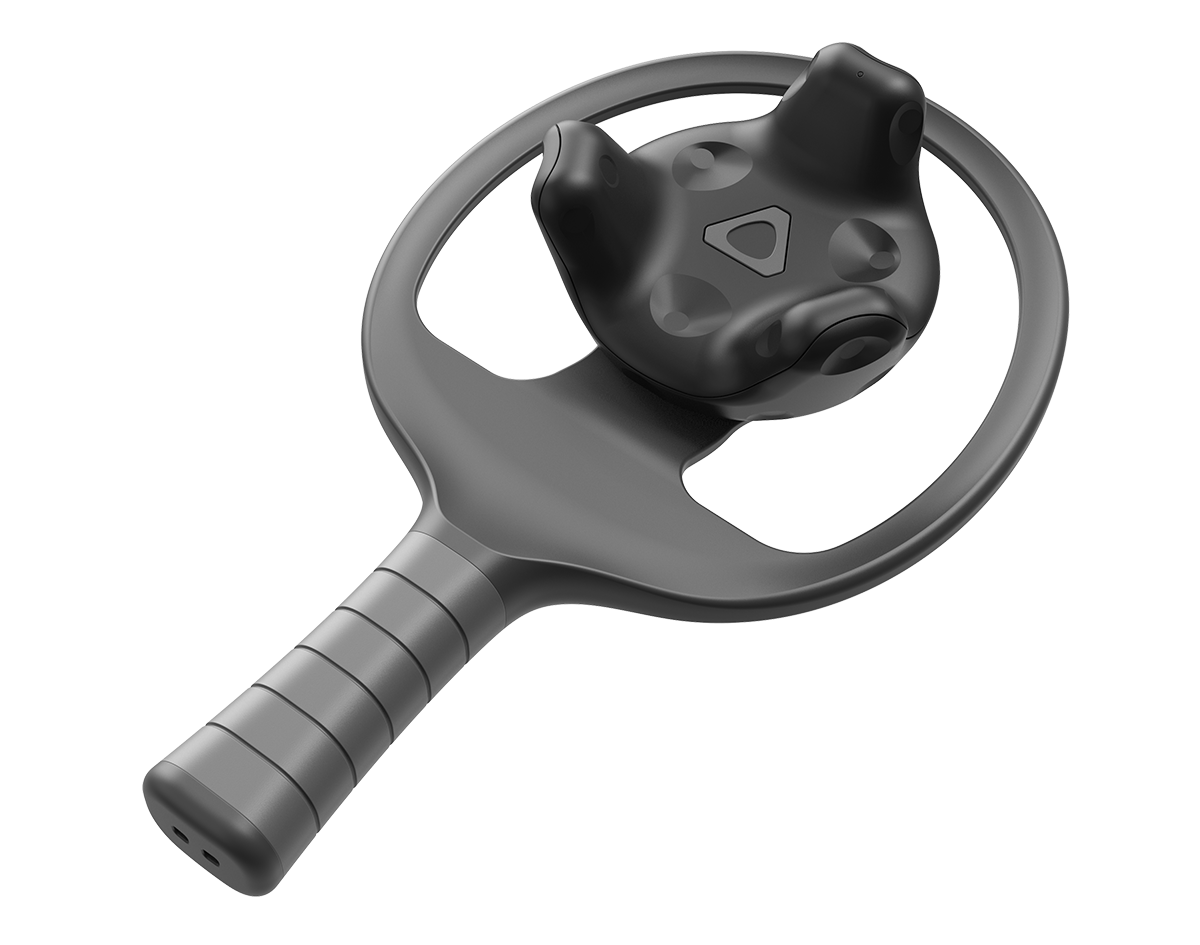} }}
  \mycaption{Virtual Reality Trackers.}{The trackers allow the position and orientation of objects to be tracked with sub-millimeter accuracy.  In the VR environment, these trackers make it possible to capture the paddle motions generated by human players.  The same trackers, or any other motion tracking technology, can be used to track the motion of table-tennis paddles in the real world. Photo credits: HTC.}
  \label{fig:tracker}
\end{figure}

\textbf{Paddle-motion state} includes the paddle's position, orientation, linear velocity, and angular velocity.  When the paddle is attached to the robot, paddle-motion state can be obtained using forward kinematics.  When learning from human games, paddle-motion state needs to be obtained from a motion-tracking system.  There are a variety of solutions that allow for tracking the paddle-motion state with high accuracy.  On the higher end, it is possible to use full-blown motion tracking systems to track marked and instrumented paddles.  On the lower end, one can use off-the-shelf tracking devices like HTC Vive, which can provide position information with sub-millimeter accuracy and jitter.  \reffig{tracker} shows two types of VR trackers that work with HTC Vive.  In fact, this is the same hardware that is used for experiments in this article when collecting human demonstrations in the VR environment.  Since such trackers are bulky, the human players would be able to use only one side of the instrumented paddles.  Lastly, a more custom tracking setup can use small IMU sensors attached to the paddles.  Visual markers on the paddles can be used to correct for the sensory drift that is common with IMUs.

Grounding the models and policies in low-dimensional state reduces the dimensionality of the learning problems and improves sample efficiency.  Moreover, employing a separate component for extracting the low-dimensional ball-motion state from visual inputs makes it possible to debug and fine-tune that component before integrating it into the implementation for the learning agent.  In contrast, using raw visual input would create a large disconnect between the distribution of sensory observations that are typically available in simulation, and raw visual data available from cameras in the real world, thereby limiting the extent to which the simulation experiments can predict real-world performance.  Another issue with using raw vision is that working with vision in the real world requires carefully designing the training environment to capture different lighting conditions, backgrounds, etc.  Any mismatches between the training and test environments would greatly disrupt the performance of policies trained with raw vision.  The next section describes how dynamics models trained on low-dimensional observations can inform the agents to make better decisions.

\subsubsection{Model-Based Learning}

Learning policies directly by interacting with the environment may require too many training samples.  Model-free RL agents often need to implicitly learn to predict the outcome of their actions by predicting how their actions changes the environment.  The approach in this article uses model-based learning to increase sample-efficiency.

The game of table tennis, despite being a complex and fast game requiring great skill to play, has relatively simple physics compared to other tasks like robot locomotion or object grasping.  In table tennis, most of the time, the ball is travelling in the air where it is only subject to gravity, drag, and Magnus forces due to its spin.  The ball experiences short contacts with two types of objects: the table, and the player paddles.  If the dynamics of the ball's motion and contact are understood, it is possible to both predict the ball's future states and to control for it by picking the right paddle motion to execute the desired contact.

The method uses observations in the environment to train dynamics models that predict the future state of the ball due to its free motion and due to contact with the player paddle.  Such dynamics models can inform the learning agents about the consequences of the actions they are exploring.   In contrast, in end-to-end model-free learning approaches, the agents are required to implicitly learn how the environment works in order to best exploit it and increase their reward.  By capturing the simple physics of table tennis in dynamics models the method allows the learning agents to focus on learning high-level behaviors, thereby improving sample-efficiency.  The next section describes how these dynamics models are trained.

\subsubsection{Learning from Demonstrations}

Training the dynamics models requires data.  However, if there is no policy to drive the robot to play table-tennis, there is no way to collect the required data.  On the other hand, the observations that are needed to learn the ball motion and contact dynamics are readily available from human games.  There is no need to use a robot to collect samples for training the dynamics models.  Similarly, there is no need for kinesthetic teaching.  Moreover, capturing human demonstrations is a lot easier than operating fragile robot setups to collect data.  So, the approach trains dynamics models from human demonstrations.

The behavior of the ball and the contact forces between the ball and the table or the paddle are the same whether the paddle is carried by a robot or a human player.  Contrast this with a task like locomotion.  As the agent learns new gaits, it starts experiencing new joint states and new contacts with the ground, requiring any contact models to be adjusted.  In table tennis, one can study the ball's free motion and contact behavior just by observing human games in instrumented environments.  While collecting robot samples is costly and time-consuming, human samples can be obtained easily and abundantly.

Intermediate players may come short in their ability to move quickly, or to control the paddle correctly to execute their desired shot, which would pose a problem if policies are trained directly from human actions.  Such policies would be able to play table-tennis only as well as the humans providing the demonstrations.  So, this method only trains dynamics models from the human demonstrations and allows the policies to choose more versatile actions beyond what is demonstrated.  The dynamics models are independent of the learning agent and stay valid as the learner's policy changes.  They dynamics models can be used predict the future states of the ball when it is moving freely, and when it is hit by the paddle.  The models can also help predict how to hit a ball so that it lands at a desired target on the opponent's side of the table.  In other words, they can be used to choose actions.  The next section describes how a rich high-level action representation gives the learning agents the ability to make very high-level yet general decisions for how to strike the ball with the paddle.

\subsubsection{Rich General High-Level Action Representations}
\label{sec:intro:actions}

\begin{figure}[htb!]
\centering
\includegraphics[width=0.6\columnwidth]{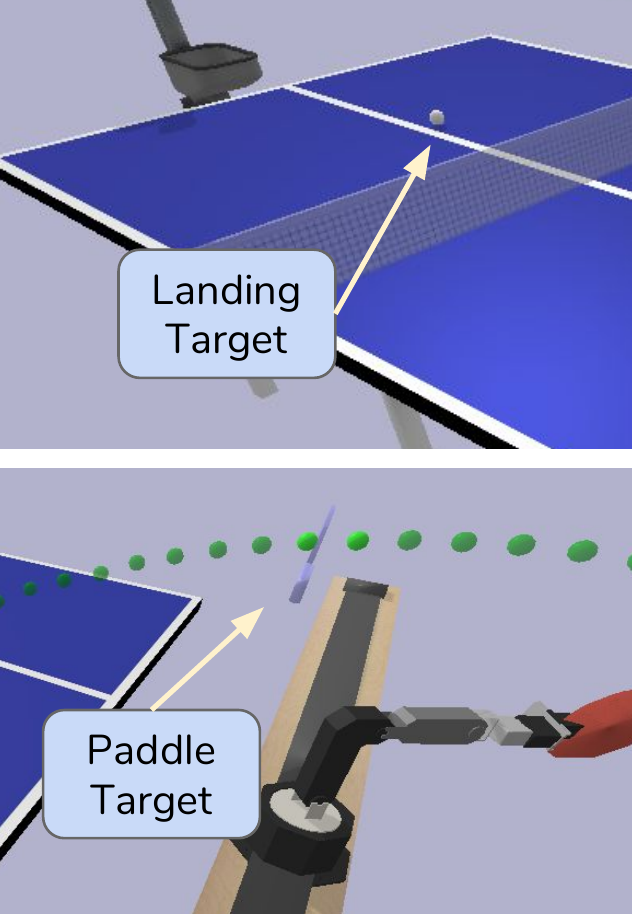}
\mycaption{Rich General High-Level Action Representations}{Top: A ball landing target specifies the desired position and speed for the ball as it lands on the opponent's side of the table.  Bottom: A paddle-motion target specifies the desired position, orientation, and velocity of the paddle at a desired time.  In the illustrated example, the paddle target is chosen to intersect with the predicted trajectory of the ball over multiple timesteps, which is shown by the string of green balls.  These high-level actions allow the agent to make high-level decisions about its game-play without losing generality in behavior.  Therefore, they increase sample-efficiency.}
\label{fig:intro:high-level-actions}
\end{figure}

Playing table tennis requires moving the paddle around all the time.  At any time during a rally, the agent is either trying to strike the ball towards the opponent, or trying to place itself optimally so that it can hit the next ball successfully.  Learning to play table tennis by continuously making decisions about how to move on every single timestep makes for a very difficult problem.  The reward for winning a rally may come only at the end of a long sequence of actions.  An RL agent would require too many samples to learn the basic techniques of striking the ball and moving around in the game just from the reward signal.  Therefore, the approach defines two rich high-level actions that allow the agent to make game-play decisions at a high-level without losing generality in behavior: ball landing targets, and paddle-motion targets.  The actions are illustrated in \reffig{intro:high-level-actions} and discussed below.

\begin{enumerate}
\item \textbf{Ball landing targets}: A landing target for the ball specifies a target motion-state for the ball at the moment that it lands.  In general, the target motion-state can encode the ball's position, velocity and spin.  In the current implementation, it encodes the position and speed (magnitude of velocity).  Such a high-level action allow the agent to specify a striking action by its desired outcome.  Learning to return the ball by specifying such targets is clearly easier than learning to return the ball by controlling the robot joints.  At the same time, such actions do not reduce from the generality of policies.   During each exchange between two players, each player's game-play can be represented by how the player lands the ball on the opponent's side of the table.  No matter what movements the player executes, the effective action from the player is how they hit the ball and how that ball behaves after contact.  In particular, the behavior of the ball at the moment of contact with the table can fully capture the ball's behavior, as the ball's motion-state at that moment fully specifies its expected motion after contact with the table.  So, although ball landing targets are high-level and easier to learn, they can capture and represent all possible striking behaviors (provided the implementation includes spin as well.)
\item \textbf{Paddle-motion targets}: A paddle-motion target specifies the target motion-state for the paddle at the moment that it hits the ball.  Paddle-motion targets are an alternative action representation for parameterizing a strike.  It is easier to learn to strike the ball by deciding a one-time target for the paddle than by deciding targets for robot joints over multiple timesteps.  At the same time, paddle-motion targets are general action as well.  During each strike, the impact of a player's paddle on the ball depends only on the motion-state of the paddle during the short period of time when the paddle and the ball are in contact.  All the actions taken by the players up to the moment of contact are just in service to achieving a paddle-motion state at the moment of contact with the ball.  So, representing a strike by the paddle-motion target that it achieves at the moment of contact does not reduce from generality of behavior.  Paddle-motion targets can also be used to position the robot when it is waiting for the opponent to act.  In those situations, the pose of the paddle is used as a proxy to control the position of the robot.  Since the paddle is the main vehicle for the robot to play the game, this action representation is very suitable for deciding positioning targets for the robot during the waiting intervals in a rally.
\end{enumerate}

Playing table tennis requires returning the ball to the opponent's side during each exchange.  However, there are so many choices in returning the ball.  The action representations used in this approach encode different striking behaviors, which permit different game-play strategies.  Moreover, using such abstract action representations simplifies the action space for the agents.  Since the actions capture the agent's behavior over multiple timesteps, they facilitates learning by eliminating the reward delay problem where a learning agent needs to figure out the actual action that leads to receiving a reward multiple timesteps into the future.  Therefore the rich high-level actions increase sample-efficiency while maintaining generality in behavior.  The following section describes how paddle-motion targets can actually be executed on the robot.

\subsubsection{Analytic Paddle-Control}

The dynamics models can inform the agent how to strike the ball, and the rich action representations allow the parameters of the strike to be specified at a high conceptual level.  However, there needs to be a controller that can actually execute such high-level actions like paddle-motion targets.  The method uses an analytic robot controller that is able to execute paddle-motion targets using information about the kinematics of the robot.

Executing paddle-motion targets requires precise control of the robot so that the paddle reaches the target position at the desired time, and that it has the desired velocity when it is crossing the target position.  In addition, reaching the target requires executing a long sequence of joint-control commands that span over multiple timesteps.  Learning to control the paddle by controlling the robot joints directly is a difficult task, as it requires learning implicitly how the robot commands affect the motion of the end-effector.  So, using RL to learn to control the execute paddle-motion targets may require too many samples.

This method develops an analytic paddle controller which uses the Reflexxes trajectory-planning algorithm to execute any paddle-motion target from any starting state for the robot, provided the target is achievable under the robot's physical limits and motion constraints.  The Reflexxes library is able to compute optimal trajectories to reach the target motion-state while satisfying velocity, acceleration, and jerk limits for each robot joint.  Employing this analytic controller removes the need for learning a paddle-control skill and improves the sample-efficiency of the method.  The next section describes how the analytic controller together with other skill policies are used in a hierarchical policy to play the whole game of table tennis.

\subsubsection{Hierarchical Policy}

Playing table tennis requires technical skills in moving the paddle and striking the ball, and tactical skills in choosing appropriate targets at different points of a rally.  This complexity is challenging for general-purpose RL algorithms.  Therefore, instead of approaching table tennis as a monolithic task, this approach uses a hierarchical policy that decomposes table tennis into a hierarchy of subtasks.  The hierarchical policy decouples the high-level skills from low-level skills in the game of table tennis, which makes it possible to implement each skill using a different method that is more suitable for it.  Moreover, the hierarchy allows each skill to be developed, evaluated, and debugged in isolation.  If necessary, the skills can be given perfect observations and perfect actions to fully evaluate their individual limits and errors.  Such a setup allows for identifying and addressing inefficiencies in each component of the system before they are integrated and fine-tuned together as a whole.

In the task hierarchy, low-level skills like how to move the paddle to a target position are implemented using analytic controllers that do not require learning.  Mid-level striking skills are implemented using dynamics models that are trained from human demonstrations with supervised learning.  Lastly, the top-level strategy skill is trained with reinforcement learning, allowing the agent to discover novel behaviors.

In contrast, learning a task end-to-end may cause the model to relearn the primitive skills over and over in various states in presence of changing inputs.  In other words, an end-to-end approach needs to learn to properly generalize its behavior to invariant states.  Doing so requires more training episodes.

As explained in \refsec{intro:actions} the action spaces used in the task hierarchy are such that they do not reduce from the generality of the policies.  In other words, the hierarchical policy does not restrict the agent's ability to explore the space of all possible game-play strategies and techniques.  As will be explained in \refsec{strategy}, model-based policies employing human data can be more sample-efficient, while model-free policies that directly pick paddle-motion targets as actions can exhibit more novel striking motions at the expense of lower sample-efficiency.

The hierarchical policy permits learning general policies in a sample-efficient manner.  The next section describes the hierarchical policy can discover interesting high-level game-play strategies.

\subsubsection{Learning Strategy with Self-Play}

The hierarchical policy design permits efficient training of general and parameterized low-level and mid-level skills which can execute different targets.  However, driving these skills requires a game-play strategy.  It is not possible to solve the game strategy analytically, as there are many choices in how to play table tennis and an effective strategy needs to factor in the behavior of the opponent as well.  So, the approach trains the strategy skill at the top of the hierarchical policy using a model-free RL algorithm, that is free to explore the space of all possible game-play strategies with no requirements other than maximizing the reward.

Since at the beginning there are no opponents to play against, the approach uses self-play to train the agent against itself.  As more self-play games are played, the strategy policy learns to adapt and respond to its own behavior.  The strategy policy picks stochastic actions that set goals for the mid-level skills in the hierarchy.  The strategy policy is encouraged to explore using an entropy term that rewards policies with more randomness.

The strategy skill allows the agent to make high-level decisions about its game plan without being concerned about how they are executed.  The strategy skill is the only skill in the task hierarchy that requires exploration and uses reinforcement learning to train.  By focusing the learning and exploration on this skill only, the method allows the agent to discover interesting general game-play strategies.

\subsection{Guide to the Reader}

The remainder of this article is organized as follows.  \refsec{sim} describes the simulation and virtual reality environment.  \refsec{method} provides a more in-depth overview of the method than what is given in this section.  \refsec{skill} explains the hierarchical policy design and the subtasks in the task hierarchy.  \refsec{env} describes how the learning environment is partitioned into a game space and a robot space so that the individual skills in the hierarchy can be trained with higher sample-efficiency.  \refsec{dyn} explains the dynamics models that allows the agents to predict the future states of the game, and evaluates the predictive ability of the trained dynamics models.  \refsec{paddle} describes the analytic paddle controller that is responsible for executing high-level paddle-motion action, and describes the implementation of the positioning policy.  \refsec{striking} describes the implementation of the different model-based striking policies and evaluates them against baseline model-free implementations that learn the striking skill from scratch using RL.  \refsec{strategy} uses self-play to train table-tennis game-play strategies in cooperative and adversarial games.  \refsec{disc} provides a discussion on the work presented in this article and outlines steps for future work, including how the method can handle vision and observation noise with continuous closed-loop control.  \refsec{related-work} discusses related work on robotic table-tennis and hierarchical RL and provides a short review of the underlying learning method used in this work.  Finally, \refsec{conclusion} lists the contributions made in this work and concludes the article.


\section{Simulation and Virtual Reality Environments}
\label{sec:sim}

This section describes the simulation and Virtual Reality (VR) environment that is used for data collection, training, and evaluation of the table-tennis agent.  First, the simulator and the virtual reality environment are introduced.  Then, the reinforcement learning environment and its state and action spaces are described.


\subsection{The Simulator}\label{sec:simsim}

\begin{figure}[htb!]
\centering
\includegraphics[width=0.9\columnwidth]{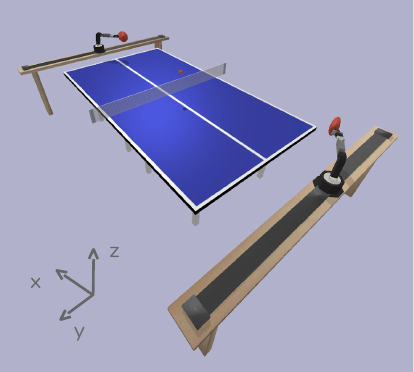}
\mycaption{Simulation Environment.}{Two WidowX arms are mounted on linear actuators that allow the arms to move sideways.  The two robot assemblies are at the same height as the table.  The robot assembly consists of a linear actuator and a robot arm.  The arm shown in the image is a WidowX arm with five joints.  The original arm has a gripper, which has been removed in this setup and replaced with a fixed link holding a paddle.}
\label{fig:sim}
\end{figure}

\reffig{sim} illustrates the simulator's setup.  The arm is mounted on a linear actuator, which allows the robot to move sideways.  This configuration has a wider reach compared to a stationary robot.  The linear actuator is implemented by one prismatic joint.  The arm and the linear actuator are treated as one robot assembly with six joints.  Fusing the linear actuator and the arm together in a single assembly simplifies inverse and forward kinematics calculations.

A different version of the simulation environment contains one robot playing against a table-tennis ball launcher.  The ball launcher can shoot table-tennis balls with controlled initial conditions (position and velocity).  By varying the initial conditions of every episode, the ball launcher makes it possible to explore the space of game conditions for the learning agents.  This version of the environment is also used in evaluations.

The simulation environment is implemented on top of the PyBullet~\cite{coumans2010bullet} physics engine.  Simulation objects are defined by their geometries and their physics parameters including mass, coefficient of restitution (bounciness), friction coefficients, etc.  The physics simulation in PyBullet is deterministic.  So, there is no inherent noise in the simulation.

The physics are simulated at \SI{1}{\kilo\hertz}.  At each physics timestep, the object states and forces are recomputed and any collisions are recorded.  Simulating physics at a high frequency increases the fidelity of the simulation and avoids glitches like missed collisions due to the fast motion of the table-tennis ball.


\subsection{Virtual Reality Setup}

The simulator described in the \refsec{simsim} is connected to a virtual reality setup, allowing a human player to control a free-moving paddle.  Using the VR setup makes it possible to create an immersive game environment where human demonstrations can be captured.  The VR environment is a good proxy for capturing human demonstrations in the real world with instrumented paddles.  In fact, the same trackers that are used in the VR setup can be used to instrument real table-tennis paddles and track their motion.  This setup for capturing the human demonstration data makes it more likely that the methodology and the results would transfer to the real world.

\begin{figure}[htb!]
\centering
\includegraphics[width=0.9\columnwidth]{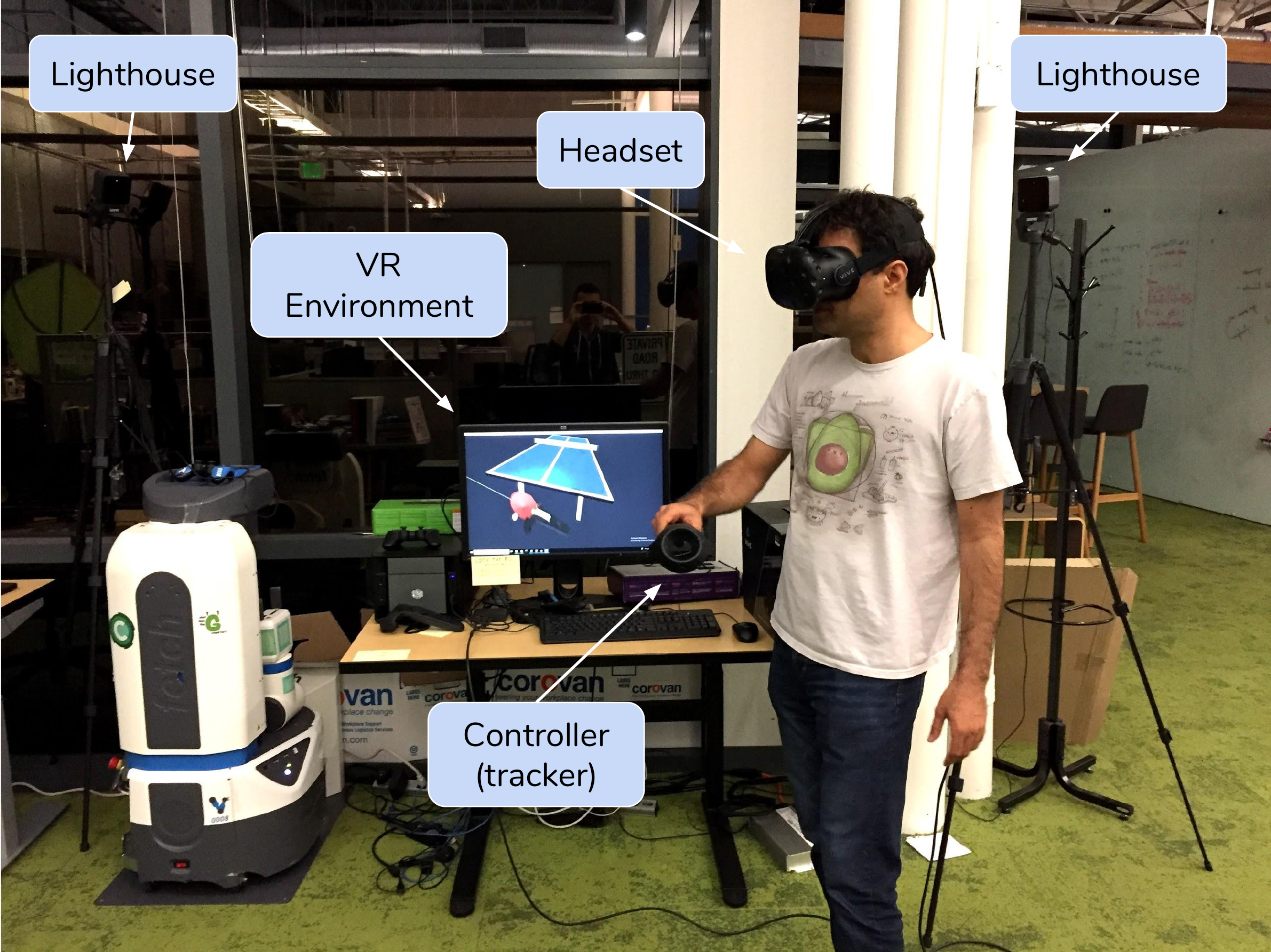}
\mycaption{Virtual Reality Setup.}{A person is using the VR environment to play table tennis against a ball launcher in simulation.  The VR hardware consists of two lighthouses, a headset, and a controller (tracker).  The simulator is connected to the VR setup, such that moving the VR controller in the real world moves a floating paddle in the simulator.  The paddle is used to hit the ball thrown by the launcher and return it to the other side of the table.  The VR environment permits capturing realistic paddle motions as demonstrated by humans.  The demonstrations are used to train dynamics models, which are then used by robotic agents playing table tennis against the ball launcher, or against each other.}
\label{fig:vr}
\end{figure}

The VR setup uses an HTC Vive headset, a controller (a.k.a. tracker), and two lighthouses.  The components are shown in \reffig{vr}.  The lighthouses continuously track the position and orientation of the player's headset and the controller in the player's hand.  The HTC VR hardware uses active lighthouses and passive headset and controllers.  The lighthouses emit vertical and horizontal sweeping laser lights at a fixed frequency.  The headset and the controller have an array of light-detecting sensors that fire whenever they receive the laser light.  Since the configuration of the sensors on the headset and controller are known, the timing of light-detection events reported by the different sensors contains enough information to decide the 3D position and orientation of each tracker.  As long as a tracker is exposed to one of the two lighthouses and a sufficient number of its light sensors are visible to it, the device can be tracked with the same accuracy.  So, if the paddle or the player hide the tracker from one of the lighthouses, it does not pose a problem.  \reffig{tracker} shows two types of VR trackers that work with HTC Vive.


\subsection{Learning Environment}

The learning environment is implemented using the OpenAI Gym~\cite{brockman2016openai} API.  The environment encapsulates the simulator and exposes the simulation object states as the environment state.  At every timestep $t$, the environment exposes the following information on the objects:

\begin{itemize}
\item The ball-motion state $b_t$, which includes its 3D position $l(b_t)$, and velocity vector $v(b_t)$;
\item The paddle-motion state $p_t$, which includes its 3D position $l(p_t)$, orientation $r(p_t)$, linear velocity $v(p_t)$, and angular velocity $\omega(p_t)$;
\item Robot joint positions $q_t$, and velocities $\dot{q}_t$;
\item Most recent collision and the ball's location and velocity at the moment of collision.
\end{itemize}

Each learning agent defines its own observation space, which is a subset of the environment state.  The action space of the environment includes the six robot joints.  The simulator supports position and velocity control modes for actuating the robot assembly.

There are three frequencies operating in the environment.  The simulator runs at \SI{1}{\kilo\hertz}, allowing for smooth simulation of physics and control of the robot.  The learning environment has a frequency of \SI{50}{\hertz}.  Every \SI{20}{\ms}, the environment state is updated based on the most recent physics state of the objects.  Collisions that are detected in between two environment timesteps are accumulated and reported together.  The collisions contain no time information, so they appear to have happened at the end of the environment timestep.  The high-level agents operate at a lower frequency in the environment.  They receive observations and choose actions only once during each ball exchange between the players.  Running at a lower frequency makes learning easier for the agents.  The high control frequency is appropriate for smooth control of the robot, but the agents do not need to make decisions at every simulation or environment timestep.  The lower frequency shortens the reward delay between the time the agent makes a decision and when it observes the consequence.

\subsection{Conclusion}

This section described the simulation and virtual reality environments that are used for simulating table-tennis games.  The next section provides an overview of the proposed method for learning to play table tennis with high sample efficiency.

\section{Method Overview}
\index{Method Overview@\emph{Method Overview}}%
\label{sec:method}

This section gives an overview of the approach and its key components.  It depicts a high-level picture of how the different components work together as part of the method.  \refsec{method:task}) discusses decomposing the task of playing table tennis into subtasks that can be learned or solved more efficiently.  \refsec{method:env} describes decomposition of the environment to separate the problem of robot control from the table-tennis game-play.  \refsec{method:dyn} discusses environment dynamics models that are trained from human demonstrations.  \refsec{method:analytic} describes an analytic robot controller which can execute target paddle-motion states (pose and velocity).  \refsec{method:strategy} discusses using self-play to learn high-level table-tennis strategies for cooperative and adversarial games.


\subsection{Policy Design}
\label{sec:method:task}

Robot table tennis is a complex task, and therefore it may be difficult for reinforcement learning.  The method decomposes the task into a hierarchy of skills where higher-level skills depend on lower-level skills.  The low-level skills are easy to learn; in turn, exposing the functionality of these low-level skills as primitives to higher-level skills makes those skills less complex and easy to learn as well.

\begin{figure}[H]
  \centering
  \includegraphics[width=0.85\columnwidth]{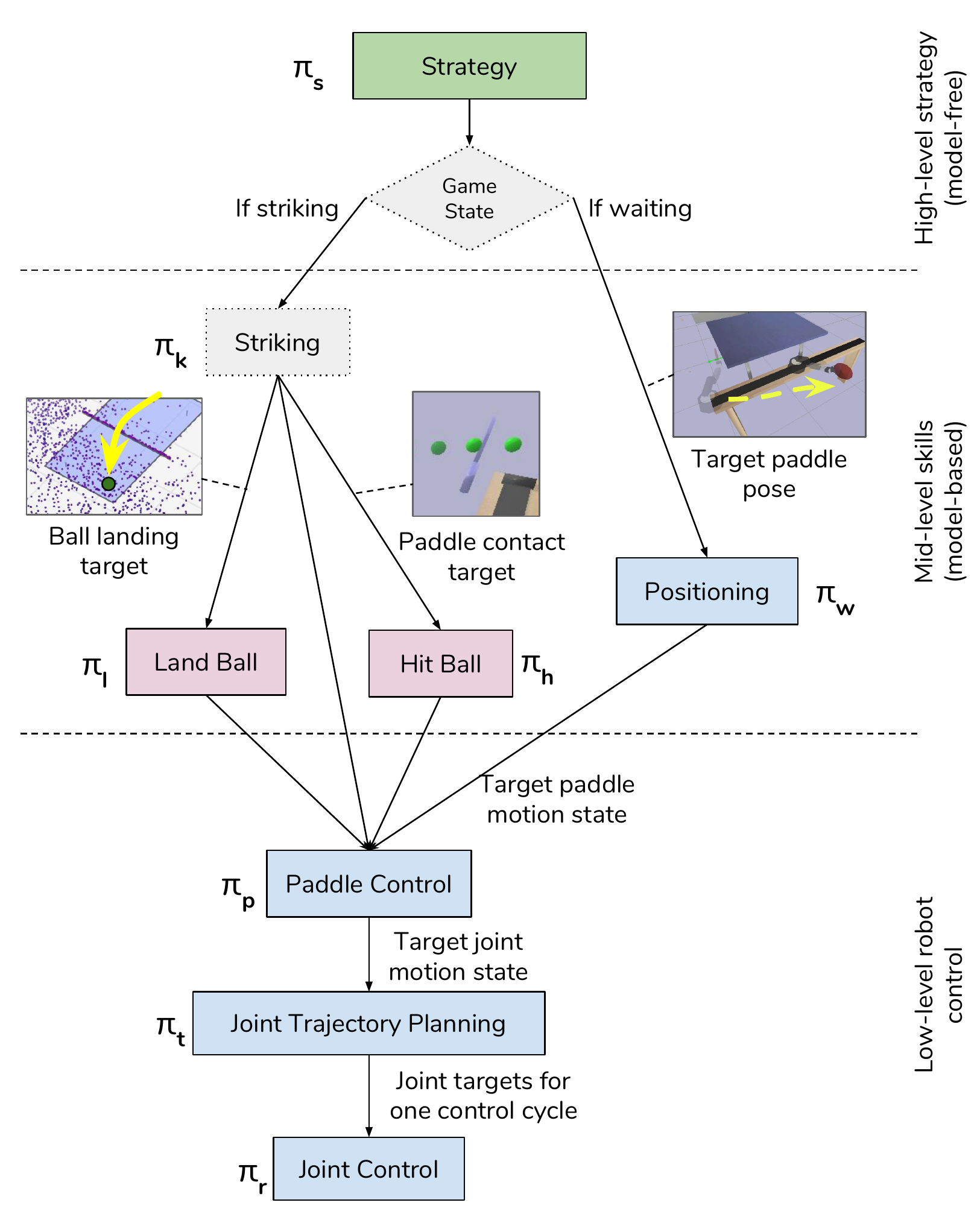}
  \mycaption{The Skill Hierarchy.}{The hierarchy consists of three levels of control (high-level, mid-level, and low-level), two modes (striking and waiting), and seven tasks.  At any point in the game, the agent is either in striking mode or waiting mode. When in striking mode, the agent strikes the ball using one of three different skills: land-ball, hit-ball, or directly with paddle-control.  Each variant of the policy uses only one of the three striking skills.}
  \label{fig:skill:hierarchy}
\end{figure}

The task hierarchy, which is illustrated in \reffig{skill:hierarchy}, offers a high-level view of the task to the learning agents.  Instead of continuously making decisions at every timestep, they make one high-level decision during each exchange with the opponent.  The high-level decisions determine how to strike the ball when the agent is returning a shot, and how to position the robot when the agent is waiting for the opponent.  \reffig{skill:hierarchy} shows three variants of the policy based on different striking skills.  The three striking skills are:

\begin{enumerate}
\item \textbf{Land-Ball}: Given the state of the incoming ball, hit it such that the ball lands at a desired location on the opponent's side with a desired speed.
\item \textbf{Hit-Ball}: Given the state of the incoming ball, hit it with a desired paddle orientation and velocity.
\item \textbf{Paddle-Control}:  Given the state of the incoming ball, hit it at a desired position, with a desired paddle orientation and velocity.
\end{enumerate}

The learning agents decide only targets for the striking skills.  Using the striking skills as the action space for the agents eliminates the reward delay problem, and consequently, the agents require fewer training episodes to learn.  At the same time, the striking skills does not reduce from the generality of policies.  Any sequence of low-level actions can be reduced to, or represented by the state of the paddle and the ball at the moment of contact.

\subsection{Environment Design}
\label{sec:method:env}

\begin{figure}[htb!]
  \centering
  \includegraphics[width=0.9\columnwidth]{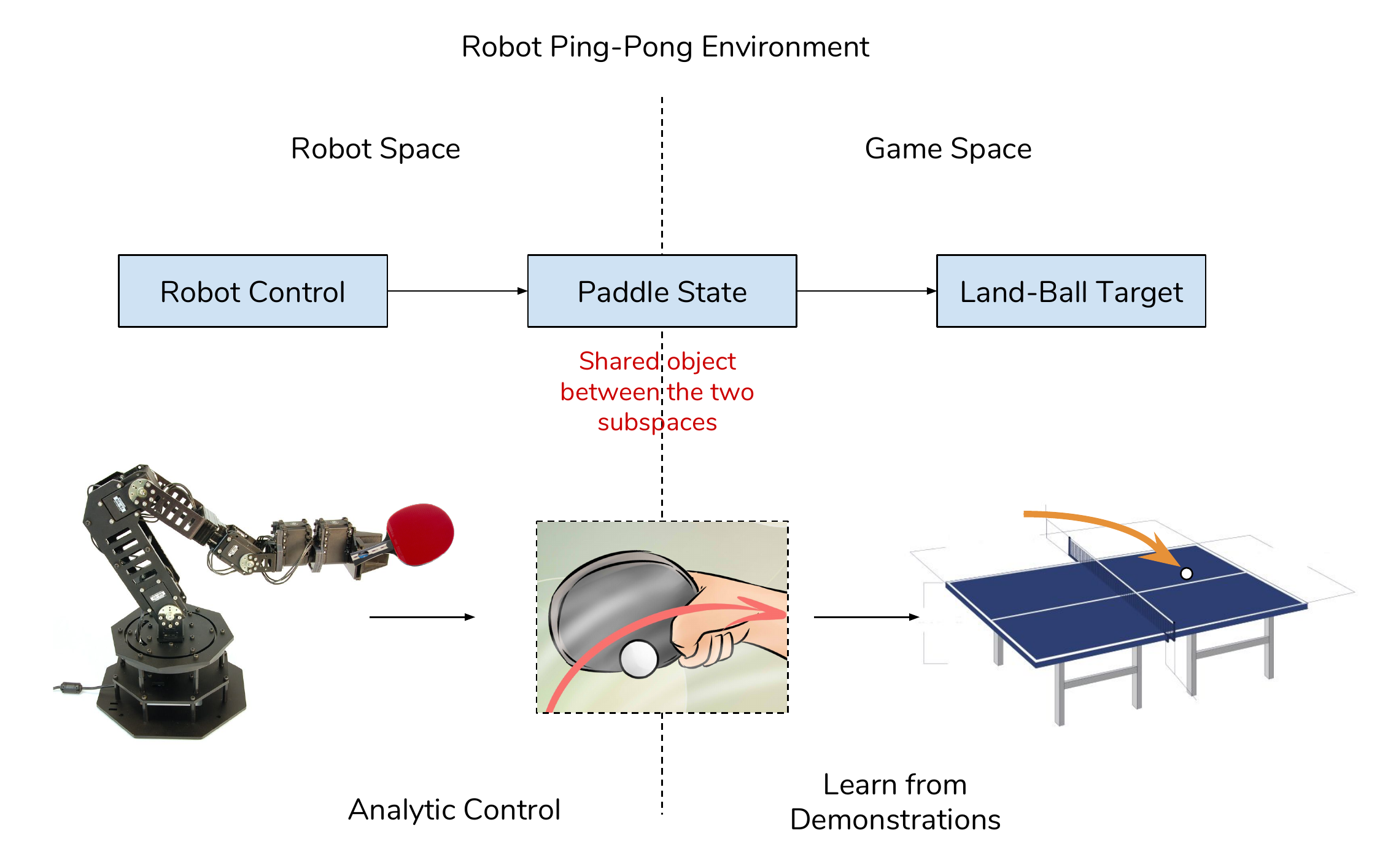}
  \mycaption{Decomposition of the Robot Table-Tennis Environment.}{The environment is divided into two spaces: the robot space and the game space.  The robot space deals with the physics and control of the robot.  The game space deals with the dynamics of table tennis.  The only shared object between the two spaces is the table-tennis paddle.  The decomposition makes it possible to learn the dynamics of table tennis from human demonstrations without using a robot.  On the other hand it simplifies the robot control problem to just accurately controlling the paddle.}
  \label{fig:env:decomposition}
\end{figure}

To make learning the skills in the hierarchical policy easier, the method decomposes the table-tennis environment into two spaces: the game space, and the robot space, as shown in \reffig{env:decomposition}.  The game space is concerned with the physics of the table-tennis game involving a table, ball and paddles, independently of how the paddles are actuated.  The robot space is concerned with the control of a robot that has a table-tennis paddle attached to the end of its arm.  The game space includes only the table, the ball, and the paddle; it deals only with the game of ping pong.  The robot spaces includes only the robot and the paddle; it deals only with the physics of the robot and end-effector control.  The only object shared between the two spaces is the table-tennis paddle.

This separation makes it possible to study and model the physics of table tennis without any robot controllers or agent policies.  In particular, it permits modelling the dynamics of the game just by observing humans playing table tennis with instrumented paddles.  On the other hand, isolating the robot space makes it possible to focus on the problem of paddle-control without any complications from the game of table tennis.

Moreover, decomposing the environment makes it easier to replace the robot with a different model, since there is no need to retrain the game models from scratch.


\subsection{Dynamics Models}
\label{sec:method:dyn}

Learning starts with observing human games with instrumented table-tennis paddles.  The human demonstrations are used to train dynamic models for the environment.  These models mainly capture the physics of ball motion and paddle-ball contact dynamics.  Once such dynamics models are trained, they can be used to predict the future trajectory of the ball.  They can also be used to make predictions about where a given ball will end up if it is hit with a given paddle pose and velocity.

In addition to predicting the outcome of actions, such dynamics models can be used for picking actions that can lead to desired states.  For example, they can be used to decide how to hit an incoming ball to achieve a desired landing location and speed for the ball.  In other words, they can be used to find the right paddle pose and velocity at the time of contact to achieve a desired land-ball target.  The dynamics models reduce the land-ball task to a paddle-control task with a particular pose and velocity target for the paddle.  Since the paddle needs to hit the ball, the paddle target also includes a time component.


\subsection{Analytic Robot-Control}
\label{sec:method:analytic}

The task and environment decomposition simplify the control task in robot table tennis.  Task decomposition reduces the game-play to accurate paddle-control and environment decomposition allows the robot controller to focus only on accurate paddle-control and not be concerned with the ball, table, or opponent.

Instead of relying on learning, the method relies mainly on analytic control to execute paddle targets.  The target paddle pose is translated to target joint positions using inverse kinematics.  The target paddle velocity is also translated to target joint velocities using the end-effector Jacobian.  The Reflexxes Motion Library is used to compute an optimal trajectory starting with the current joint positions and velocities, and reaching the desired joint positions and velocities at the desired time for contact.  Using an analytic controller instead of learning increases the sample-efficiency of the method.  This controller also makes it possible to switch the robot without having to retrain the paddle-control skill.

The method can thus produce table-tennis agents that follow fixed policies with minimal experimentation on the robot itself.  Experiments are needed only to calibrate the motion constraints of the robot, and to model imperfections in the underlying robot control stack, \eg imperfections in the robot's body and the PID controller.


\subsection{Learning Strategy with Self-Play}
\label{sec:method:strategy}

The method uses a strategy skill whose job is to pick high-level targets for the striking and positioning skills.  In a cooperative game where two agents try to keep a rally going for as long as possible, a good strategy might pick landing targets near the center of the opponent side of the table.  In an adversarial game where the agents try to defeat the opponent, a good strategy might pick targets that make it difficult for the opponent to return the ball.

For tasks like land-ball and paddle-control, it is possible to evaluate any given action and determine whether it accomplishes the task.  However, when looking at the whole game of table tennis and the space of all possible strategies, it is not immediately clear what action is more likely to help an agent win a point.  It is exactly this part of the policy that benefits the most from reinforcement learning and evolutionary strategies and their ability to discover novel solutions.

This skill is trained with a self-play setup involving two robots.  Training happens over a number of self-play levels.  In the first self-play level, the agent plays against a fixed policy.  In every subsequent level, the agent plays against a frozen copy of its most recent policy.  Only one of the two robots is learning during self-play.

The strategy skill is the only skill in the hierarchy that requires every component to be in place.  It requires all tasks controllers to be available.  It also works across the decomposition in the environment as it engages both the robot and game spaces.  Despite these complex interactions, the strategy skill remains relatively simple, since its observation and action spaces are low-dimensional.  Therefore, training the skill requires far fewer episodes compared to training an end-to-end agent.

A key challenge in learning with self-play is maintaining efficient exploration.  Often with self-play learning the agent may converge to a narrow policy.  Since the method uses self-play only for training the strategy, the exploration problem remains confined at the strategy level.  A failure to fully explore the space of all strategies does not reduce the coverage of the dynamics models over the space of game physics, since they are learned independently.  Also, a failure in exploration does not affect the analytic robot controller.  In contrast, in an end-to-end setup a failure in exploration at the highest level may restrict the predictive ability of the underlying components as well.

The strategy skill also accounts for the imperfections in the underlying skills.  The land-ball skill is not perfect.  It misses the target by some error distance based on predictive accuracy of the dynamics models and the precision of the robot controller.  During training, the strategy skill implicitly observes the accuracy of the underlying skills through its reward signal and learns to choose better targets for them.  One common problem with task decomposition is that errors can accumulate through the task hierarchy.  However, since the strategy skill does not have an externally-specified target in the method, it leaves room for the learning agent to compensate for the imperfections in the underlying skills.

\subsection{Conclusion}

To increase sample-efficiency in learning, the method decomposes both the task space and the environment space.  Decomposition the task makes it possible to learn the skills one at a time.  Decomposing the environment allows most skills to focus either on the dynamics of the game, or the dynamics of the robot.  Therefore, task and environment decomposition increases sample-efficiency.  At the same time, the decomposition is done in a manner not to reduce from the generality of the solutions.  The following sections describe each component of the method in detail.

\section{Policy Design}
\label{sec:skill}

This section explains the different tasks that make up the control hierarchy in the method.  The section starts with an overview of the task hierarchy.  It then lists example inputs/outputs for some of the skills.  Then, it discusses each task/skill in more detail.  A high-level overview of the skill implementations are also provided.  The terms task, subtask, and skill are used interchangeably.  Often, skill is used when learning is required and task is used when an algorithmic controller is used.

The main advantage to task decomposition is to create subtasks that be implemented with different mechanisms.  For subtasks that use some form of learning, decomposition improves sample-efficiency.  Also, the training and debugging process is more manageable when focusing on one skill at a time.  Decomposing the task makes it easier to develop and evaluate each component before it is integrated into the whole system.


\subsection{Skill Hierarchy}

The skill hierarchy is shown in \reffig{skill:hierarchy}.  The hierarchy consists of three levels of control (high-level, mid-level, and low-level), two modes (striking and waiting), and seven tasks.  At the top, there is the strategy skill, which just picks targets for the skills in the middle layer.  Each skill in the hierarchy depends on skills below it and used the lower-level skills to accomplish its task.  Except for the strategy skill, all other skills have parameterized targets.  The striking and positioning skills provide an abstraction over the agent's plan for a single exchange with the opponent.  The strategy skill, on the other hand, provides an abstraction over the agent's entire game plan by picking different striking and positioning targets over the course of multiple exchanges as the game goes on.

Decomposing the robot table-tennis task into smaller tasks makes learning easier and more sample-efficient.  The green strategy node is learned with reinforcement learning.  The purple nodes have algorithmic policy implementations using dynamics models trained from human demonstrations.  The blue nodes are implemented by an analytic robot controller.

At any point during the game, the agent is either in striking mode or waiting model.  In striking mode, the agent uses a striking skill to hit the ball toward the opponent.  The agent can strike the ball using land-ball, hit-ball, or directly using the paddle-control skill.  \reffig{skill:hierarchy} shows the three variants of the policy with alternative striking skills in the same image.  In waiting mode, the agent picks a position for itself in anticipation of the opponent's next shot.  This position is specified by a target paddle pose and executed using the positioning skill.  The agent's mode changes automatically based on the current state of the game.  As soon as the opponent hits the ball, the agent enters the striking mode.  The agent stays in the striking mode until it hits the ball back.  At that moment, it enters the waiting mode.

As discussed in subsequent sections, the strategy skill is implemented by a PPO policy trained through self-play (\refsec{strategy}).  The striking skills land-ball and hit-ball are implemented by algorithmic policies (\refsec{striking}) which employ game dynamics models trained from human demonstrations (\refsec{dyn}).  The other four skills in the hierarchy are implemented using analytic control (\refsec{paddle}).  The remainder of this section defines each skill in detail.


\subsection{Strategy}

The strategy skill allows the agent to make high-level decisions about its game plan without being concerned about how they are executed.  Depending on the state of the game, the strategy skill either picks a land-ball target or a positioning target and passes that target to one of its underlying skills.  More specifically, the strategy skill is described by the policy

\begin{equation}
  \pi_s(b_s) = \begin{cases}
    \pi_k(b_s), & \text{if returning the ball}\\
    \pi_w(l(p), \sgn N_x(p)), & \text{if waiting for opponent to return the ball},
  \end{cases}
\end{equation}
\\
where $\pi_s$ denotes the strategy policy, $b_s$ denotes the current ball-motion state\footnote{In the current implementation, the strategy and striking skills receive only the current ball-motion state as the observation.  Experiments have shown that adding other observations like the agent's paddle position $l(p_t)$ or the opponent's paddle position do not improve performance in this implementation.  It is likely that in a setup where the agent can make more than one decision per exchange including such additional observations would be useful.}, $\pi_k$ denotes one of the three striking policies (defined later in this section), $\pi_w$ denotes the positioning policy (defined later), $l(p)$ denotes a target position for the paddle, and $\sgn N_x(p)$ denotes the sign of the $x$ component of the paddle normal vector, specifying a forehand or backhand pose.

The strategy skill separates the tactical dimension of game-play from the technique needed to execute desired movements on the robot.  By providing an abstraction over the game plan, it allows the agent to explore the space of game-play strategies while reusing the technical know-how captured in the underlying skills.  The actions selected by the strategy skill stay active during one ball exchange, which lasts for a relatively long time interval (about 70-100 environment timesteps).  So, it shortens the reward delay between the time the agent makes a decision and when it observes the consequence.


\subsection{Striking Skills}

During an exchange, when the agent is expected to return the ball coming from the opponent, it should execute some action to make contact with the ball and have it land on the opponent's side successfully.  The method offers three variants of the striking policy which achieve this goal with different approaches.  Each learning agent uses only one variant of the hierarchical policy, so for any given agent only one of the three striking policies is available.  The three variants differ in the number of dynamics models they use in their implementation (ranging from three to zero) and are used in experiments to evaluate the impact of dynamics models on learning sample efficiency.  The striking skill $\pi_k$ is specified by one of the three policies
\\
\begin{equation}
  \pi_k(b_s) = \begin{cases}
    \pi_l(g \mid b_s) & \text{land-ball variant}\\
    \pi_h(l_x(p_t), N(p_t), v(p_t), \omega(p_t) \mid b_s) & \text{hit-ball variant}\\
    \pi_p(t, p_t \mid b_s) & \text{direct paddle-control variant},
  \end{cases}
\end{equation}
\\
where $\pi_l$ denotes the land-ball policy (defined later), $\pi_h$ denotes the hit-ball policy (defined later), $\pi_p$ denotes the paddle-control policy (defined later), $b_s$ denotes the current ball-motion state, $g$ denotes a ball landing target, $l_x(p_t)$ denotes the $x$ coordinate of the paddle at time of contact (which maps to the desired distance from the net), $N(p_t), v(p_t), \omega(p_t)$ denote the normal, linear velocity and angular velocities for the paddle, $t$ denotes the time of contact between the paddle and ball, and $p_t$ denotes the full paddle-motion state.

Each of the three striking skills has a different approach to returning the ball.  The land-ball skill parameterizes the strike by a desired landing target for the ball on the opponent's side of the table.  Hit-ball parameterizes the strike by a desired paddle orientation and velocity at the time of contact with the ball.  The hit-ball skill helps the agent make contact by automatically computing the required paddle position based on the predicted ball trajectory.  Striking directly using the paddle-control skill requires fully specifying a target paddle-motion state at the time of contact.  As described in \refsec{striking}, land-ball requires three dynamics models to work, while hit-ball requires only one dynamics model.  The paddle-control skill does not use any trainable dynamics models.  So the three alternative striking skills are used to evaluate the impact of using dynamics models on sample-efficiency of learning the different skills.

\refsec{skill:land-ball} and \refsec{skill:hit-ball} describe the land-ball and hit-ball striking skills.  Striking with direct paddle-control is covered in \refsec{skill:paddle}.

\subsubsection{Land-Ball}
\label{sec:skill:land-ball}

The objective of the land-ball skill is to execute the necessary actions to hit back the opponent ball $b_s$ in a way that it eventually lands at the desired target $g$.  The target $g$ specifies a target location $l(g)$, and a target landing speed $\lvert v(g) \rvert$ over the opponent's side of the table:

\begin{equation}
  g = l(g), \lvert v(g) \rvert
\end{equation}
\\
The target landing location $l(g)$ specifies the desired position of the ball at the moment of landing:

\begin{equation}
  l(g) = l(b_g)
\end{equation}
\\
where $l(b_g)$ denotes ball position at the moment of landing.  Note that there is no constraint on the landing time.  The subscript $g$ here denotes some unspecified landing time.  The target landing speed $\lvert v(g) \rvert$ specifies the desired magnitude of the ball's velocity vector at the moment of landing.

In the current implementation the landing target does not specify a desired landing angle for the ball.  Since the trajectory of the landing ball is constrained at two points (by $b_t$ and $g$), the landing speed specifies the landing angle to some extent, as faster balls tend to have lower landing angles.  However, the implementation can be extended to also include a target landing angle.  In environments where the ball spin affects it motion, topspin would increase the landing angle while backspin would decrease it.  In such environments a desired landing angle can further constrain the desired ball motion.

An example land-ball target can be:

\begin{align*}
l(g) & = (0.9, 0.2) \text{ m}, \\
\lvert v(g) \rvert & = 6 \text{ m/s}.
\end{align*}
\\
In a coordinate system where the center of the table's surface is at $(0, 0, 0.76)$~m, the two-dimensional target $(0.9, 0.2)$ specifies a point $0.9$~m behind the net and $0.2$~m to the left of the center divider (from the agent's perspective) on the opponent's side.  The z coordinate of the target is always equal to $0.78$~m, which is the height of a standard table, $0.76$~m, plus the radius of a table-tennis ball, $0.02$~m.

The land-ball skill is described by the policy

\begin{align}
\pi_l(g \mid b_s) = \pi_p(t, p_t \mid p_s),
\end{align}
\\
where $\pi_p$ denotes the paddle-control policy (defined later), $t$ denotes the planned paddle-ball contact time picked by the land-ball skill, and $p_t$ denotes the target paddle-motion state at time $t$, also picked by the land-ball skill.

The implementation for the land-ball policy is described in detail in \refsec{striking}. Given the incoming ball's motion state $p_s$, the land-ball policy predicts the ball's future trajectory and plans to make contact with it at some time $t$.  The land-ball policy chooses a target motion state $p_t$ for the paddle at the time of contact.  To ensure contact with the ball, the target position for the paddle $l(p_t)$ is always chosen to be equal to the predicted position of the ball $b_t$ at that time.  The land-ball policy also chooses the paddle's target orientation, linear velocity, and angular velocity at the time of contact.  To accomplish its requested goal, the land-ball policy should pick the paddle orientation and velocity such that hitting the ball with that paddle-motion state sends the ball to the landing target $g$.  The target contact time $t$ and the target paddle-motion state $p_t$ computed by the land-ball skill are passed as a high-level action to the paddle-control skill.

The land-ball skill provides a high-level abstraction over the agent's game-play during a single exchange with the opponent.  This high-level abstraction does not cause a reduction in generality of behavior.  Barring deceptive movements to hide the agent's intention from the opponent, any sequence of paddle actions can be reduced to the resulting landing state for the ball.\footnote{A fully-specified landing state should capture the ball's 3D velocity and spin at the moment of landing.  Since the simulator used in this article does not model spin, the implementation uses a simplified representation for the landing state.  However, the land-ball skill can be extended to accept more expressive landing targets.}  In other words, the specification of the land-ball skill makes it possible to specify the agent's behavior by its desired outcome.  Learning to use the land-ball skill is easier for the agents as its action space has fewer dimensions than a fully-specified paddle-motion state, yet its action space can specify complex behaviors.  Furthermore, land-ball's action space maps to the geometry of the world and allows for exploiting the symmetries and invariances that exist in the domain (\refsec{dyn:norm}).


\subsubsection{Hit-Ball}
\label{sec:skill:hit-ball}

The hit-ball skill is an alternative striking skill.  Rather than specifying the strike by how the ball should make contact with the table, the hit-ball skill specifies the strike by how the paddle should make contact with the ball.  Unlike land-ball which aims for a particular point on the opponent's side, hit-ball has not specific target for when the ball lands.  The hit-ball skill is described by the policy
\\
\begin{align}
\pi_h(l_x(p_t), N(p_t), v(p_t), \omega(p_t) \mid b_s) = \pi_p(t, p_t \mid p_s).
\end{align}

The hit-ball skill helps the agent make contact with the ball by computing the time of contact $t$ and the target paddle position $l(p_t)$ based on its inputs and the predicted ball trajectory.  It expects that the agent provide the other contact parameters like the orientation of the paddle encoded by the paddle normal $N(p_t)$ and its linear and angular velocities $v(p_t), \omega(p_t)$.  An example hit-ball target can be:
\\
\begin{align*}
l_x(p_t) & = -1.7 \text{ m}, \\
N(p_t) &= (0.97, -0.02, 0.22), \\
v(p_t) & = (1.23, 1.19, 0.06) \text{ m/s}, \\
\omega(p_t) & = (-0.2, -0.05, 2.85) \text{ rad/s}.
\end{align*}
\\
In this example, the skill is requested to make contact with the ball when it is $1.7$\,m away from the net ($10$\,cm in front of the robot).

The implementation for the hit-ball policy is described in detail in \refsec{striking}. It uses a model to predict the ball's future trajectory.  It then considers the intersection of the predicted ball trajectory with the target contact plane $x = l_x(p_t)$ which is an input parameter to the hit-ball skill.  The point in the predicted trajectory that is closest to this plane is picked as the target contact point.  This point determines the contact time $t$ and the full position of the paddle $l(p_t)$ at the time of contact.  The other attributes of the paddle's motion state $N(p_t), v(p_t), \omega(p_t)$ are given as inputs to the hit-ball skill.  Together with the computed paddle position, they fully specify the paddle's motion state $p_t$, which is passed as a high-level action to the paddle-control skill.

Unlike the land-ball skill which can only execute strikes demonstrated by humans, the hit-ball skill can be used to execute arbitrary paddle strikes.  So a strategy trained over the hit-ball skill can better explore the space of paddle strikes.  At the same time, as the experiments show learning with the hit-ball skill is less sample-efficient and requires more training time.


\subsection{Positioning}
\label{sec:skill:positioning}

The positioning skill is in effect in waiting mode -- when the ball is moving toward the opponent and the agent is preparing to respond to the next ball from the opponent.  The objective of this skill is to move the robot to the requested positioning as as quickly as possible.  Instead of specifying the requested position with the full robot pose, the skill accepts a target paddle position.  This formulation makes the action space of the positioning skill independent of the robot and more likely to transfer to new configurations and new robots.  The positioning skill is described by the policy

\begin{align}
\pi_w(l(p), \sgn N_x(p)) = \pi_p(p \mid p_s),
\end{align}
\\
where $p$ denotes some paddle state that satisfies the requested paddle position $l(p)$ and normal direction indicated by $\sgn N_x(p)$.  Unlike the land-ball skill which requests a specific time target for its desired paddle-motion state, the positioning skill does not specify a time.  In this case, the paddle skill is expected to achieve the desired paddle state as fast as possible.  In other words, it is expected to find some minimum time $m$ and achieve the target paddle $p_m = p$ at that time.  An example positioning target can be:

\begin{align*}
l(p) & = (-2.13, 0.07, 1.02) \text{ m}, \\
\sgn N_x(p) & = +1 \text{  (forehand)}.
\end{align*}

The positioning skill maps the requested paddle position and paddle normal direction to some robot pose that satisfies them.  The positioning policy is discussed in detail in \refsec{skill:positioning}.  In the current implementation, this skill requests a target velocity of zero for the joints at their target positions.  However, a more complex implementation can arrange to have non-zero joint velocities at target to reduce the expected reaction time to the next ball coming from the opponent.


\subsection{Paddle-Control}
\label{sec:skill:paddle}

The objective of the paddle-control skill is to bring the paddle from its current state $p_s$ to a desired target state $p_t$ at time $t$.  This skill is invoked both by the land-ball skill and the positioning skill.  The paddle-control skill is described by the policy

\begin{align}
\label{eq:skill:paddle}
\pi_p(t, p_t \mid p_s) = \pi_t(t, q_t, \dot{q}_t \mid q_s, \dot{q}_s),
\end{align}
\\
where $\pi_t$ denotes the trajectory planning skill (defined later), $q_t$ denotes the target joint positions, $\dot{q}_t$ denotes the target joint velocities, $q_s$ denotes the current joint positions, $\dot{q}_s$ denotes the current joint velocities.  An example paddle target can be:

\begin{align*}
t & = 0.72 \text{ s}, \\
l(p_t) & = (-1.79, 0.22, 1.05) \text{ m}, \\
N(p_t) &= (0.97, -0.02, 0.22), \\
v(p_t) & = (1.23, 1.19, 0.06) \text{ m/s}, \\
\omega(p_t) & = (-0.2, -0.05, 2.85) \text{ rad/s},
\end{align*}
\\
where $l(p_t)$ denotes the target paddle position, $N(p_t)$ denotes the target paddle surface normal, $v(p_t)$ denotes the target paddle linear velocity, and $\omega(p_t)$ denotes the target paddle angular velocity.  The paddle's angular velocity at the time of contact can be used to better control the spin on the ball.  Although the PyBullet simulator does not support ball spin, controlling the angular velocity is useful in real environments.

The paddle-control skill chooses some paddle orientation to satisfy the requested surface normal.  An example paddle orientation can be:

\begin{align*}
r(p_t) & = 0.42 \mathbf{i} + 0.45 \mathbf{j} + 0.51 \mathbf{k} + 0.58, \\
\end{align*}
\\
where $r(p_t)$ is a four-dimensional quaternion representing the paddle orientation and $\mathbf{i, j, k}$ are unit vectors representing the three Cartesian axes.

As discussed in \refsec{paddle}, the policy for the paddle-control skill can be implemented analytically, thereby reducing the need for training and increasing the sample-efficiency of the method.  The analytic controller uses inverse kinematics and the robot Jacobian to translates the target paddle-motion state into a set of joint positions and velocities $q_t, \dot{q}_t$ such that achieving those joint states brings the paddle to the state $p_t$.  It then passes the target joint positions and velocities to the trajectory planning skill.  An example target joint state for a 6-DOF robot assembly can be:

\begin{align*}
t & = 0.72 \text{ s}, \\
q_t & = 0.71 \text{ m}, (-1.31, 0.50, -0.35, -0.31, -0.17) \text{ rad}, \\
\dot{q}_t & = 1.06 \text{ m/s}, (2.23, -0.02, 0.19, 0.61, -0.09) \text{ rad/s}.
\end{align*}
\\
where the first DOF corresponds to the prismatic joint at the base of the robot assembly, and the next five DOFs correspond to the revolute joints in the arm.

The paddle-control skill provides a high-level abstraction over robot-control for table tennis.  It allows higher-level policies to specify their desired robot actions by just specifying the impact of those actions on the paddle at a particular point in time.  At the same time, this high-level abstraction does not cause a reduction in generality of behavior.  Barring deceptive movements, any sequence of joint commands can be reduced to and represented by the final paddle-motion state at the time of contact.  It is only during the short contact time that the state of the robot affects its impact on the ball.  Lastly, since the paddle-control skill works with long-term actions lasting throughout one ball exchange, it allows the agent to observe the reward for its actions with little delay and learn more easily.


\subsection{Joint-Trajectory Planning}

The trajectory planning skill is responsible for bringing the joints from their current positions and velocities $q_s$ and $\dot{q}_s$ to their target positions and velocities $q_t$ and $\dot{q}_t$ at the requested time $t$, while observing the motion constraints on each joint.  By doing so, it brings the paddle from its current state $p_s$ to the target state $p_t$.  The trajectory planning skill is used only by the paddle-control skill and is described by the policy

\begin{align}
\begin{split}
\label{eq:skill:trajectory}
\pi_t(t, q_t, \dot{q}_t \mid  q_s, \dot{q}_s) = & \pi_r(\{ (q_j, \dot{q}_j, \ddot{q}_j) \mid s \le j \le t \}),\\
& \text{subject to:} \\
& q_\text{min} \leq q_j \leq q_\text{max}, \\
& \dot{q}_\text{min} \leq \dot{q}_j \leq \dot{q}_\text{max}, \\
& \ddot{q}_\text{min} \leq \ddot{q}_j \leq \ddot{q}_\text{max}, \\
& \dddot{q}_\text{min} \leq \dddot{q}_j \leq \dddot{q}_\text{max}, \\
\end{split}
\end{align}
\\
where $\pi_r$ denotes the joint-control skill (defined later), $q_j, \dot{q}_j, \ddot{q}_j, \dddot{q}_j$ denote the planned joint positions, velocities, accelerations, and jerks at times $s \le j \le t$, the motion constraints $q_\text{min}, \dot{q}_\text{min}, \ddot{q}_\text{min}, \dddot{q}_\text{min}$ specify the minimum joint positions, velocities, accelerations, and jerks, and $q_\text{max}, \dot{q}_\text{max}, \ddot{q}_\text{max}, \dddot{q}_\text{max}$ specify the maximum joint positions, velocities, accelerations, and jerks.

The trajectory planning task receives long-term targets in the future.  In turn, it generates a series of step-by-step setpoints for the robot to follow on every timestep starting with the current time $s$ and leading to the target time $t$.  Each trajectory setpoint specifies target joint positions and velocities $q_j, \dot{q}_j$.  In the implementation, setpoints are computed at \SI{1}{\kilo\hertz}.  So, for the example given in \refsec{skill:paddle} the computed trajectory would contain 720 points over 0.72 s.

The motion constraints depend on the robot's physical properties, mass, maximum motor torques, etc, and would be different for each robot.  The position limits depend on the configuration of the robot joints and links.  The velocity and acceleration limits depend on the motors and their loads.  Jerk limits are usually imposed to limit the wear and tear on the robot.  The velocity and acceleration limits can be determined from the robot data sheets, or measured empirically.  As explained in \refsec{paddle}, trajectory planning is implemented using the Reflexxes Motion Library~\cite{kroger2011opening}, which can evaluate the feasibility of trajectories and return the next step in under \SI{1}{\ms}.  Reflexxes assumes the motion constraints to be constant and independent of the current joint velocities.  In reality, motors have limit profiles that vary based on their current load.  Still, the limits can be averaged and tuned for the duration of typical motions in table tennis, which last about 0.5 s.

Once the trajectory is computed, it is sent to the robot controller to execute.


\subsection{Joint-Control}

The objective of the joint-control task is to execute the necessary control actions on the robot to follow the next setpoint in the trajectory.  Typically this skill is already realized by an existing PID controller or an inverse dynamics controller.  Then, the joint-control task simply represents the low-level controller on the robot.  The task is described by the policy

\begin{align}
\label{eq:skill:joint}
\pi_r(\{ (q_j, \dot{q}_j, \ddot{q}_j) \mid s \le j \le t \}) = \{u_j \mid s \le j \le t - 1\}
\end{align}
\\
where $u_j$ denotes the control action at time $j$.

At each point in time, the controller observes the current joint states $(q_j, \dot{q}_j, \ddot{q}_j)$, and decides the joint actions or torques $u_j$ to best satisfy the requested next joint states at the next timestep $j + 1$.


\subsection{Conclusion}

The skill hierarchy breaks down the task of playing table tennis into a hierarchy of simpler subtasks that depend on each other.  This breakdown makes it possible to focus on each skill individually and develop and evaluate them separately.  It helps isolate the source of problems and reduces the effort needed for debugging.  Also, learning subtasks can be done with fewer training samples than learning the whole task at once.

The skill definitions do not impose any restriction on how they should be implemented.  In order to achieve higher sample-efficiency, the approach uses reinforcement learning and supervised learning strategically.  The robot-control skills and the positioning skill are implemented using an analytic robot controller that does not require learning.  The striking skills are implemented by algorithmic policies that use dynamics models trained from human demonstrations using supervised learning.  Human games provide observations on the physics of how the ball moves, and how it is affected by contact with the table or player paddles.  Using the demonstrations to train dynamics models instead of policies has the advantage that it allows the agent to exhibit different game-play strategies beyond what is demonstrated by humans.  Since the strategy requires complex reasoning about opponents and is difficult to learn in a model-based way, model-free reinforcement learning is used to learn the strategy skill.

The next section discusses how decomposing the environment simplifies the implementation of skills and improves the method's sample efficiency.  The skill implementations are discussed in future sections.  \refsec{paddle} discusses an analytic controller that implements paddle-control, joint-trajectory planning, and joint-control skills, plus the positioning skill.  The striking skills land-ball and hit-ball are discussed in \refsec{striking}.  Finally, \refsec{strategy} discusses how model-free reinforcement learning is employed to discover creative table-tennis strategies in cooperative and adversarial games using the skills described in this section.

\section{Environment Design}
\label{sec:env}

The skill hierarchy offers a decomposition of the problem on the behavioral axis.  The method also decomposes the robot table-tennis environment into two spaces: the game space, and the robot space.  The decomposition of the environment is shown in \reffig{env:decomposition}.  The game space is concerned with the physics of the table-tennis game involving a table, ball and paddles, independently of how the paddles are actuated.  The robot space is concerned with the control of a robot that has a table-tennis paddle attached to the end of its arm.  The robot space does not deal with the table-tennis ball or the table.  The paddle is the only shared object between the robot space and the table-tennis space.  Also, note that the game space and its constituent objects are exactly the same in robot games and human games, which makes it possible to learn the game dynamics from human demonstrations and use them for robot games.


\subsection{Game Space}

\begin{figure}[htb!]
\centering
\includegraphics[width=0.9\columnwidth]{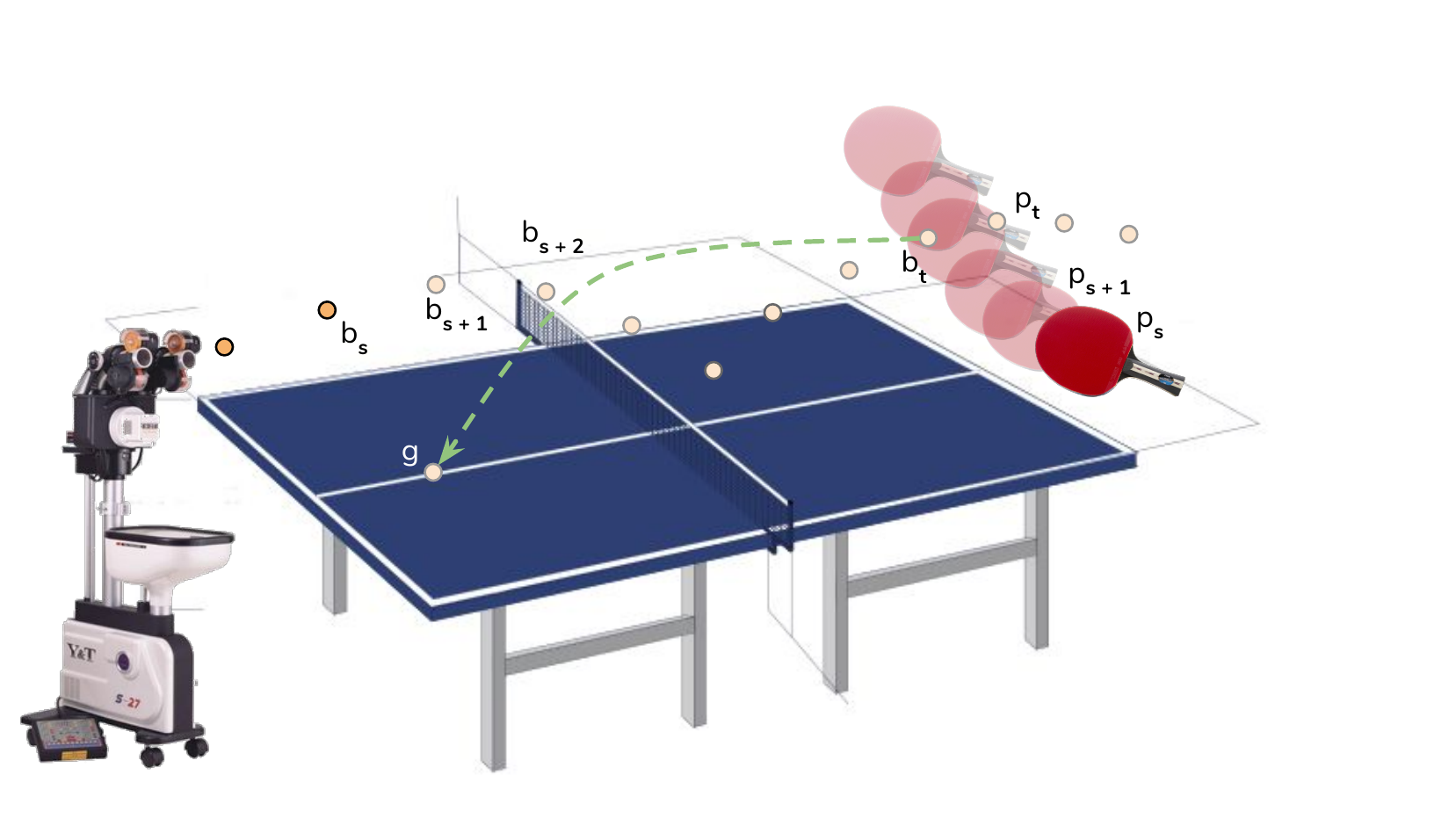}
\mycaption{The Game Space of the Environment.}{The game space includes only the table, paddle, and ball.  The variables $b_s$ and $b_t$ denote the ball motion-state at times $s$ and $t$.  Likewise, $p_s$ and $p_t$ denote the paddle motion-state at times $s$ and $t$.  The game space does not include any robot.  Separating the game objects from the robot permits training dynamics models that predict the dynamics of the game itself independently of how the game is to be played by the robot.  In particular, the game dynamics models can be trained from human demonstrations without using any robot.}
\label{fig:env:gamespace}
\end{figure}

\reffig{env:gamespace} illustrates the game space and visualizes some of the variables which are used in the definition of the skills in \refsec{skill}.  The game space contains only the table, paddle, and ball.  Separating the game objects from the robot makes it possible to learn the dynamics and physical behavior of the objects in the game of table tennis independently of any particular robot.  The data that is relevant to the game space and the dynamics models that are trained in this space can be used by agents driving different robots.  Separating the game physics from robot control increases the sample-efficiency of training skills and policies that depend on them.

\subsection{Robot Space}

\begin{figure}[htb!]
\centering
\includegraphics[width=0.9\columnwidth]{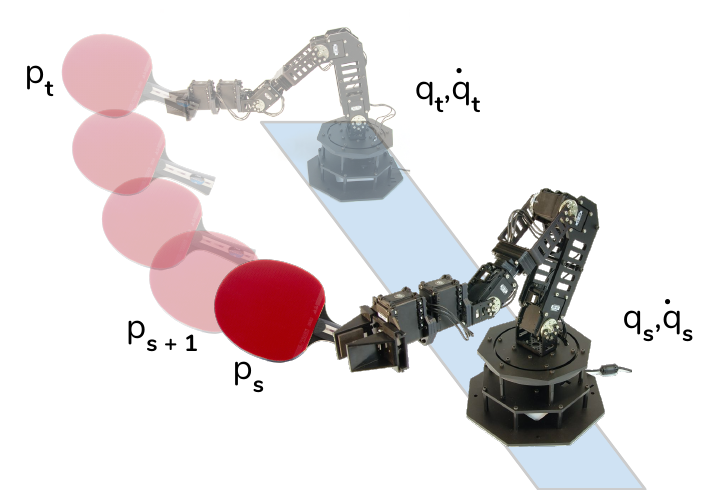}
\mycaption{The Robot Space of the Environment.}{The robot spaces includes only the robot and the paddle attached to it.  The main task in the robot space is paddle-control, which is defined by executing a paddle-motion target $p_t$ starting from the current paddle-motion state $p_s$.  The analytic controller maps the paddle-motion target to joint position and velocity target $q_t, \dot{q}_t$ and executes them using a trajectory-planning algorithm.  The robot space does not include any of the game objects including the ball and table.  Separating the robot from the game objects simplifies the implementation of the paddle-control skill.  In particular, it permits solving the paddle-control problem with an analytic robot controller that does not need training.}
\label{fig:env:robotspace}
\end{figure}

The robot space deals with the control of the paddle attached to the robot's arm.  \reffig{env:robotspace} illustrates the robot space.  The only entities that exist in this space are the links and the joints of the robot assembly and the paddle.  The robot space is not concerned with any of the game objects, including the ball, the table, or any opponent.  The main task in the robot space is paddle-control, which requires taking the paddle from its current motion-state $p_s$ to a target motion-state $p_t$ at time $t$.  Separating the robot from the game objects simplifies the robot-control problem in the hierarchical policy, and reduces it to precise control of the end-effector to achieve target motion-states for the paddle.

\subsection{Separating Physics of the Robot from the Physics of the Game}

The key advantage of decomposing the environment is that it allows for separating the physics of the game objects from the physics of robot control.  This decomposition makes it possible to create models of the interactions between the paddle, the ball, and the table independently of how the robot agent acts in and manipulates the environment.  

Since the game space does not include the robot, it is possible to train dynamics models for it just by observing human games or human practice sessions against a ball launcher.  Such models capture the physics of table tennis and can be used to predict the future states of the game given past observations.  As shown later in \refsec{striking}, these models are used to create a functioning table-tennis agent without using any training episodes.

Furthermore, this decomposition simplifies the problem of robot control, as the robot space does not contain any elements of the game besides the paddle that is attached to the robot.  When attached to the end of the robot arm, the paddle becomes the robot end-effector.  Control of an arm end-effector is a well-studied problem in robotics, so the method uses classic control methods to develop an analytic controller for moving the paddle.  In contrast, when the robot is part of the table-tennis environment, robot control becomes embedded in the agent's behavior.  In such a monolithic environment, the table-tennis agent would need solve the robot control problems as well.  For example, if the agent controls the robot in joint space, it would need to implicitly solve the inverse kinematics problem to be able to interact with objects in the environment.  However, inverse kinematics has an analytic solution and does not require training samples.  Using an analytic controller where possible decreases the complexity of the agent and increases sample-efficiency as no training samples are used by the agent to learn the internal dynamics of the robot.


\subsection{One Game, Different Robots}

An advantage of decomposing the environment into the game and robot spaces is that the robot can be replaced with a completely different type of robot without affecting the dynamics models that are trained on the game space.  All that is needed to get the new robot to play table tennis is to supply a new robot controller that knows how to drive the new robot's end-effector properly.  As long as the new controller can move the paddle attached to the robot as demanded, the new agent can continue to use the existing game dynamics models from the previous setup.  The game strategy learned on an old robot might transfer to some extent to a new robot as well.  However, since different robots have different reaches and different motion constraints, the strategy skill would need to be retrained for best exploit the physics of the new robot.


\subsection{Reduction in Dimensionality}

Another benefit to decomposing the environment is that it lowers the dimensionality of the inputs to the models and policies.  In a monolithic environment, a typical choice for the observation space is to include the joint states in addition to the state of the world objects.  Consider the problem of predicting the trajectory of the incoming ball.  The future states of the ball only depend on the state of the ball and its impact with the table.  Yet, in a monolithic environment, the joint states are always part of the observations.  Similarly, the state of the robot paddle depends only on the joint actions, and is independent of the ball's position.  On the other hand, when learning to land the ball on the opponent's table, the agent needs to line up the paddle with the ball.  For this task, the agent needs to take into account both the state of the paddle and the state of the ball.

In a monolithic environment, more samples or more training episodes are required for the agent to learn when the dimensions are independent and when they interact with each other.


\subsection{Interaction with Task Decomposition}

In the method, environment decomposition is used in combination with the task decomposition in the skill hierarchy.  In the skill hierarchy in \reffig{skill:hierarchy}, the strategy and land-ball skills interact mainly with the game space of the environment.  The positioning skill, the paddle-control skill and the skills below it interact only with the robot space of the environment.  However, the usefulness of decomposing the environment does not depend on using task decomposition as well.  Even with an end-to-end policy which treats robot table tennis as one monolithic task, decomposition of the environment into game and robot spaces can facilitate learning.

In a model-based setup, decomposing the environment allows the agent to learn something about the robot and something about the game from each training episode.  Regardless of how good the current policy is, each training episode provides the agent with a sample from which it can learn how the robot's body responds to the chosen actions.  At the same time, the episode provides the agent with observations showing the dynamics of interactions between objects in the table-tennis environment.

As an example, consider the task of producing striking motions with the goal of making contact with the ball.  In a monolithic environment, as the policy improves and the agent learns to hit the ball, the distribution of the inputs to the dynamics models may change considerably.  In a decomposed environment, the models that predict the impact of joint commands on the motion of the robot paddle will observe the same outcome regardless of whether the strike is successful at making contact with the ball or not.  Suppose at some point the agent learns to make contact.  The experience from before the agent's policy works well transfers fully to after when it learns to make contact.


\subsection{Conclusion}

Separating the environment and robot models creates a setup where the table-tennis dynamics models can be trained independently from the robot's control mechanism.  This separation helps increase sample efficiency during training, as any shortcomings in the robot control stack would not limit the space of observations for the game dynamics models.  Similarly, in the other direction, any shortcomings in the agent's table-tennis strategy would not hinder an exploration in the space of the robot's end-effector behavior.  The next section describes the dynamics models that make predictions over the game space of the environment.

\section{Dynamics Models}
\label{sec:dyn}

This section describes the dynamics models that can make predictions over future states in the game space in the table-tennis environment.  \refsec{dyn:design} describes some of the design choices behind implementing dynamics models as neural networks.  \refsec{dyn:models} describes the ball-trajectory and landing prediction models.  \refsec{dyn:norm} describes the normalization process for reducing the number of dimensions for the dynamics models and thereby reducing the number of samples needed for training them.  \refsec{dyn:data} describes the process for collecting training samples in the VR environment.  \refsec{dyn:e} evaluates the trained dynamics models.


\subsection{Learning Dynamics with Neural Networks}
\label{sec:dyn:design}

This section discusses why the human demonstrations are used in this work to train dynamics models rather than policies, and why the dynamics models are implemented as neural networks rather than physics models or a combination of the two.


\subsubsection{Learning Dynamics Instead of Policy}

Human demonstrations can be used to teach a learning agent about optimal actions to choose.  In other words, the demonstrations can be used to directly train a policy.  In this article, the human demonstrations are used only to train dynamics models of the game space of the environment.  It has the advantage that the dynamics models transfer to different policies.  Also, not using the human policies allows the strategy agent to come up with novel approaches to playing table tennis different from what was tried by the humans.

\subsubsection{Using Physics vs. Neural Networks}
\label{sec:dyn:physics}

A good approach to implementing the dynamics models is to use a combination of physics models and neural networks.  The physics models would include parameters for gravity, coefficients of friction, restitution, the drag and Magnus forces acting on the ball, etc.  The values for these parameters can be obtained by computing a best fit over observations from the training data.  Such physics models can produce an approximation over the future states of the objects.  Then, neural networks can be trained to predict only a \emph{residual} over the predictions from the physics models.  The targets for the neural networks can be obtained by running more accurate offline state estimation models that take complete trajectories into account, including data points that are to be observed in future timesteps.  Using neural networks in combination with physics models is expected to increase the sample efficiency.

However, since the method is evaluated only in simulation, all dynamics models are implemented using neural networks.  The reason is that the physics simulations in Bullet are deterministic.  If the right parameters were included in the physics models, their values could be discovered rather easily, without relying on the predictive ability of the neural networks.  By relying only on neural networks, the experiments evaluate whether the dynamics models are able to model physical interactions.  If they do so in the simulator, they are likely to do so in the real world as well.


\subsection{Dynamics Models}
\label{sec:dyn:models}

The models are used to make two types of predictions:

\begin{enumerate}
\item Future trajectory of the current ball: Once the opponent hits the ball, the learning agent predicts the future trajectory of the incoming ball and picks a desired point and time of contact for hitting the ball back.
\item Landing location and speed resulting from a paddle action: The learning agent uses landing predictions to choose among potential paddle actions to best satisfy the requested landing target.  The landing model predicts the eventual landing location/speed of the ball on the opponent side given the pre-contact ball-motion and paddle-motion states.
\end{enumerate}

The following sections go over each dynamics model in detail.


\subsubsection{Ball-Trajectory Prediction Model}

The ball trajectory prediction model receives the estimate on the current state of the ball and predicts a sequence of ball positions and velocities at future timesteps.  The model produces a full trajectory to give the agent multiple options for picking a contact point for hitting the ball with the paddle.  The model is described by the function

\begin{align}
b_{s+1}, b_{s+2}, \dots, b_{s+n} = B(b_s),
\end{align}
\\
where $B$ denotes the ball trajectory prediction model, $b_s$ denotes the estimate on the current ball-motion state, and $b_{s+1} .. b_{s+n}$ denote predictions on the future ball-motion state over the next $n$ timesteps.

\begin{figure}[htb!]
  \centering
  \includegraphics[width=0.9\columnwidth]{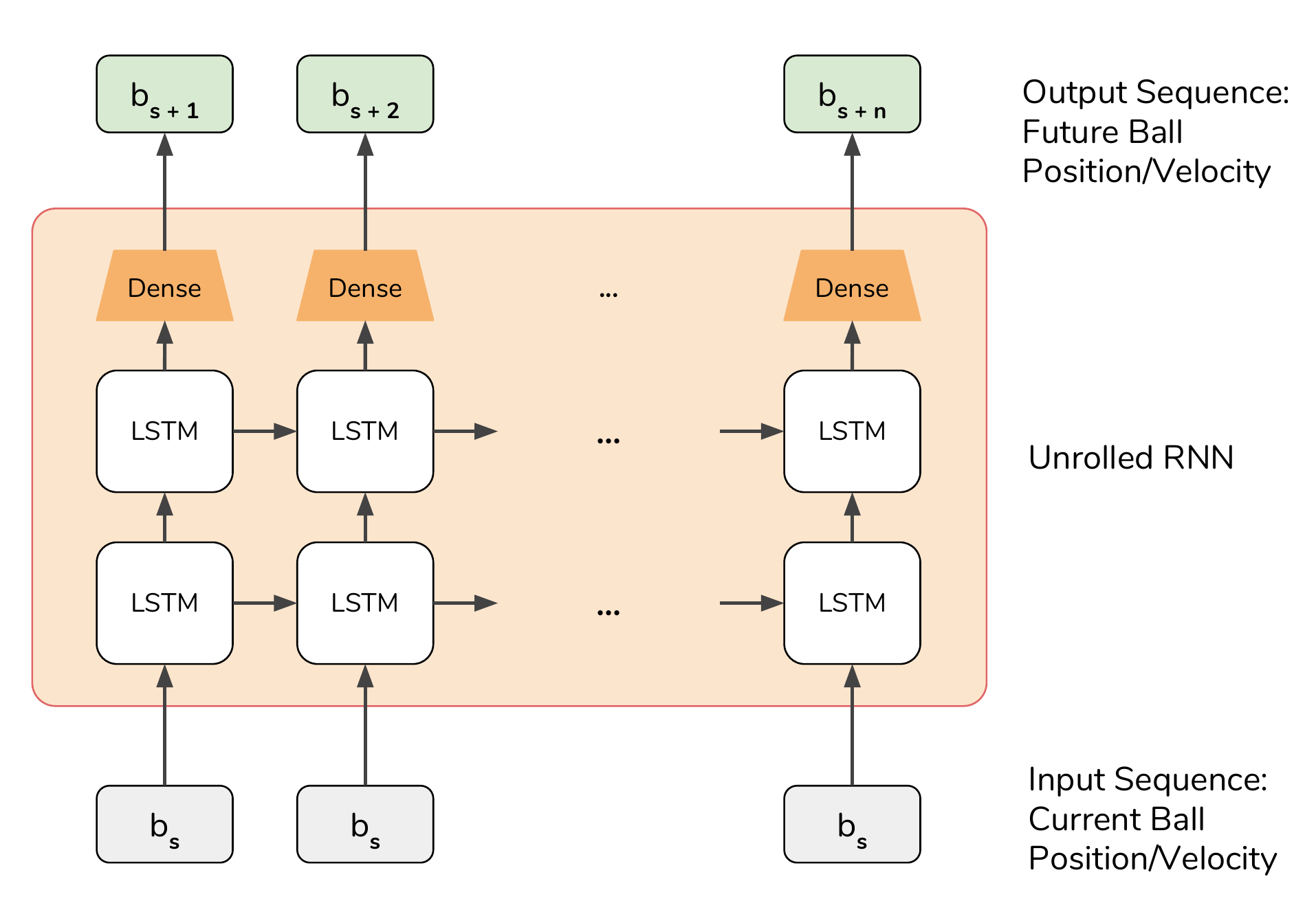}
  \mycaption{Ball-Trajectory Prediction Model.}{Given an estimate on the current position and velocity of the ball, predicts the future position and velocity of the ball over multiple timesteps.  The recurrent model has two LSTM layers followed by a fully-connected layer.  The figure shows the model unrolled through time for better visualization.  At all timesteps, the model receives the same input $b_s$, which represents the current ball position and velocity.}
  \label{fig:dyn:ball2}
\end{figure}

The model's network architecture is shown in \reffig{dyn:ball2}.  It is a recurrent model, however, it receives the same input at every timestep.  The model outputs form a complete trajectory for the ball's future position and velocity over multiple timesteps.

Given the training data, the model learns to predict the physical forces that operate on the ball.  In the simulator, the free motion of the ball in the air is affected by its velocity, gravity, and a drag force that is proportional to the velocity of the ball.  The Bullet simulator does not simulate ball spin and the Magnus forces causing the ball's trajectory to curve.  The motion of the ball after contact with the table is affected by the friction coefficients (lateral, rolling, spinning) between the ball and table, and the restitution coefficients of the ball and table.  In order to predict the future states of the ball, the neural network learn to implicitly model these physical forces.  The ball trajectory may include arcs after its contact with the table.  The training data allows the model to learn about the location and geometry of the table and predict whether the ball will collide with the table and how its trajectory is affected after contact.


\subsubsection{Landing-Prediction Model}\label{sec:dyn:fwdlanding}

The landing-prediction model allows the agent to make predictions about the eventual landing location and speed of the ball given pre-contact ball and paddle-motion states.  Suppose the land-ball agent has predicted a trajectory for the incoming ball and has picked a particular point $b_t$ as a desired contact point for hitting the ball with the paddle.  Given a candidate paddle-motion state at time of contact $p_t$, the landing-prediction model predicts where and with what speed the ball will land on the table if the paddle-motion state is achieved.

The landing-prediction model is described by the function

\begin{align}
\hat{g} = L(p_t, b_t),
\end{align}
\\
where $L$ denotes the landing model, and $\hat{g}$ denotes the ball's predicted position and speed at landing time. 

\begin{figure}[htb!]
  \centering
  \includegraphics[width=0.7\columnwidth]{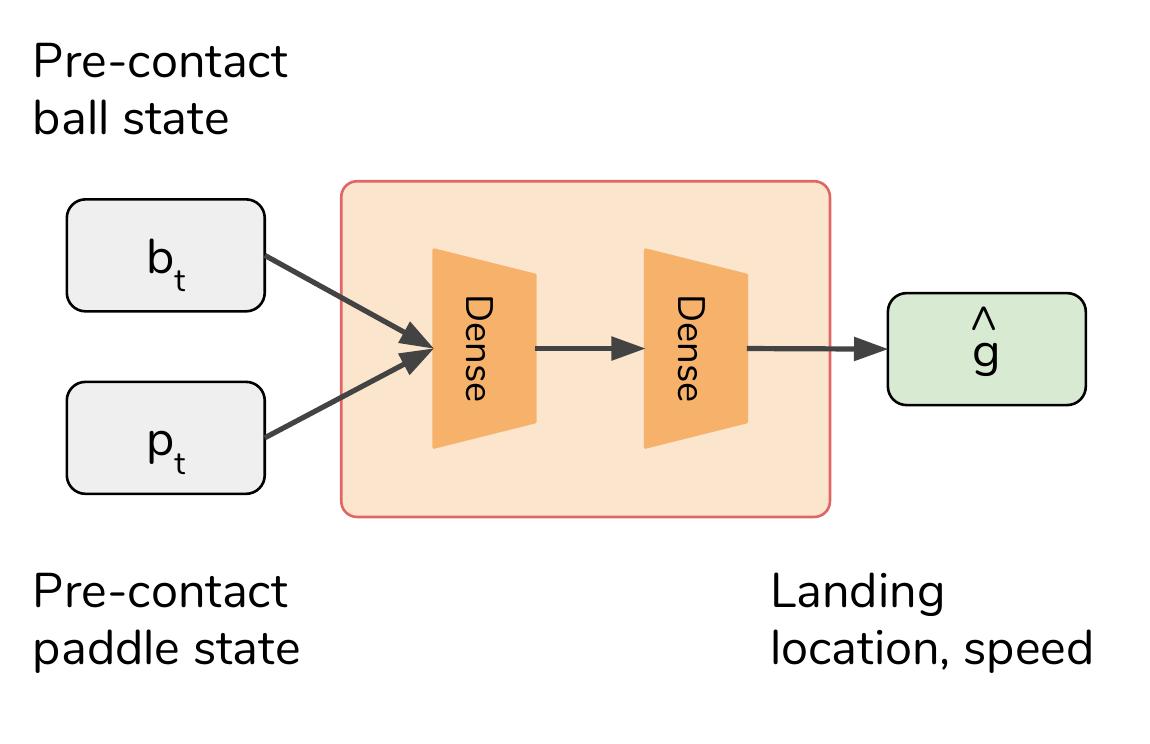}
  \mycaption{Forward Landing-Prediction Model.}{It is a feed-forward network with two fully-connected hidden layers.  Given pre-contact ball-motion and paddle-motion states, the model predicts the eventual landing location and speed of the ball when it hits the opponent side.  Such a prediction can inform the agent about the outcome of available actions.}
  \label{fig:dyn:fwdlanding}
\end{figure}

The model's architecture is shown in \reffig{dyn:fwdlanding}.  It is a feed-forward network, which produces its output in one step.  Since the land-ball policy is only concerned with the eventual landing location and speed of the ball, a feed-forward is preferred since it is faster at inference time and easier to train.

The training data for this model can be limited to the ball trajectories that actually hit the table after contact with the paddle.  Alternatively, the notion of landing location can be extended to include positions off the table.  In that case, the landing location is picked to be the last location for the ball before it falls below the plane of the table's surface.

The landing-prediction model is used by the land-ball skill to search in the space of candidate paddle actions and select one that is more likely to achieve the desired landing target.  In other words, given a set of potential paddle actions $p_k$, the land-ball policy selects the action whose predicted landing location and speed is closest to the requested target $g$.


\subsubsection{Inverse Landing-Prediction Model}

The search process for selecting paddle actions is computationally expensive.  The inverse landing-prediction model addresses this issue by directly predicting the paddle action given a desired landing target.

The landing-prediction model is a forward model, as it predicts the future state of the environment given observations from the current state and an action.  It is possible to train an inverse landing model from the same data that is used to train the forward landing-prediction model.  The inverse landing model is described by the function

\begin{align}
p_t = L^{-1}(b_t, g)
\end{align}
\\
where $L^{-1}$ denotes the inverse landing model.

\begin{figure}[htb!]
  \centering
  \includegraphics[width=0.7\columnwidth]{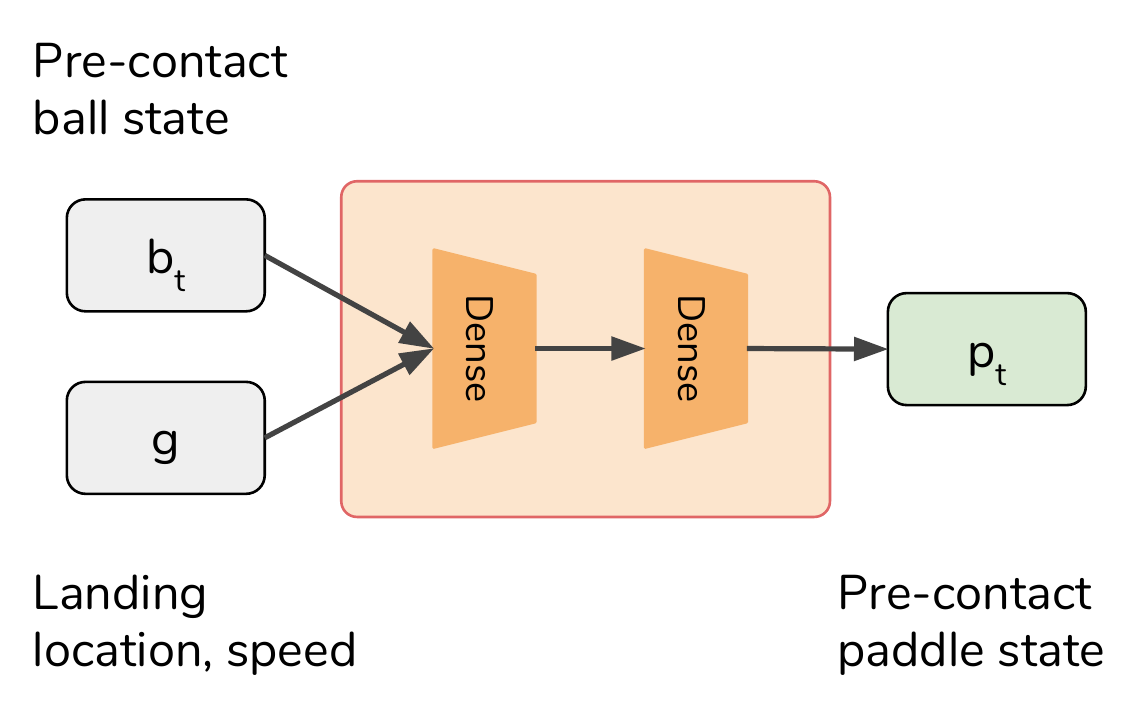}
  \mycaption{Inverse Landing-Prediction Model.}{It is a feed-forward network with two fully-connected hidden layers.  Given pre-contact ball-motion state and a desired landing target, the model predicts the paddle-motion state right before contact.  The predicted paddle-motion can be used as an action for the paddle-control skill.}
  \label{fig:dyn:invlanding}
\end{figure}

The model's architecture is shown in \reffig{dyn:invlanding}.  Given a pre-contact ball-motion state and a desired landing location and speed, the inverse landing model predicts the pre-contact paddle-motion state that would achieve the desired landing location and speed.  The predicted paddle-motion state can be used as an action for the paddle-control skill.

Ignoring the noise in the environment and imperfections in paddle-control, the forward landing-prediction model should have a single answer.  That is, if the pre-contact ball and paddle-motion states are accurately known, there is only one possible outcome for the landing location and speed of the ball.  The same statement does not hold for the inverse landing model.  In general, there might be more than one way to hit a given ball back toward a given target.  However, it is still useful to build an inverse landing model by training a neural network on all observed trajectories from human data.  Such a model would predict the mean of all observed paddle actions taken by humans.  Due to the nonlinearities in the action dimensions, it is possible that the mean action would not be a good solution to the inverse landing problem.  However, it can serve as a starting point for a search using the forward landing model.

The inverse landing model is more complex than the forward landing model, since the paddle-motion state $p_t$ has more dimensions than the landing target $g$.  In the current implementation, $p_t$ is 12-dimensional, while $g$ has up to four dimensions.  Note that some dimensions do not need to predicted.  For example, the paddle position can be decided directly based on the predicted position of the ball at the moment of contact, and the height of the landing target is equal to the height of the table plus the radius of the ball.


\subsection{Domain Invariances and Data Normalization}
\label{sec:dyn:norm}

This section describes the normalizing transformations that are applied to the collected training data when training the dynamics models discussed in \refsec{dyn:models}.

\subsubsection{Invariances in Table Tennis}

Normalizing the data reduces the dimensionality of the inputs and outputs to the models and improves sample efficiency.  Consider a trajectory containing a number of observations on the position and velocities of the ball and the paddle.  In the table-tennis domain, the following invariances hold:

\begin{itemize}
\item \textbf{Translation invariance across $x, y$}:  Shifting the $x, y$ coordinates of the ball and paddle positions by a constant amount for all the points in the trajectory produces a valid trajectory.  This transformation does not affect the object orientations or velocities.
\item \textbf{Rotation invariance around $z$}:  Rotating all object poses and velocity vectors around any line orthogonal to the $x-y$ plane produces a valid trajectory.
\item \textbf{Rotation invariance around paddle normal $N$}:  Rotating the paddle poses around the surface normals produces a valid trajectory.  The force exerted by the paddle on the ball depends on the contact angle between the ball and their velocities before contact, but it is not affected by the placement of the paddle handle.  Rotating the paddle handle around the paddle surface normal does not change the contact dynamics.
\item \textbf{Inversion invariance for paddle normal $N$}:  Inverting all three elements of the paddle normals produces a valid trajectory.  Flipping the paddle normal amounts to changing a forehand to a backhand.  As long as the paddle position and velocity stay the same, a forehand and backhand contact have the same impact on the ball.
\end{itemize}

These variances can be exploited when training the dynamics models from observation trajectories in two ways:

\begin{enumerate}
\item Data Augmentation:  For any collected trajectory, random perturbations based on the explained invariances can be applied to all points consistently to generate augmented training trajectories.
\item Data Normalization:  The collected trajectories can be normalized to remove the redundant dimensions.
\end{enumerate}

Data augmentation has the advantage that it has has a simple implementation; it just results in generating more data.  The clients that query the dynamics models do not need to be modified.  Another advantage to data augmentation is that it can factor in the position of the table and its impact on the trajectory.  For example, a ball that bounces once on the table may not hit the table at all if its $x$ and $y$ coordinates are shifted too much.  Similarly, if an augmented trajectory hits the net, it can be removed.  A disadvantage of data augmentation is that it introduces additional hyperparameters.  What translation and rotation values are likely to generate trajectories that are valid and likely to occur during the game?  How many augmented trajectories should be generated from each recorded trajectory to capture different types of transformations over object states sufficiently?  Lastly, the expected accuracy of the data augmentation approach is upper bounded by the data normalization approach.

The data normalization approach does not add any new trajectories.  Instead, it just modifies the collected trajectories to remove the redundant dimensions.  For example, all ball trajectories can be shifted so that the initial $x, y$ coordinates of the ball are at $(0, 0)$.  It has the advantage that it does not introduce any hyperparameters and does not increase the size of the training dataset.  Also, reducing the number of dimensions simplifies the problem for the neural network, whereas with data augmentation the neural network needs to capture the invariances in its hidden layers.  The disadvantage of normalization is that it cannot model the table.  For example, normalizing the trajectory of a bounced ball assumes implicitly that the ball will bounce in any direction and will never hit the net.  Lastly, querying a model trained on normalized data requires normalizing the inputs and un-normalizing the outputs.  Therefore, it complicates the implementation.

The method implemented in this article uses data normalization.  Since the models are not aware of the location of the table and the net, it is up to the higher-level skills like strategy to pick targets that are feasible and increase the agent's reward.  In the case of the ball-trajectory prediction model, it is harmless to assume that the opponent's ball will always bounce on the player's table.  The agent can plan to respond to a bouncing ball.  If that does not happen, the agent simply wins a the point.  For the landing model, the strategy skill is expected to pick targets such that the actions recommended by the models are effective.  For example, if the strategy skill picks a landing target close to the net, it should pick a lower landing velocity to avoid hitting the net.  Therefore, the dynamics models do not need to be aware of the location of the table and the net.

The following section explains the normalizing transformations that are applied to the data used for training each dynamics model.


\subsubsection{Normalizing Ball Trajectories}

Ball trajectories are normalized as follows:

\begin{enumerate}

\item All ball-motion states $\{b_0, \ldots, b_n\}$ are shifted such that the $x, y$ coordinates of the first ball-motion state $b_0$ become zero.  Suppose the position of the first ball $b_0$ in the original trajectory is

\begin{align}
l(b_0) = l_x(b_0), l_y(b_0), l_z(b_0),
\end{align}
\\
where $l_x, l_y, l_z$ denote the $x, y, z$ coordinates of the ball's position.  Then all points $b_i$ in the original trajectory are transformed to points $b'_i$ in the normalized trajectory as

\begin{align}
l(b'_i) \gets l(b_i) - (l_x(b_0), l_y(b_0), 0).
\end{align}
\\
In particular, the first point is transformed to $b'_0$ such that
\begin{align}
l(b'_0) = 0, 0, l_z(b_0).
\end{align}

This transformation does not affect the ball velocity vectors.

\item All ball positions and velocity vectors are rotated such that the $y$ component of the velocity vector of the first ball $b_0$ becomes zero.  More specifically, the objects and their velocity vectors are rotated around the vertical line

\begin{align}
x = l_x(b_0), y = l_y(b_0)
\end{align}
\\
by an an angle $\psi$ equal to
\\
\begin{align}
\psi = \angle(\text{proj}_{z=l_z(b_0)} v(b_0), (1, 0, 0)),
\end{align}
\\
where $\angle$ denotes the angle between the two enclosed vectors, $\text{proj}_{z = l_z(b_0)}$ denotes the projection of the velocity vector $v(b_0)$ onto the horizontal plane specified by $z = l_z(b_0)$, and $(1, 0, 0)$ is the unit vector parallel to the $x$ axis.

\end{enumerate}

These transformations remove three of the six dimensions in the input to the ball-trajectory prediction model.  Therefore, they simplify the job of the neural network that is modeling ball trajectories.


\subsubsection{Normalizing Landing Trajectories}

Landing trajectories are normalized as follows:

\begin{enumerate}

\item All ball-motion states $\{b_0, \ldots, b_n\}$ and paddle-motion states $\{p_0, \ldots, p_n\}$ are shifted such that the $x, y$ coordinates of the pre-contact ball $p_t$ become zero.  Suppose the position of the pre-contact ball $b_t$ in the original trajectory is

\begin{align}
l(b_t) = l_x(b_t), l_y(b_t), l_z(b_t).
\end{align}
\\
Then all ball-motion states $b_i$ and paddle-motion states $p_i$ in the original trajectory are transformed in the normalized trajectory as

\begin{align}
l(b'_i) \gets l(b_i) - (l_x(b_0), l_y(b_0), 0), \\
l(p'_i) \gets l(p_i) - (l_x(b_0), l_y(b_0), 0).
\end{align}

This transformation does not affect any velocity vectors or paddle orientations.

\item All ball and paddle poses and velocity vectors are rotated such that the $y$ component of the velocity vector of the pre-contact ball $b_t$ becomes zero.  More specifically, the objects and their velocity vectors are rotated around the vertical line

\begin{align}
x = l_x(b_t), y = l_y(b_t)
\end{align}
\\
by an angle $\psi$ equal to
\\
\begin{align}
\psi = \angle(\text{proj}_{z=l_z(b_t)} v(b_t), (1, 0, 0)).
\end{align}

\item All paddle orientations are replaced by paddle normals.
\item All paddle normals with negative $x$ components are inverted.  In other words, all backhand paddles are replaced with forehand paddles.

\end{enumerate}

The first two transformations above remove three of the six dimensions from the ball-motion state input to the landing-prediction model.  The third transformation removes one of the 13 dimensions from the paddle-motion state input to the model.  The last transformation cuts the space of three of the paddle-motion state dimensions in half.  Therefore, normalizing the landing trajectories makes it easier for the neural network to predict landing targets.

It is useful to note that the trajectories are not invariant to translation across $z$.  Changing the height of the ball and paddle invalidates the trajectory if the ball contacts the table at any point.  On the other hand, for the landing-prediction model, data augmentation is used to generate augmented trajectories with \emph{reduced} ball and paddle heights.  The landing model is not concerned with the motion of the ball after its contact with the table.  Given a ball trajectory that collides with the table at the end, it is possible to compute where the collision point would be if the ball was shot from a lower height.  The same does not hold for increasing the height of the ball.  This property is used to generate multiple landing trajectories with lower heights from each original landing trajectory.


\subsection{Learning Dynamics from Demonstrations}
\label{sec:dyn:data}

The data for training the dynamics models is collected in a VR environment that is integrated with the simulation environment.  The VR environment is specially-designed for the purpose of data collection only.  The paddle-strike trajectories and ball-motion data collected from human demonstrations 

\subsubsection{Data Collection in VR Environment}

\begin{figure}[htb!]
  \centering
  \includegraphics[width=0.9\columnwidth]{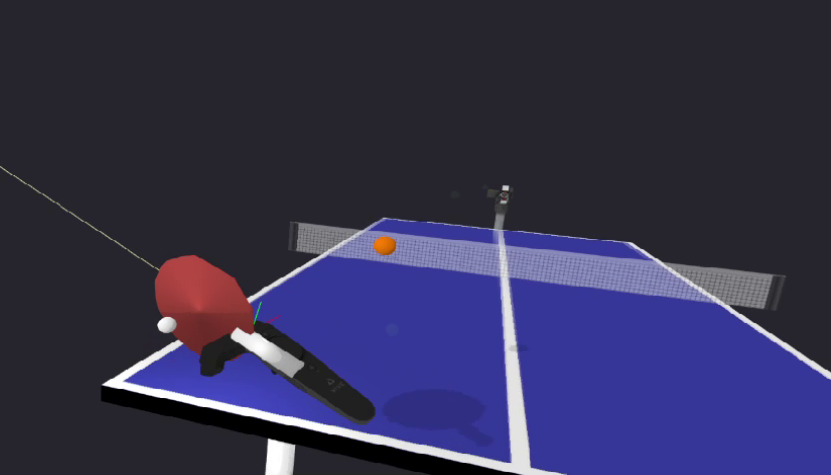}
  \mycaption{Data Collection in VR Environment.}{The VR environment allows a human player to control a simulated paddle by moving the VR controller in the real world.  The simulated paddle follows the VR controller (visualized as the black object holding the paddle).  The player returns the balls coming from a ball launcher on the other side of the table.  The paddle and ball trajectories are recorded and used in training the dynamics models.}
  \label{fig:dyn:vr-data}
\end{figure}

\reffig{dyn:vr-data} shows the data collection process in the VR environment.  A player is controlling the simulated paddle by moving the real VR controller in their hand.  The player returns the balls coming from a ball launcher on the other side of the table.  The paddle and ball trajectories are recorded and used in training the dynamics models.

Since the VR hardware has only one headset and does not support two players, the data collection environment emulates a two-player game by having the ball launcher match the direction and velocity of the balls that are successfully returned by the player.  If the player lands the ball on the other side, the simulator sends the next ball to the corresponding location on the player's side of the table, making the player respond to their own shot.  This setup allows the distribution of ball trajectories to be closer to what might be observed in a real two-player game.  If the player sends the ball out, the next ball is shot from the ball launcher.

Once the data is collected, it is used to extract ball-motion trajectories and landing trajectories.  Ball motion trajectories start at the moment when the ball launcher throws the ball and include the motion state of the ball in every subsequent timestep.  Landing trajectories start a few timesteps before contact between the ball and paddle happens and continue until the ball lands on the opponent side, or crosses the plane at the surface of the table.  The ball-trajectory prediction model is trained on the ball trajectories and the landing-prediction model is trained on the landing trajectories.  For training the landing-prediction model, only two timesteps of the trajectory are used: a timestep before contact happens, and the final timestep when the ball has landed or has gone out.  Since the dynamics models are normalized, a ball that goes out contains useful information as well, since the same paddle motion can be useful for landing a similar ball from a different region of the table.

\subsubsection{Data Augmentation}

The data collected from human demonstrations contained about 300 successful paddle strikes where the human player was able to make contact with the ball.  Strikes where the ball hit the edge of the paddle were removed from the dataset, since that type of contact is not a behavior from which the agent is expected to learn.  To speed up data collection, a data augmentation process was used to generate more samples from the 300 human demonstrations.  During data augmentation, the paddle and ball trajectories observed during demonstrations were replayed in the simulator with small amounts of noise to produce additional samples.  Each additional sample created this way is counted as an extra sample.

\subsubsection{Subsampling}

A subsampling process is used to extract multiple training samples from each trajectory.  Since the sensors have a frequency higher than the environment, each trajectory can be subsampled with multiple offsets.  For example, in a landing trajectory, there are multiple sensory observations of the ball and paddle in the 20 milliseconds (one environment timestep) leading up to contact between the ball and paddle.  Any one of those observations can be used for training the forward and inverse landing-prediction models.  So, with the subsampling process, every landing trajectory can be used to produce 20 samples.

In addition, the ball trajectories are replicated by considering any of the observations in the first ten timesteps as the starting point for the remainder of the trajectory.  The training samples extracted with subsampling were not counted as additional samples.

Although not attempted in the current implementation, there is also an opportunity to extract additional landing trajectories by considering a height reduction and computing where the landing location would have been if the height of the paddle and the ball at the moment of contact were reduced.  The new landing location can be computed as the intersection of the observed trajectory and an imaginary plane of where the table would be given the height adjustment.  This same process can not be used for augmenting free-moving ball trajectories, as the ball-trajectory prediction model is expected to predict the behavior of the ball after it bounces off the table as well, and that behavior can not be computed since reducing the initial height of the ball changes the contact parameters for when it hits the table.  The height reduction technique was not used in the current implementation.





%


\subsection{Evaluation}
\label{sec:dyn:e}

To better evaluate the impact of the number of samples on the accuracy of the dynamics models, they are trained with two dataset sizes.  The smaller dataset contains about 7,000 successful paddle strikes where the human player was able to make contact with the ball.  The larger dataset contains 140,000 strikes.  Once trained, the dynamics models are evaluated on 1000 strikes generated against a ball launcher.  These strikes are not part of the training or evaluation datasets.  The observed ball and paddle trajectories resulting from the 1000 strikes are recorded and compared against the predictions from the dynamics models.  The following sections report the mean errors for the ball-trajectory and landing-prediction models.

\subsubsection{Ball-Trajectory Prediction}

\begin{figure}[htb!]
  \centering
  \includegraphics[width=0.9\columnwidth]{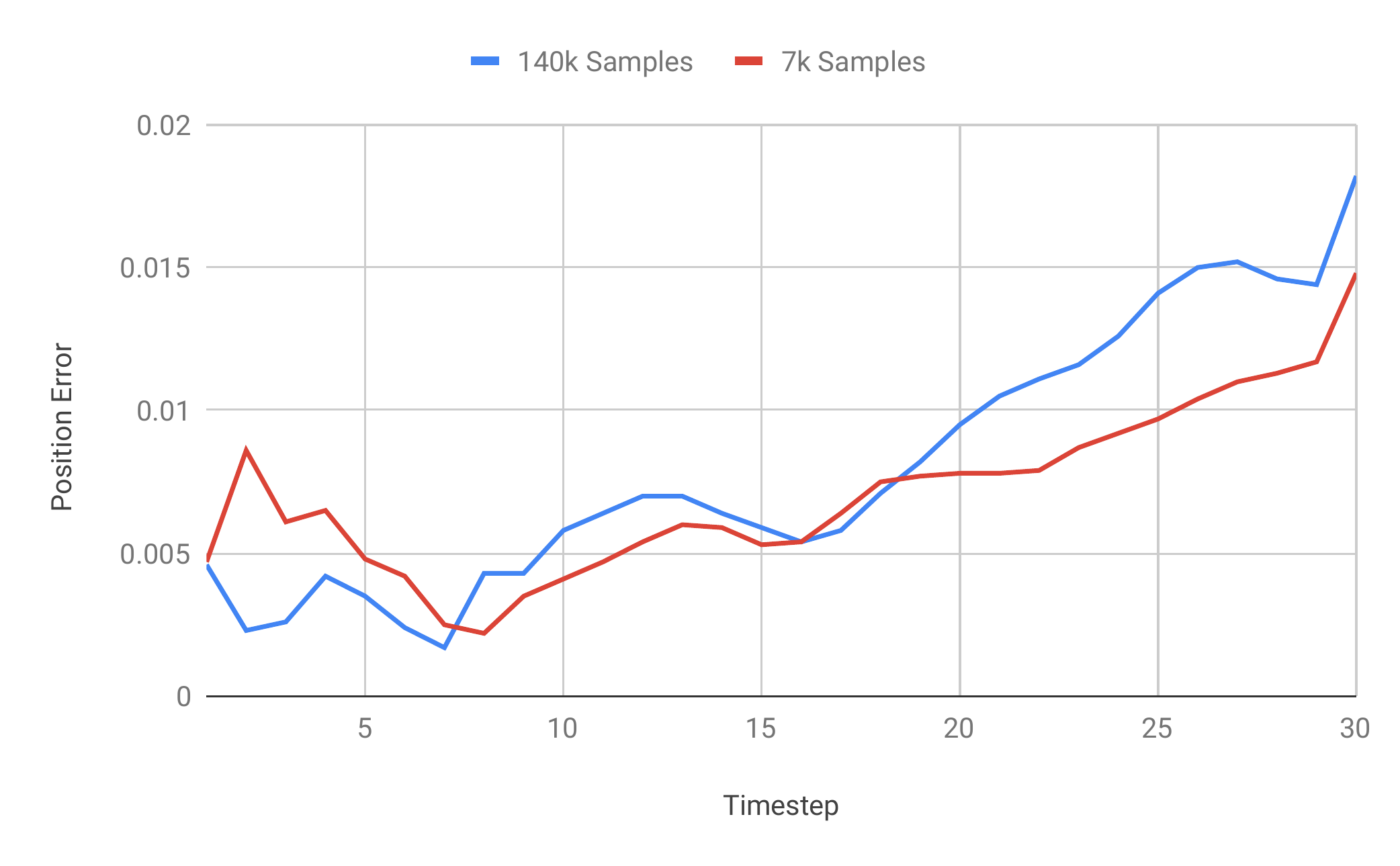}
  \mycaption{Mean Position Error in Ball-Trajectory Predictions.}{The plot shows the mean position error over 1000 ball trajectories containing 30 timesteps of observations each.  The error reported is the Euclidean distance between the predicted position and the observed position of the ball.  The average position error stays less than 1 cm after 25 timesteps, which suggests that the approach is effective.}
  \label{fig:edyn:pos}
\end{figure}

\begin{figure}[htb!]
  \centering
  \includegraphics[width=0.9\columnwidth]{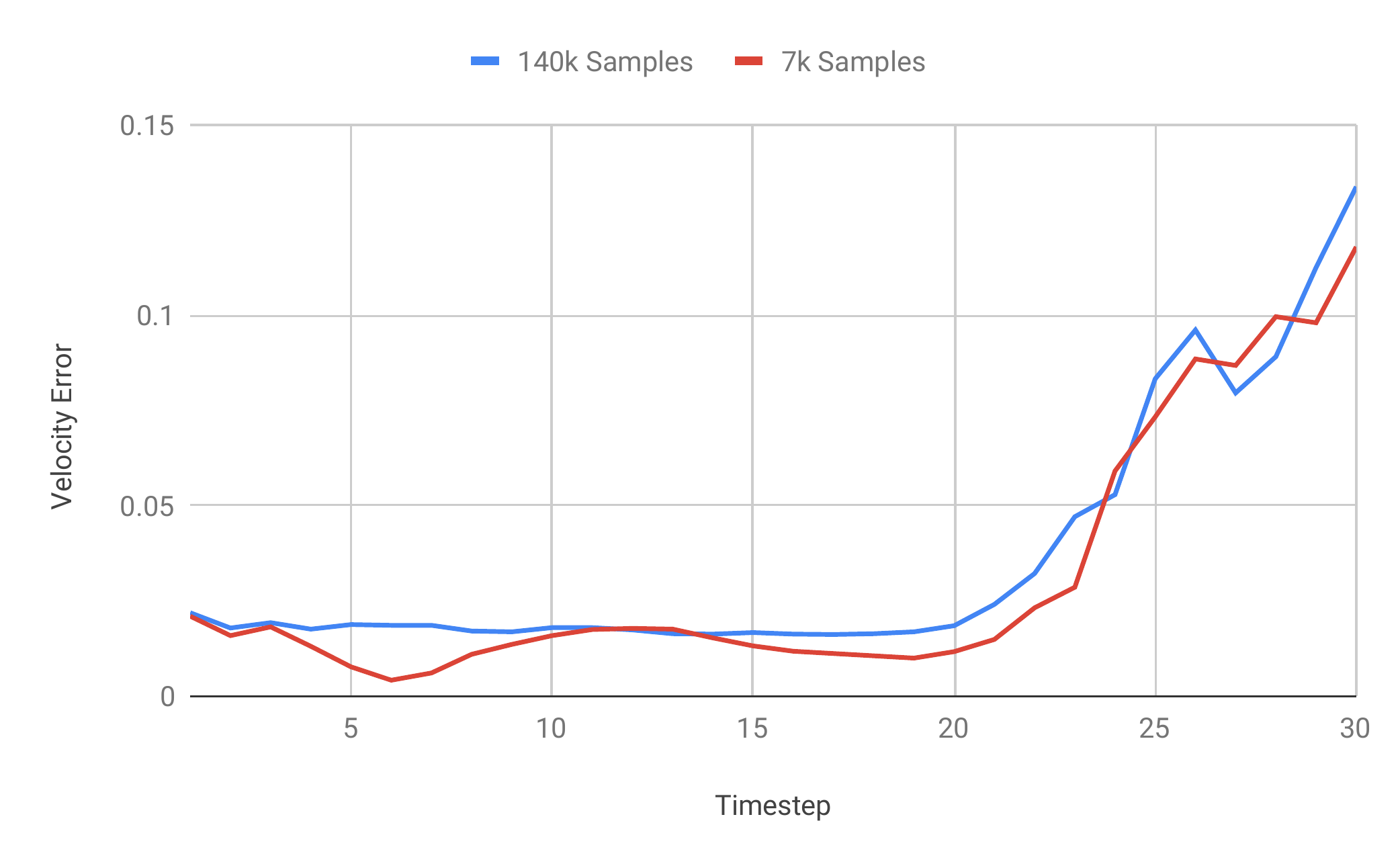}
  \mycaption{Mean Velocity Error in Ball-Trajectory Predictions.}{The plot shows the mean velocity error over 1000 ball trajectories containing 30 timesteps of observations each.  The error reported is the Euclidean distance between the predicted 3D velocity  and observed 3D velocity vectors for the ball.  The average velocity error stays around 0.02 m/s for the first 20 timesteps and starts climbing from there.  The 20th timestep is around the time when the ball hits the table.  It is likely that predicting the behavior of the ball after contact is more challenging than predicting its free motion for the model.  At any rate, the prediction error on velocity remains low compared to the magnitude of observations (around 6 m/s), which suggests that the approach is effective.}
  \label{fig:edyn:vel}
\end{figure}

\reffig{edyn:pos} and \reffig{edyn:vel} show the average position and velocity errors in predicted ball trajectories.  Each evaluated ball trajectory contains 30 timesteps of observations corresponding to 0.6 seconds of data.  This amount of time is usually enough for the ball to cross the table and reach the striking player.  As the plots show, the average position error stays less than 1 cm after 25 timesteps, and the average velocity error stays less than 0.1 m/s after 25 timesteps.

There is little difference between the accuracy of the model trained on the large dataset and the model trained on the small dataset.  With data normalization, the number of inputs to the ball-trajectory prediction model is reduced to three.  Moreover, the subsampling process generates many more samples from each recorded ball trajectory.  In addition, the physical laws governing the behavior of a bouncing ball are relatively simple to learn.  Therefore, it seems this model does not need many samples to learn to predict the behavior of the ball.

\subsubsection{Landing Prediction}

\reftab{edyn:fwdlanding} shows the mean position error over 1000 landing predictions from models trained on the small and large datasets.  The error is about 0.19 m when the model is trained on 7,000 samples and about 0.114 m when the model is trained on 140,000 samples.  The landing-prediction model is more complex than the ball-trajectory prediction models, since its inputs include both the ball and paddle states.  Moreover, the model is expected to predict the eventual position of the ball after it has travelled for 1-2 meters.  These reasons might be why the landing-prediction model benefits from more training data.

\begin{table}[htb!]
\centering
  \begin{tabular}{|r|r|}
  \hline 
  \thead{Samples} & \thead{Mean Position Error} \\
  \hline 
  7,000 & 0.190 m \\
  140,000 & 0.114 m \\
  \hline 
  \end{tabular}
\mycaption{Mean Position Error for the Landing-Prediction Model.}{Mean position error for models trained from 7,000 trajectories and 140,000 trajectories.}
\label{tab:edyn:fwdlanding}
\end{table}


\subsection{Conclusion}

The dynamics models discussed in this section are only concerned with the physics of the table-tennis environment.  They do not deal with the physics of the robot.  Therefore, the models can be trained from data collected from human games or practice sessions against a ball launcher.  The evaluations show that models are able to predict the motion of the ball over multiple timesteps in the future.  There is no noise in the simulation, which makes the simulation outcomes deterministic and easier to predict.  Yet, the experiment results show that the models have the ability to capture the physics of interactions between the objects in the environment.  \refsec{disc} describes an extension to the method that can handle observation noise as well.  The next section describes the analytic paddle controller that can be used to execute target paddle-motion states to land a given ball at a desired target.

\section{Paddle-Control Policy}
\label{sec:paddle}

This section discusses the implementation of the paddle-control skill.  \refsec{paddle:problem} revisits the definition of the paddle-control task and defines some variables used in the rest of the section.  \refsec{paddle:analytic-control} describes an analytic paddle controller, which is derived mathematically based on the kinematics of the robot links and the motion constraints on the robot motors. \refsec{paddle:analytic-model} describes an analytic paddle-dynamics model that allows higher-level skills to make predictions about expected paddle-motion states resulting from executing the high-level paddle-motion targets with the paddle-control skill.  Lastly, \refsec{paddle:learning} discusses an alternative implementation for the paddle-control skill that uses learning.

To increase sample-efficiency, the method in this article uses the analytic paddle controller and does not rely on training to learn the internal dynamics of the robot.  The alternative controller that uses learning is studied in an ablation experiment in \refsec{striking}.


\subsection{Paddle-Control Problem}\label{sec:paddle:problem}

As described in \refsec{skill:paddle}, the objective of paddle-control skill is bring the paddle from its current motion-state $p_s$ at time $s$ to the desired paddle-motion target $p_t$ at time $t$.  The target requested from the paddle skill includes the paddle pose $x(p_t)$ and its time derivative $\dot{x}(p_t)$:

\begin{align}
p_t = x(p_t), \dot{x}(p_t).
\end{align}

The paddle pose in turn includes the paddle position $l(p_t)$ and surface normal $N(p_t)$:

\begin{align}
x(p_t) = l(p_t), N(p_t).
\end{align}

Note that in this formulation, $p_t$ does not fully specify the paddle pose, as there are generally many possible paddle orientations that satisfy the specified paddle normal $N(p_t)$.  Specifying the paddle pose with its surface normal instead of a fully-specified orientation has the advantage that it gives the paddle skill the freedom to choose any orientation that satisfies the normal.  Also, it is easier to replicate paddle normals from human demonstrations than paddle orientations.  The forces exerted from the paddle on the ball at the moment of contact depend on the paddle's normal and stay fixed if the paddle is rotated around its surface normal vector.  So any possible orientation that satisfies the given normal will have the same impact on the ball.

The time derivative of the paddle pose $\dot{x}(p_t)$ includes the paddle's linear velocity $v(p_t)$ and angular velocity $\omega(p_t)$:

\begin{align}
\dot{x}(p_t) = v(p_t), \omega(p_t).
\end{align}

The paddle's linear and angular velocity at the time of contact affect the forces exerted on the robot and affects the ball's trajectory after contact.


\subsection{Analytic Paddle-Control}\label{sec:paddle:analytic-control}

The analytic paddle controller uses 3D geometry, inverse kinematics, the robot Jacobian, and Reflexxes trajectory planning to achieve the paddle-motion target $p_t$.  It works through the following steps to move the paddle from its current state $p_s$ at time $s$ to a desired state $p_t$ at time $t$:

\begin{enumerate}
\item Find a paddle orientation $r(p_t)$ that satisfies the desired paddle normal in $N(p_t)$.
\item Map the target paddle pose $x(p_t)$ to target joint positions $q_t$.
\item Map the target paddle velocity $\dot{x}(p_t)$ to target joint velocities $\dot{q}_t$.
\item Compute a joint-trajectory starting with current positions and velocities $q_s, \dot{q}_s$ and reaching the target joint states exactly at time $t$.
\item Use robot's controller (\eg a PID controller) to execute joint commands between times $s$ and $t$ to follow the computed trajectory.
\end{enumerate}

The following sections describe each step in detail.


\subsubsection{Mapping Paddle's Normal to Orientation}

The analytic controller computes a paddle orientation $r(p_t)$ based on the requested paddle position $l(p_t)$ and surface normal $N(p_t)$.  First, it uses inverse kinematics to find an intermediate pose that satisfies only the requested position $l(p_t)$:

\begin{align}
\label{eq:paddle:ikc}
q^l_t \gets \text{IK}_c(l(p_t)),
\end{align}
\\
where $\text{IK}_c$ denotes the inverse kinematics function starting with canonical robot rest poses, and $q^l_t$ denotes the joint positions corresponding to an intermediate pose that satisfies $l(p_t)$.

In the coordinate system introduced in \refsec{simsim} and shown in \reffig{sim}, the $x$ coordinate of the paddle normal points toward the opponent.  A normal with a positive $x$ coordinate specifies a forehand paddle, while a normal with a negative $x$ coordinate specifies a backhand paddle.  The $\text{IK}_c$ function in \refeq{paddle:ikc} runs the IK optimization starting with either a canonical forehand or backhand rest pose for the robot depending the $x$ coordinate of the requested paddle.  Inverse kinematics is typically implemented as a local optimization process that iteratively uses the robot Jacobian to reduce the distance between the current pose and the requested pose.  So, the solution found by inverse kinematics depends on the initial pose of the robot.  Starting the search with a rest pose leads to an answer that is closer to that initial pose, and therefore likely to be well within the robot's reachable space.

Once the intermediate solution $q^l_t$ is found, forward kinematics is used to compute the corresponding paddle pose for this solution:

\begin{align}
x(p^l_t) \gets \text{FK}(q^l_t),
\end{align}
\\
where FK denotes the forward kinematics function.  Assuming that the requested target paddle position $l(p_t)$ is reachable by the robot, $p^l_t$ should satisfy that position, \ie, one should have:

\begin{align}
l(p^l_t) = l(p_t).
\end{align}

Next, the corresponding paddle normal $N(p^l_t)$ at the intermediate solution is computed.  Then a minimum rotation between $N(p^l_t)$ and the target paddle normal is computed as:

\begin{align}
\phi \gets \angle (N(p^l_t), N(p_t)),
\end{align}
\\
where $\phi$ denotes a 3D rotation that can move $N(p^l_t)$ to $N(p_t)$.  Applying the rotation $\phi$ to the intermediate paddle orientation produces the desired paddle orientation:

\begin{align}
r(p_t) \gets \phi(r(p^l_t)),
\end{align}
\\
where $r(p_t)$ denotes the desired paddle orientation and $r(p^l_t)$ denotes the paddle orientation corresponding to the intermediate paddle pose.

Due to its construction, $r(p_t)$ is guaranteed to have the desired paddle normal $N(p_t)$.  Also, because it is constructed with a minimum rotation from a feasible pose $p^l_t$, it is likely that $r(p_t)$ is feasible by the robot as well.


\subsubsection{Mapping Paddle's Pose to Joint Positions}

Inverse kinematics can be be used to find some joint positions $q_t$ to satisfy the paddle pose $x(p_t)$ subject to the physical limits of the robot and the limits on the range of positions for its joints:

\begin{align}
\label{eq:paddle:ik}
q_t \gets \text{IK}(x(p_t)),
\end{align}
where IK denotes the inverse kinematics function.  In other words, IK maps the desired paddle pose $x(p_t)$ to a robot pose $q_t$.

In general, there are multiple solutions to the IK problem.  The method in this article uses null-space control to prefer robot poses that are closer to some canonical forehand and backhand poses.


\subsubsection{Mapping Paddle's Linear and Angular Velocities to Joint Velocities}

To map the desired linear and angular velocities for the paddle $\dot{x}(p_t)$ to some joint velocities, the end-effector Jacobian is computed at pose $q_t$:

\begin{align}
\label{eq:paddle:jac}
J_t \gets \frac{\partial x(p_t)}{\partial q_t},
\end{align}
\\
where $J_t$ denotes the Jacobian at $q_t$.  \refeq{paddle:jac} can be rewritten as:

\begin{align}
\frac{\partial x(p_t)}{\partial q_t} & = J_t \\
\frac{\partial x(p_t)}{\partial t} \frac{\partial t}{\partial q_t} & = J_t \\
\frac{\partial x(p_t)}{\partial t} & = J_t \frac{\partial q_t}{\partial t} \\
\dot{x}(p_t) & = J_t \dot{q}_t,
\end{align}
\\
where $\dot{x}(p_t)$ and $\dot{q}_t$ denote the time derivatives of $x(p_t)$ and $q_t$.  In other words, the Jacobian establishes a relationship between the paddle's linear and angular velocity and the robot's joint velocities.

In order to solve for $\dot{q}_t$ given $\dot{x}(p_t)$, the Jacobian needs to be inverted.  To handle non-square Jacobians when the robot assembly has more than six joints, and also to avoid failing on singular matrices, the pseudo-inverse method is employed to invert the matrix:

\begin{align}
J^{\dagger}_t \gets \text{pseudo-inverse}(J_t).
\label{eq:paddle:pinv}
\end{align}
\\
Then, the required joint velocities at target can be obtained as:

\begin{align}
\label{eq:paddle:jacinv}
\dot{q}_t \gets J^{\dagger}_t \dot{x}(p_t).
\end{align}

The current joint positions and velocities $q_s, \dot{q}_s$ can be obtained directly by reading them from the robot's controller.  Therefore, the paddle-control policy can analytically obtain the inputs it needs to pass to the trajectory planning skill as described in \refeq{skill:paddle}:

\begin{align}
\pi_p(t, p_t \mid p_s) = \pi_t(t, q_t, \dot{q}_t \mid q_s, \dot{q}_s)
\end{align}


\subsubsection{Trajectory Planning}

At this point the problem of executing the paddle-motion target $p_t$ is reduced to executing target joint states $q_t, \dot{q}_t$ given the initial joint states $q_s, \dot{q}_s$, as described in the trajectory planning skill in \refeq{skill:trajectory}.  This task is accomplished by employing Reflexxes to compute a joint state trajectory between times $s$ and $t$.  Reflexxes is an analytic algorithm which computes the intermediate joints states between times $s$ and $t$ solely based on a set of motion constraints defined on individual joints:

\begin{align}
\begin{split}
\{ (q_j, \dot{q}_j, \ddot{q}_j) \mid s \le j \le t - 1 \} & \gets \text{Reflexxes}(q_s, \dot{q}_s, t, q_t, \dot{q}_t) \\
& \text{subject to:} \\
& q_\text{min} \leq q_j \leq q_\text{max}, \\
& \dot{q}_\text{min} \leq \dot{q}_j \leq \dot{q}_\text{max}, \\
& \ddot{q}_\text{min} \leq \ddot{q}_j \leq \ddot{q}_\text{max}, \\
& \dddot{q}_\text{min} \leq \dddot{q}_j \leq \dddot{q}_\text{max}. \\
\label{eq:reflexxes}
\end{split}
\end{align}

Reflexxes is able to produce trajectories at the desired control frequency, which is \SI{1}{\kilo\hertz} in this implementation.  It is a fast library and can plan for the trajectory and return the next step typically within \SI{1}{\ms}.

Reflexxes can compute trajectories that take the minimum time, or trajectories that complete precisely at some specified time $t$.  The method in this article uses both modes for different skills:

\begin{enumerate}
\item The positioning skill is active when the agent is awaiting the opponent's action.  Its objective is to put the robot in some pose that is suitable for responding to the next incoming ball.  So, for this skill, it is desirable to reach the target position as soon as possible.  When the positioning skill is active, the paddle skill requests a trajectory that reaches the target in minimum time.
\item The objective of the land-ball skill is to hit the ball back at a planned contact time $t$.  When producing trajectories for this skill, the paddle skill is always given a desired target time $t$.  In such cases, usually the robot starts moving slowly and builds up speed toward the target just at the right time to allow it to achieve the desired joint velocities $\dot{q}_t$ exactly when it is reaches the pose specified by the joint positions $q_t$.
\end{enumerate}

It is possible that before the robot reaches the target of the positioning skill, the opponent hits the ball back and the land-ball skill becomes active again.  In that case, the trajectory planned for the positioning skill is not completed, and the trajectory for the land-ball skill starts with the current joint states as its initial condition.

There are situations where no feasible trajectories exist that satisfy the constraints.  For one, the initial joint states at time $s$ or the final joint states at time $t$ might violate the position and velocity constraints of the robot.  In the hierarchical setup, this may happen due to the higher-level skills requesting a paddle-motion target that requires executing joint velocities beyond the limits of the robot.  In such cases, $\dot{q}_t$ already violates the constraints in \refeq{reflexxes}.  Even with conservative limits on the paddle's velocity $\dot{x}(p_t)$, the joint velocities $\dot{q}_t$ may end up being high when the paddle is close to the singularity points of the robot.  In such regions, the inverse Jacobian matrix computed in \refeq{paddle:pinv} contains elements with large magnitudes.  In situations where the requested  final joint positions and velocities $q_t, \dot{q}_t$ are invalid, the analytic controller does not send any commands to the robot.

Another class of infeasible trajectories are those with insufficient time.  For example, if a higher-level skill demands that the paddle moves from one side of the table to the other side in 0.1 seconds, the trajectory would violate the acceleration and jerk limits of a typical robot.  In such cases, Reflexxes can still compute a minimum-time trajectory towards $q_t, \dot{q}_{t}$.  However, due to having insufficient time, at time $t$ the robot will be at some state $\hat{q}_{t}, \hat{\dot{q}}_{t}$ somewhere in between the starting state and the target state.  This state can be queried and used to evaluate the expected outcome of the action under consideration:

\begin{align}
\label{eq:paddle:reflexxes-query}
\{..., (\hat{q}_{t}, \hat{\dot{q}}_{t}, \hat{\ddot{q}}_{t})\} \gets \text{Reflexxes}(q_s, \dot{q}_s, t, q_t, \dot{q}_t).
\end{align}


\subsubsection{Joint-Control}

Once a trajectory is computed, it needs to be executed on the robot.  The joint-trajectory includes joint positions, velocities and accelerations for each timestep.  When the control frequency is high enough, the points on the trajectory are so close to each other that just sending the joint positions $q_{i}, ..., q_{t}$ to a PID controller can control the robot smoothly.  Some robots have PID controllers that also handle velocity targets.  For such controllers, the joint velocities $\dot{q}_{i}, ..., \dot{q}_{t}$ from the trajectory can also be fed to the robot's controller.  Another option for joint-control is inverse dynamics control, where the kinematics and inertial properties of the robot links are used to directly compute the forces or torques for joints.  In either case, the underlying controller that is available for the robot implements the policy

\begin{align}
\begin{split}
\pi_r(u_j \mid & \: q_j, \dot{q}_j, \ddot{q}_j, q_{j + 1}, \dot{q}_{j + 1}, \ddot{q}_{j + 1}), \\
& s \le j \le t - 1,
\end{split}
\end{align}
where $q_j, \dot{q}_j, \ddot{q}_j$ denote the current joint positions, velocities, and accelerations, $q_{j+1}, \dot{q}_{j+1}, \ddot{q}_{j+1}$ denote the desired joint positions, velocities and accelerations at the next timestep, and $u_j$ denotes the joint-control command to execute in order to achieve the desired joint states at the very next timestep.


\subsection{Paddle-Dynamics Model}
\label{sec:paddle:analytic-model}

The objective of the paddle-control skill is to control the robot in a way to achieve the requested paddle-motion target $p_t$ at time $t$.  As outlined in \refsec{paddle:analytic-control}, this skill is implemented with an analytic controller.  However, the solution found by the analytic controller may not always achieve the desired paddle-motion target $p_t$ due to the following reasons:

\begin{enumerate}
\item Failure in inverse kinematics: The desired paddle-motion target $p_t$ may not be physically feasible for the robot.  The target paddle position may be out of reach for the robot, or it may have an orientation that is not possible for the robot's anatomy.
\item Failure in trajectory planning: The desired target time $t$ may be too close.  In that case, the trajectory planning skill cannot move the robot to the target in time without violating the specified motion constraints.
\end{enumerate}

The analytic controller can predict the error in achieving the paddle-motion target due to either of the above two causes.  If there is enough time for the trajectory to reach the target while satisfying the motion constraints, then the trajectory would be feasible.  As shown in \refeq{paddle:reflexxes-query}, the final point on the trajectory computed by Reflexxes contains information about where the robot will be at target time $t$.  When the trajectory is feasible by the given time $t$, the final joint positions and velocities $\hat{\dot{q}}_{t}, \hat{\dot{q}}_{t}$ will be equal to the planned target $q_t, \dot{q}_t$.  In other words, the following holds:

\begin{align}
\hat{q}_{t} = & q_{t}, \\
\hat{\dot{q}}_{t} = & \dot{q}_{t}.
\end{align}

When the time given is not enough to complete the trajectory, $\hat{q}_{t}, \hat{\dot{q}}_{t}$ can be used to predict the final paddle-motion state at time $t$.  First, forward kinematics is used to predict the paddle pose from executing $\hat{q}_t$:

\begin{align}
x(\hat{p}_t) \gets \text{FK}(\hat{q}_t)
\label{eq:paddle:fk}
\end{align}
\\
where $x(\hat{p}_t)$ denotes the predicted resulting paddle pose.

Then, the end-effector Jacobian is used to produce a prediction for the paddle's final linear and angular velocities given the predicted final joint velocities $\hat{\dot{q}}_{t}$: 

\begin{align}
\dot{x}(\hat{p}_t) \gets J(\hat{q}_{t}) \hat{\dot{q}}_{t}.
\label{eq:paddle:fkjac}
\end{align}

Equations ~\ref{eq:paddle:fk},~\ref{eq:paddle:fkjac} combined specify a prediction for the full paddle-motion state at time $t$:

\begin{align}
\hat{p}_t = x(\hat{p}_t), \dot{x}(\hat{p}_t).
\label{eq:paddle:phat}
\end{align}

In \refsec{paddle:analytic-control}, inverse kinematics (\refeq{paddle:ik}), the robot Jacobian (\refeq{paddle:jacinv}), and Reflexxes (\refeq{paddle:reflexxes-query}) were used to map the paddle-motion target $p_t$ to joint positions and velocities $\hat{q}_{t}, \hat{\dot{q}}_{t}$.  In this section, forward kinematics (\refeq{paddle:fk}) and the robot Jacobian (\refeq{paddle:fkjac}) are used to map $\hat{q}_{t}, \hat{\dot{q}}_{t}$ to a predicted paddle-motion state $\hat{p}_t$ resulting from running the analytic controller.  Combining the above equations permits defining a forward paddle-control model, which given the current paddle-motion state $p_s$ and a paddle-motion target $p_t$ produces a prediction for the paddle-motion state resulting from running the analytic controller: 

\begin{align}
\label{eq:paddle:fwdmodel}
\hat{p}_t \gets P(p_t, p_s)
\end{align}
\\
where $P$ is the forward paddle-control model under the motion constraints specified in \refeq{reflexxes}.

The prediction $\hat{p}_t$ shows the expected state of the paddle at the planned contact time with the ball.  This prediction can be used in conjunction with the forward landing model from \refsec{dyn:fwdlanding} to inform the agent about the expected landing location and velocity resulting from executing the paddle action $p_t$.

The controller discussed in \refsec{paddle:analytic-control} and the paddle-control model discussed in this section are entirely analytic.  They are derived mathematically based on the kinematics of the robot links and the motion constraints on the robot motors.  This approach increases sample efficiency since no training episodes are being spent on learning robot-control.  Moreover, the abstract control space exposed by the analytic controller makes the remaining parts of the problem easier to learn with reinforcement learning.


\subsubsection{Learning Paddle-Dynamics}

One of the advantages of using a trajectory planning module like Reflexxes is that it generates smooth targets which already account for the physical motion limits of the robot.   This smoothness and continuity in the targets can hide away some imperfections in the underlying robot controller.  For example, if the PID gains are not tuned very well, they would cause smaller errors or oscillations when the targets are closer to each other.

However, robot controllers and robots are ultimately imperfect and imprecise.  There are various causes for deviations to exist between the expected behavior and observed behavior of a robot.  Such deviations exist due to:

\begin{itemize}
\item{Imperfections in the controller's implementation, gains, and other parameters.}
\item{Round-trip delay of executing commands on the robot.}
\item{Malfunctioning or worn out robot parts.}
\item{Mechanical backlash caused by gaps between components.}
\item{Misspecified motion constraints.  If the velocity, acceleration, and jerk limits given to Reflexxes are higher than the actual limits of the robot, Reflexxes would compute trajectories that are beyond the physical limits of the robot.}
\end{itemize}

It is possible to extend the notion of the robot's model to also capture such imprecisions in control of the robot.  Unlike the analytic paddle-control model in \refsec{paddle:analytic-model}, which was derived mathematically, it is best to learn a dynamics model over robot's behavior by experimenting and observing outcomes.  A neural network can be trained to predict inaccuracies in robot-control regardless of the underlying cause.  As the robot is executing the target joint-motion states specified by $q_t, \dot{q}_t$, the resulting joint positions and velocities at time $t$ can be recorded and used as labels for training a model.  The trained model can then make predictions of the form:

\begin{align}
\doublehat{q}_t, \doublehat{\dot{q}}_t \gets R(q_s, \dot{q}_s, q_{t}, \dot{q}_{t})
\label{eq:paddle:learned-model}
\end{align}
\\
where $R$ denotes the forward robot-control model, and $\doublehat{q}_t, \doublehat{\dot{q}}_t$ denote the expected joint position and velocity observations at time $t$.

This constitutes a forward prediction, which when combined with \refeq{paddle:fwdmodel} can produce a more accurate prediction about the future state of the paddle.  Such a forward prediction can be used to produce a more accurate estimate on the landing location of a particular strike specified by $p_t$.

In the other direction, the same training data can be used to learn a corrective robot-control model as in:

\begin{align}
{q_t}', {\dot{q}_t}' \gets R^{-1}(q_s, \dot{q}_s, q_{t}, \dot{q}_{t}),
\end{align}
\\
where $R^{-1}$ denotes the inverse robot-control model, and ${{q}_t}', {\dot{q}_t}'$ are alternative targets such that if they are requested from the robot, it is expected that the observed joint states at time $t$ would be close to the actual targets $q_t, \dot{q}_t$.  In other words, it is expected that:

\begin{align}
q_{t}, \dot{q}_{t} \approx R(q_s, \dot{q}_s, {{q}_t}', {\dot{q}_t}').
\end{align}

The inverse robot-control model can be used to adjust the joint position and targets before they are passed to the trajectory planning skill to increase the likelihood that the requested paddle-motion target $p_t$ is going to be achieved.


\subsection{Learning Paddle-Control}
\label{sec:paddle:learning}

The primary approach in this article uses the analytic paddle controller discussed in \refsec{paddle:analytic-control}.  This section discusses an alternative implementation for the paddle-control skill using learning.  The learning approach is undesirable when an analytic solution exists.  The learning approach discussed here is implemented and evaluated in an ablation experiment in \refsec{striking}.

As shown in the task hierarchy from \refsec{skill}, the paddle-control skill depends on the trajectory planning skill, which in turn depends on the joint-control skill.  However, it is possible to treat paddle-control as a single problem.  Combining the task specifications for these three skills from \refeq{skill:paddle}, \refeq{skill:trajectory}, and \refeq{skill:joint} produces a contracted definition for the paddle-control skill as:

\begin{align}
\label{eq:paddle:fused2}
\pi_p(t, p_t \mid p_s) = & \{u_j \mid s \le j \le t - 1\} \\
& \text{subject to motion constraints in \refeq{skill:trajectory}} \nonumber.
\end{align}
\\
\refeq{paddle:fused2} relates the high-level paddle-control policy with $p_t$ as target to the low-level joint-control actions $u_j$ over multiple timesteps from $s$ to $t - 1$.

Given this formulation, the paddle-control task can be treated as a learning problem where the objective is to find the right joint commands to bring the paddle to the desired state at the desired time.  A basic approach to learning paddle-control with standard reinforcement learning may use random actions on joints with the intention of gradually discovering the right actions that move the paddle toward the requested targets.  This approach is not desirable, since it is very inefficient and random joint actions can break the robot.

An alternative approach may consider action spaces that span over time intervals longer than one timestep.  Effective strikes generally have some continuity in joint motions.  An effective strike usually maintains the direction of motion for most joints.  So, one can sample a set of fixed velocity/acceleration profiles for each joint and use those to create paddle strikes.  However, it is hard to determine whether such motion profiles would cover the space of all useful strikes.  Another problem is the issue of the initial pose of the robot at the start of the strike.  If the paddle is already in front of the robot, a successful strike may need to bring the paddle back first, before it can be moved forward again to hit the ball with enough force.  Moreover, the robot may be in the middle of some motion and have non-zero velocities on some joints when a new strike is to be executed.  These requirements greatly increases the space of possible motions that need to be tried in order to develop a high-level paddle controller via training.


\subsection{Positioning Policy}
\label{sec:paddle:positioning}

The positioning skill is active when the agent is awaiting the opponent's action, \ie, when the opponent is striking the ball.  This skill is invoked either if the episode starts with the launcher sending the ball toward the opponent, or right after the agent hits the ball back toward the opponent.  The skill stays active until the opponent makes contact with the ball, at which point the striking skill becomes active.

The positioning skill receives a paddle position target $l(p)$ from the strategy skill.  The objective of the skill is to move the paddle to the requested position as quickly as possible.  Note that the requested paddle target $p$ has no specified time.

The paddle position is a proxy to the robot's pose.  Specifying the paddle pose instead of the robot pose has the advantage that it makes the policy less dependent on the specific robot that is in use.  Moreover, a paddle position can be specified with three values, while the full robot pose typically requires specifying six or more joint positions.

As discussed in \refsec{skill:positioning}, the positioning skill $\pi_w$ is defined by the policy:

\begin{align}
\pi_w(l(p), \sgn N_x(p)) = \pi_p(p \mid p_s),
\end{align}
\\
where $\pi_p$ denotes the paddle-control policy, $p$ denotes some paddle-motion target that satisfies the requested paddle position $l(p)$ and normal direction indicated by $\sgn N_x(p)$.  The paddle skill is expected to achieve paddle-motion target $p$ as fast as possible.

The positioning skill is implemented analytically.  It simply computes a fully-specified paddle-motion target $p$ which satisfies the paddle position $l(p)$ and forehand/backhand pose indicated by $\sgn N_x(p)$ and requests $p$ from the paddle-control skill, which in turn executes it using the trajectory planning skill.  To compute $p$, the positioning skill first uses inverse kinematics to compute a robot pose as in:

\begin{align}
q^w \gets \text{IK}_c(l(p)),
\end{align}
\\
where $\text{IK}_c$ denotes the inverse kinematics function starting with some canonical robot rest pose, and $q^w$ denotes the joint positions that satisfy the requested paddle position $l(p)$.  Satisfying $\sgn N_x(p)$ is achieved by starting the IK search with a canonical pose which has the same forehand or backhand direction as specified by $\sgn N_x(p)$.  Since IK algorithms perform a local search, a carefully-chosen canonical forehand and backhand pose allow the positioning policy to satisfy $l(p)$ without flipping the forehand/backhand orientation.  Once target joint positions $q^w$ are computed, forward kinematics is used to compute a paddle pose from them:
\\
\begin{align}
x(p) \gets \text{FK}(q^w).
\end{align}

Assuming that the requested target paddle position $l(p)$ is reachable by the robot, the paddle pose $x(p)$ should satisfy that position.  Otherwise, $x(p)$ will get as close as possible to $l(p)$.  The fully-specified paddle-motion target should also include the paddle's linear and angular velocities.  In the current implementation, the positioning skill requests a stationary target by setting the target velocities to zero:

\begin{align}
p \gets x(p), 0.
\end{align}

However, a more complex implementation could request to have non-zero joint velocities at target to reduce the expected reaction time to the next ball coming from the opponent.  The full paddle-motion target $p$ is then sent to the paddle-control policy to execute.

\subsection{Conclusion}

This section discussed the analytic controller setup that handles paddle-control, trajectory planning, and joint-control tasks.  It also discussed the analytic paddle-dynamics model $P$ which predicts the expected paddle-motion state $\hat{p}_t$ resulting from executing a desired paddle-motion target $p_t$.  The next section explains how the analytic paddle controller and model can be used in conjunction with the dynamics models trained from human data over the environment's game space to implement the land-ball policy.

\section{Striking Policies}
\label{sec:striking}

The dynamics models trained from human demonstrations allow the agent to predict the trajectory of an incoming ball, and predict the landing locations resulting from hitting the incoming ball with given paddle strikes.  This section describes how the model-based land-ball policy uses these dynamics models and the paddle-control policy described in the previous section to execute arbitrary landing targets.  The land-ball policy is evaluated on a target practice task with random landing targets.  In order to determine whether using arbitrary strikes from human demonstrations imposes a restriction on the land-ball policy, the policy is also evaluated on dynamics models trained from data generated directly on the robot.  To evaluate the sample-efficiency of the model-based land-ball policy, the land-ball task is also learned directly using a model-free reinforcement algorithm.  Lastly, the alternative hit-ball striking policy is described.  The hit-ball policy does not use the strikes demonstrated by the humans and is suitable for learning new striking motions.


\subsection{Model-Based Land-Ball Policy}
\label{sec:landball:modelbased}

The objective of the land-ball skill is to execute a paddle strike to send an incoming ball with motion state $p_s$ to a landing target $g$ consisting of a target position and speed at the moment the ball lands on the opponent's side of the table.  This section describes two implementations for the land-ball policy and evaluates both.

\subsubsection{Policy Implementation}

\begin{figure}[htb!]
\centering
\includegraphics[width=1.0\columnwidth]{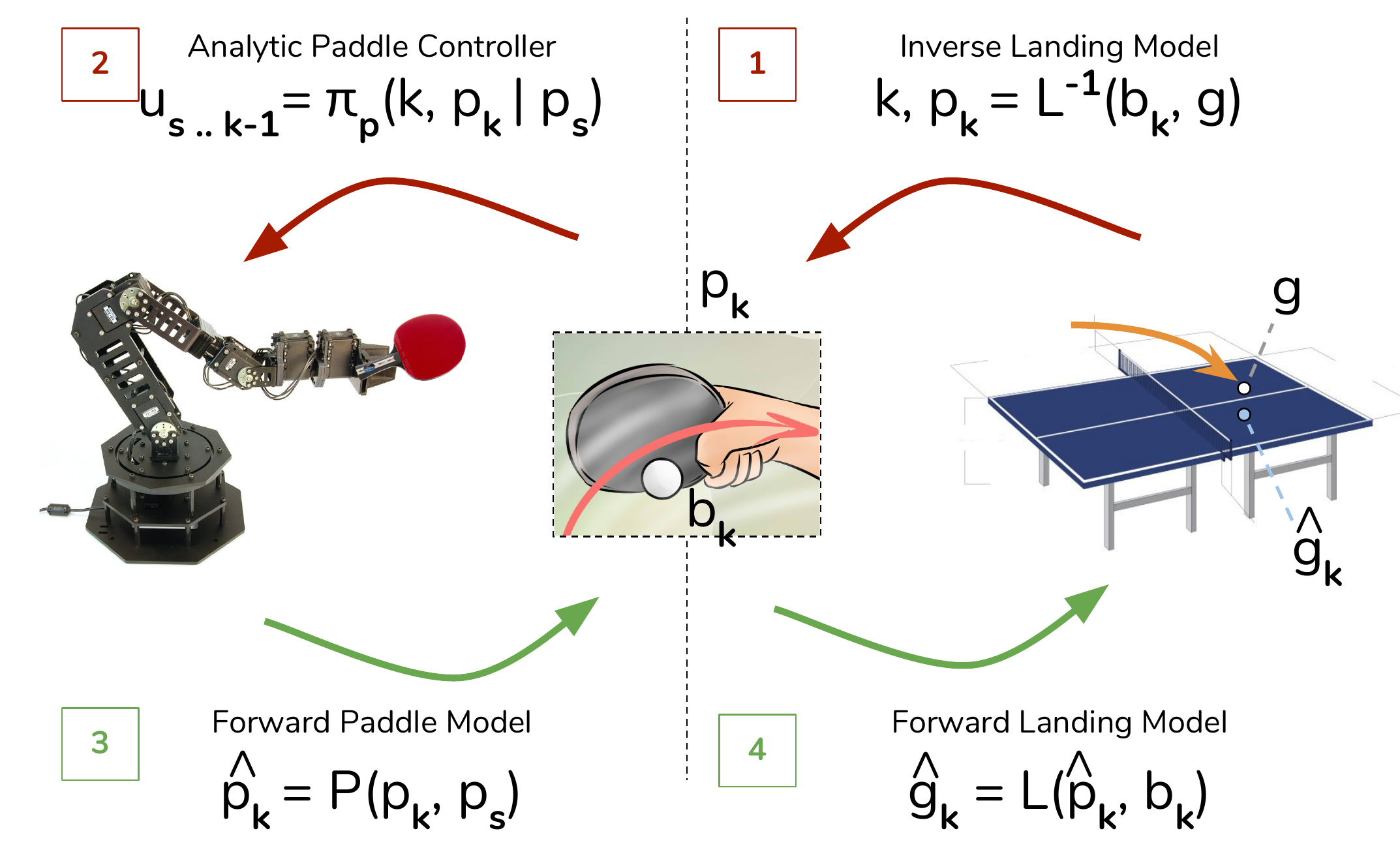}
\mycaption{Model-Based Land-Ball Policy.}{The land-ball policy uses three dynamics models learned in the game space and an analytic paddle model derived in the robot space to execute land-ball targets specifying a desired position and speed for the ball at the moment of landing.  The policy uses the ball-trajectory prediction model (not visualized) to predict the ball's trajectory.  For any candidate pre-contact point in the trajectory and the corresponding ball-motion state $b_k$, it uses the inverse landing model to compute a pre-contact paddle-motion target $p_k$.  For each $p_k$, the analytic paddle controller computes a joint-trajectory and joint actions to achieve the target $p_k$, as well as a prediction $\hat{p}_k$ on the expected paddle-motion state resulting from executing the target $p_k$.  Given $\hat{p}_k$ and $b_k$, the forward landing model can predict $\hat{g}_k$, the landing position and speed resulting from executing $p_k$.  The dynamics models permit implementing the land-ball policy with dynamics models trained from human demonstrations.  This policy can be deployed on any robot assembly to hit land-ball targets without prior training.}
\label{fig:landball:policy}
\end{figure}

\reffig{landball:policy} illustrates the implementation of the model-based land-ball policy using four dynamics models: ball-trajectory prediction, forward landing-prediction, inverse landing-prediction, and the analytic paddle-control model.  Algorithm~\ref{alg:landball:policy} outlines the policy steps in detail.

\begin{algorithm}[htb!]
  \SetKwInOut{Input}{input}
  \SetKwInOut{Output}{output}
  \Input{Current ball-motion state $b_s$}
  \Input{Desired landing target $g$}
  $T = b_{s+1}, b_{s+2}, \dots, b_n \gets B(b_s)$\;
  \ForEach{$b_k \in T$ such that $b_k$ is reachable}{%
    $p_k \gets L^{-1}(b_k, g)$\;
    $\hat{p}_k \gets P(p_k, p_s)$\;
    $\hat{g}_k \gets L(\hat{p}_k, b_k)$\;
  }
  $t \gets \argmin_k ||\hat{g}_k - g||$\;
  $i \gets s$\;
  \Repeat{robot paddle hits the ball or episode ends}{
    emit next action $u_i$ from $\pi_p(t, p_t \mid p_s)$\;
    $i \gets i + 1$\;
  }
  \caption{Model-Based Land-Ball Policy Algorithm}
  \label{alg:landball:policy}
\end{algorithm}

\begin{figure}[htb!]
\centering
\includegraphics[width=1.0\columnwidth]{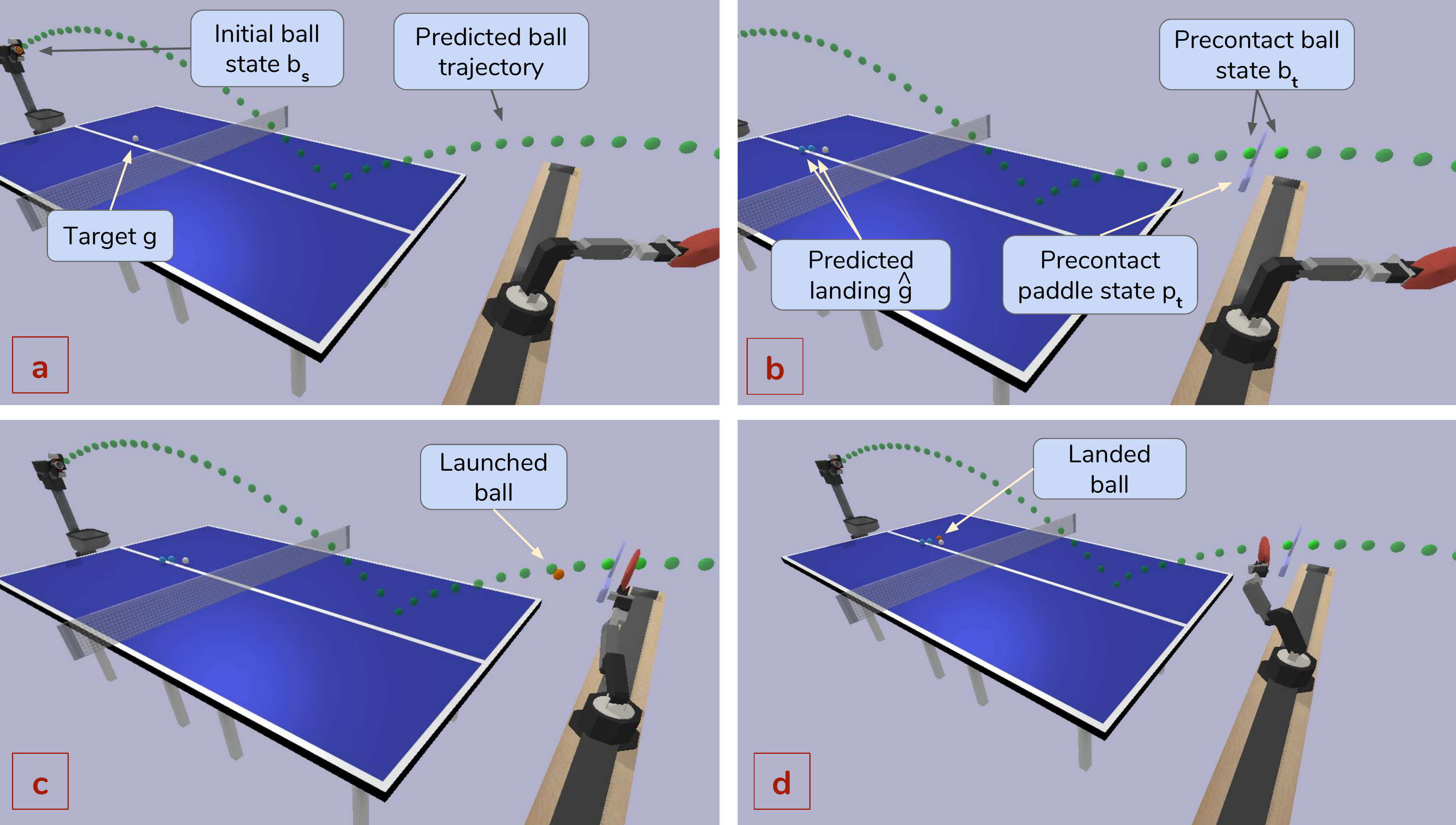}
\mycaption{Demonstration of the Land-Ball Policy.}{\textbf{a)} A target $g$ is passed as input to the land-ball skill.  Given the estimate on the current ball-motion state $b_s$ (visualized by the orange ball), the future ball trajectory (visualized as a sequence of green balls) is predicted.  \textbf{b)} Two points in the predicted ball trajectory are selected as pre-contact points (visualized by light-green balls).  For each such point, a pre-contact paddle-motion target is computed.  Only one of the paddle-motion targets is visualized by a transparent paddle.  Given each paddle-motion target and its corresponding ball-motion state, the forward landing model is used to predict the landing position and speed of the ball.  The paddle solution $p_t$ whose predicted landing $\hat{g}$ is closer to the requested target $g$ is selected.  \textbf{c)} The paddle-motion target is sent to the analytic paddle controller, which executes it.  \textbf{d)} The ball lands on the table.  Since neither the inverse landing nor the forward landing models are completely accurate, the ball may land at a location that is close to, but not exactly on either the requested target $g$, or the predicted landing target $\hat{g}$.}
\label{fig:landball:nocem}
\end{figure}

\reffig{landball:nocem} demonstrates the different stages of the land-ball policy's algorithm.  Given the incoming ball's motion state $p_s$, the policy predicts the ball's future trajectory $T$.  The predicted ball trajectory contains future ball position and velocity observations at the resolution of the environment timestep (20 ms).  There are multiple options for selecting a subset of the predicted trajectory $T$ as potential striking targets.  In the current implementation, a heuristic is used to select all points that lie between the two planes $x = -1.8$~m, $x = -1.6$~m, corresponding to a 20~cm band in front of the robot assembly.  This band typically contains 2-3 predicted observations in the predicted ball trajectory.  These points are highlighted as light-green balls in \reffig{landball:nocem}.  Considering multiple potential striking points allows the land-ball policy to come up with multiple striking solutions and pick one that is most likely to satisfy the requested landing target $g$.  Another potential solution for selecting striking targets from the trajectory is to use a heuristic to prefer balls that are closer to a certain preferred height for the robot.  It is also possible to leave this decision entirely up to a higher-level skill like the strategy skill.  In other words, the strategy skill could additionally specify the striking target by requesting a desired height or distance from the net for the point of strike.

For each potential striking point $b_k$ corresponding to time $k$, the policy uses the inverse landing-prediction model to compute a target paddle-motion $p_k$ to land the ball at target $g$.  To ensure contact with the ball, $l(p_k)$ is always set to $l(b_k)$.  Since the landing-prediction models are trained from human demonstrations, querying the inverse landing-prediction model amounts to computing the expected mean striking motion as demonstrated by humans when aiming for a target $g$.  For each $p_k$, the forward paddle model is used to compute a prediction for the expected paddle-motion state $\hat{p}_k$ that would result from executing $p_k$ with the analytic paddle controller.  If the pose specified by $p_k$ is not feasible for the robot, or if the target motion state is not reachable in time given the kinematic motion constraints, $\hat{p}_k$ informs the agent about how close the paddle can come to the requested state $p_k$.  In addition, $\hat{p}_k$ captures any imprecisions resulting from the iterative IK search process.  Given each predicted $\hat{p}_k$, the policy uses the forward landing-prediction model to compute $\hat{g}_k$, the expected landing position and velocity resulting from hitting the ball $b_k$ with the paddle strike specified by $\hat{p}_k$.  The policy then picks the strike $p_k$ such that it has the smallest predicted landing error and executes it using the analytic paddle-control policy as $\pi_p(t, p_t \mid p_s)$.

\subsubsection{Automatic Forehand/Backhand}

Any given strike can be executed with forehand or backhand paddle motions.  As described in \refsec{dyn:norm}, inverting the paddle normal (from forehand to backhand or vice versa) does not change its impact on the ball.  The normalization process used in training the dynamics models replaces all backhand strikes from human demonstrations with forehand strikes.  So, the inverse landing-prediction model always suggests forehand paddle strikes.  However, for some targets backhand strikes might be better choices.  For example, if the predicted striking point $b_k$ is close to the left side (from the robot's perspective) of the robot's reachable space, $p_k$ may only be feasible with a backhand orientation.  Also, if the current paddle state $p_s$ at the beginning of the strike is a backhand pose, the target $p_k$ may only be reachable in time with a backhand target; switching from backhand to forehand may require more time.  For these reasons, the land-ball policy considers both forehand and backhand executions for each $p_k$ and picks the one whose $\hat{p}_k$ is closer to $p_k$.

\subsubsection{Improved Policy with Cross-Entropy Method (CEM)}

The landing target resulting from a given pre-contact paddle-motion and ball-motion states has a unimodal distribution, \ie there is only one possible expected outcome.  The inverse landing problem, however, may have multiple solutions.  The inverse landing-prediction model captures the mean of all paddle-motion states that can send the ball to the target $g$.  However, the mean action may be skewed and it may not do well.

\begin{figure}[htb!]
\centering
\includegraphics[width=1.0\columnwidth]{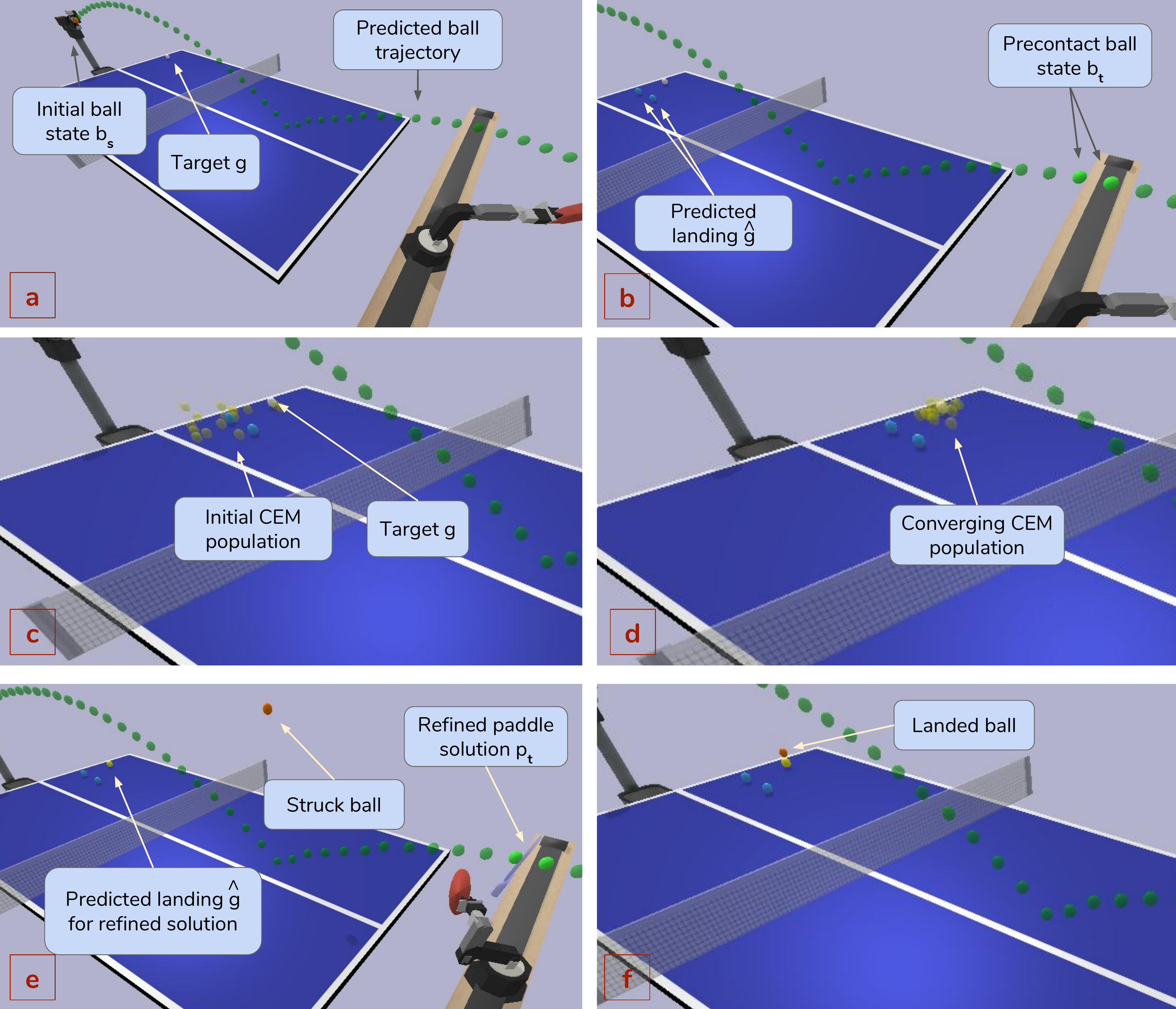}
\mycaption{Demonstration of the Land-Ball Policy with CEM Search.}{\textbf{a)} See \reffig{landball:nocem}.a. \textbf{b)} See \reffig{landball:nocem}.b. \textbf{c)} Given the best candidate paddle-motion target $p_t$ computed by the inverse landing model, a population of paddle-motion targets is created randomly around $p_t$.  The translucent yellow balls visualize the predicted landing targets for each paddle-motion target in the CEM population.  \textbf{d)} During each iteration of CEM, the population may get closer to the target $g$. \textbf{e)} Once the CEM population has converged, the final mean of the population of paddle-motion targets is passed as an action to the paddle-control skill.  \textbf{f)} The ball lands on the table.}
\label{fig:landball:cem}
\end{figure}

A variant of the model-based land-ball policy uses an iterative refinement process based on CEM~\cite{rubinstein1999cross} to improve the final solution $p_k$ returned from Algorithm~\ref{alg:landball:policy}.  The refinement process performs a search in the space of paddle-motion states around $p_k$.  \reffig{landball:cem} demonstrates the different stages of the land-ball policy with a forward CEM search.  A population of solutions is created with the mean of $p_k$ and an initial standard deviation for each dimension in $p_k$.  During each iteration of the algorithm, each individual in the population is evaluated using the forward landing-prediction model and its predicted target error is recorded.  The predicted errors are used by the CEM algorithm to compute an updated mean and standard deviation for the next iteration.  The process continues until either the maximum number of iterations is reached, or the predicted error falls below a fixed threshold.  At that point the final mean value is returned as the target paddle-motion specifying the strike.

\subsubsection{Evaluation}
\label{sec:landball:eval}

The land-ball policy is evaluated by launching random balls from the ball launcher and requesting the policy to land the ball at random targets.  To allow for a fair comparison of the alternative striking policies discussed later in this section, in these experiments the forward and inverse landing-prediction models were trained only with position targets (without velocity).  The dynamics models were trained both from a small dataset containing 7,000 trajectories, and a large dataset containing 140,000 trajectories to evaluate the impact of the number of samples on the accuracy of policies.

The robot always starts at a fixed stationary forehand pose with zero joint velocities.  The ball is launched towards the robot from a random position and with random velocity.  If the launched ball goes out or hits the net, the launch is repeated.  In the simulation environment, the table is at the center of the coordinate system with its center point at $(0, 0, 0.76)$\,m.  The $x$ coordinates increase away from the robot and towards the opponent, the $y$ coordinates increase to the left side of the net, and the $z$-axis points up.  The robot's base is at $(-1.8, 0, 0.76)$\,m, \ie 1.8\,m away from the net.   The initial ball position is chosen uniformly at random from a cube defined by $l_\text{low}(b_s) = (1.4, -0.3, 0.86)$\,m, $l_\text{high}(b_s) = (2.0, 0.3, 1.26)$\,m.  The initial ball velocity is chosen uniformly at random between $v_\text{low}(b_s) = (-6, -0.5, 1.5)$\,m/s and $l_\text{high}(b_s) = (-5, 0.5, 2.5)$\,m/s.  The landing target for the ball is chosen uniformly at random from a box $0.4$\,m away from the net and $0.1$\,m away from the table edges.

\reftab{landball:modelbased} shows the evaluation results for the two model-based land-ball policy variants with and without CEM search.  A successful ball return requires that the robot strikes the ball over the net and onto the opponent's side of the table.  The mean target errors are computed only over the balls that are successfully returned.



\begin{table}[htb!]
\centering
  \begin{tabular}{|l|c|c|c|r|r|}
  \hline 
  \thead{Method} & \thead{Search} & \thead{Data} & \thead{Samples} & \thead{Return Rate} & \thead{Target Err} \\
  \hline 
  \hline 
  Model-Based & - & VR & 7,000 & 88.0\% & 0.216 m \\
  Model-Based & CEM & VR & 7,000 & 86.7\% & 0.191 m \\
  \hline 
  Model-Based & - & VR & 140,000 & 89.6\% & 0.182 m \\
  Model-Based & CEM & VR & 140,000 & 90.8\% & 0.119 m \\
  \hline 
  \end{tabular}
\mycaption{Evaluation of Model-Based Land-Ball Policies Trained with Human Demonstrations Collected in the VR Environment.}{Mean ball return rate and mean landing target error computed over 1200 attempts with random targets.  The policy uses dynamics models trained from data collected in the VR environment.  The two variants of the policy (with and without CEM search) have similar return rates.  When trained on the large dataset, the CEM variant has a higher target accuracy with an average error of about 12 cm, while the variant without CEM has an average error above 18 cm.  The policy is sample-efficient, since even when trained on the small dataset it achieves an average target position error of about 20 cm.}
\label{tab:landball:modelbased}
\end{table}

Visualizing the policy reveals that a high percentage of the failure cases happen because the ball hits the net.  The data normalization process described in \refsec{dyn:norm} normalizes all landing trajectories such that the $x, y$ coordinates of the pre-contact ball state are set to zero.  In doing so, the normalization process hides the position of the net from the model.  The forward and inverse landing-prediction models operate as if the net did not exist.  It is possible to add extra inputs to the landing models to specify a distance to the net.  However, doing so would increase the number of dimensions, which might cancel some of the benefits of the data normalization process.  This approach was not tested in experiments.  In the hierarchical policy design, it is expected that the high-level strategy skill will account for the inefficiencies in the striking skills and will pick high-level actions that work around them.

\begin{figure}[htb!]
\centering
\includegraphics[width=0.9\columnwidth]{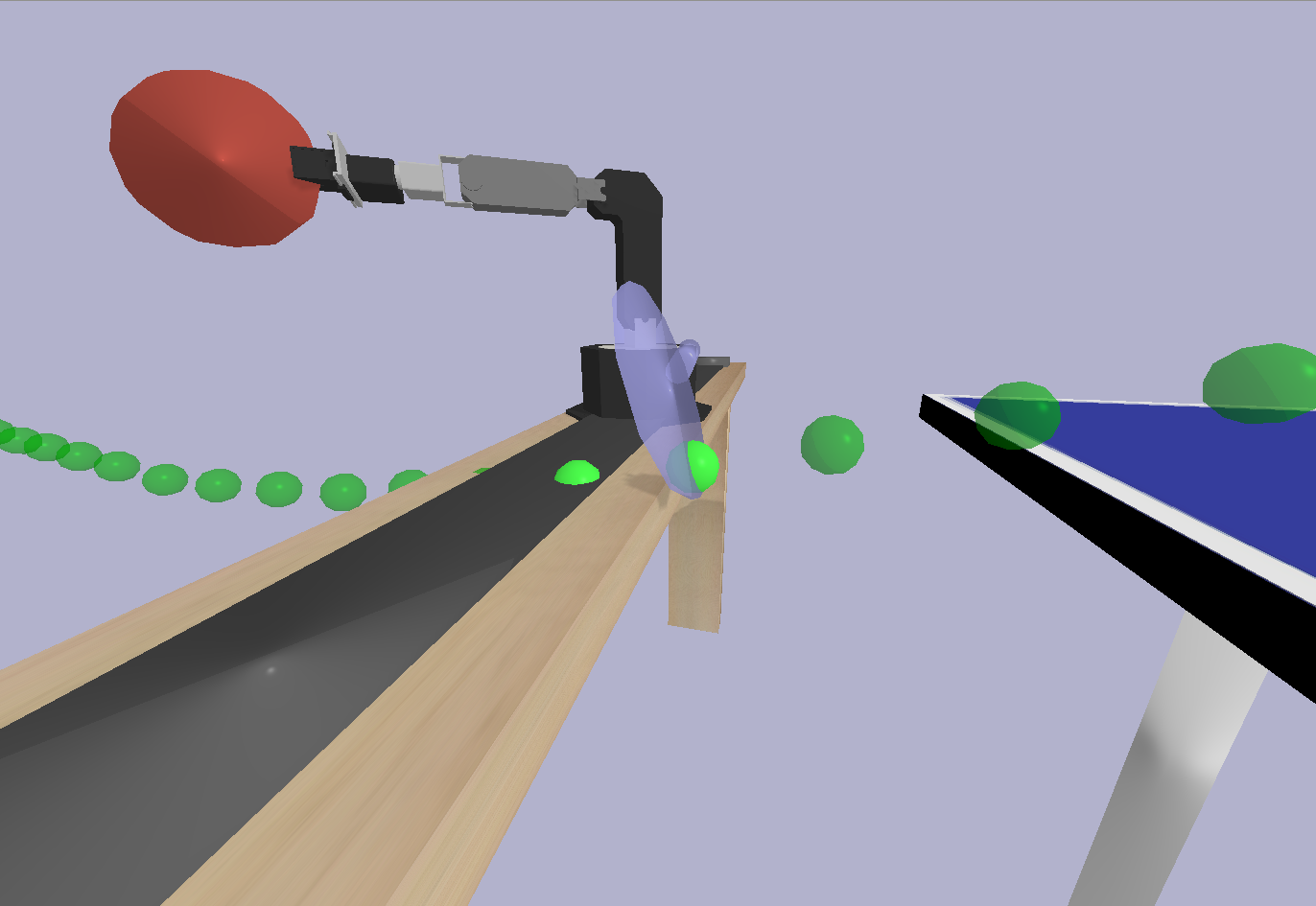}
\mycaption{Example Failure Case for Model-Based Land-Ball Policy.}{At times, the trajectory of the launcher ball collides with the table under the robot assembly.  The translucent paddle visualizes the requested paddle-motion target.  Such balls are more difficult to return, as the planned paddle trajectory may collide with the table.}
\label{fig:elandball:failure}
\end{figure}

Another failure happens when the launched ball goes straight to the table under the robot.  An example of this failure case is shown in \reffig{elandball:failure}.  In these cases, the robot is not able to execute its desired paddle strike because the paddle would collide with the table.

The land-ball skill provides a high-level abstraction over the agent's game-play during a single exchange with the opponent.  This high-level abstraction does not reduce generality of behavior.  Barring deceptive movements to hide the agent's intention from the opponent, any sequence of paddle actions can be reduced to the resulting landing state for the ball.  In other words, the land-ball skill makes it possible to specify the agent's behavior by its desired outcome.  Learning to use the land-ball skill is easier for the agents as its action space has fewer dimensions than a fully-specified target paddle-motion, yet its action space can specify complex behaviors.


\subsection{Model-Based Land-Ball Trained with Robot Data}

The analytic paddle controller executes strikes by executing paddle-motion targets at the moment of contact with the ball.  These paddle-motion targets are learned from humans who can freely move the paddle around using the high degrees of freedom in their shoulder and wrist joints.  In contrast, robots have few degrees of freedom.  In particular, the main robot used in these evaluations has only six.  This observation raises the question whether requiring the robot to imitate human strikes imposes a restriction on the policy, as either the anatomy of the robot or the analytic paddle controller may not permit replicating the paddle motions demonstrated by humans.  To answer this question, the model-based land-ball policies were also trained from strikes generated directly on the robot itself.

\subsubsection{Data Generation}

Generating successful strikes with the robot requires a functioning striking policy.  However, such a policy does not exist at the beginning and it is not straight-forward to implement a hand-coded policy that can land the ball well, and at the same time demonstrate versatile strikes.  As discussed in \refsec{paddle:learning}, generating random yet successful strikes is a hard problem.  In this experiment, random strikes are generated by sampling fixed target velocities for each of six joints on the robot and passing them to the PID controller.  The robot is always initialized to a fixed forehand pose at the beginning of the strike.  The velocity targets are kept fixed during each strike.  The range of velocity values is determined using a trial-and-error process to increase the likelihood of useful strikes.

A successful strike requires that striking paddle makes contact with the ball.  Since the paddle strike is random, a cooperative ball launcher is used to launch the ball with the right velocity to meet a swinging paddle at a desired time.  The position of the ball launcher is fixed, but the velocity of the launched balls is variable.  Such a ball launcher has a real-world counterpart as well.  Most commercial ball launchers have a user interface allowing the player to specify the desired velocity and spin of the ball.  Implementing a cooperative launcher requires interfacing with the device and programmatically setting the desired launch attributes for the ball.  To achieve contact between the ball and the paddle, this experiment trains and uses two additional dynamics models:

\begin{enumerate}
\item \textbf{Forward low-level paddle-control model}: Similar to what is described in \refsec{paddle:learning}, a model is learned to predict the trajectory of the paddle-motion states resulting from executing a set of fixed joint velocities.  The model is implemented as a recurrent neural network, which given the velocity targets, predicts the full paddle-motion state over subsequent timesteps assuming that the robot always starts from a fixed pose.
\item \textbf{Inverse ball-launcher model}: The cooperative launcher uses the inverse ball-launcher model.  The inputs to this model are a desired ball position observation in the future and a corresponding time.  The outputs are the 3D launch velocity vector, such that launching the ball from the fixed launcher position with that velocity will result in a ball that bounces on the table and reaches the desired position at the desired time.  The inverse ball-launcher model is trained from  trajectories of randomly-launched ball.
\end{enumerate}

\subsubsection{Evaluation}
\label{sec:landball:roboteval}

Once the two above dynamics models are trained, they are used to generate random strikes and collect landing trajectories.  For each sample, a set of fixed velocities are sampled for the robot joints.  The forward low-level paddle-control model is used to predict the trajectory of the paddle.  A random paddle position that lies within the same 20~cm band described in \refsec{landball:modelbased} is picked from the trajectory.  Then, the inverse ball-launcher model is used to compute a velocity for the launcher ball such that the launched ball reaches that paddle position at the same time as the paddle.  This process is likely to create contact between the ball and the paddle.  Moreover, the range of random joint velocity targets are hand-tuned in a way that some of the returned balls reach the opponent's side of the table.

\begin{figure}[htb!]
\centering
\includegraphics[width=0.9\columnwidth]{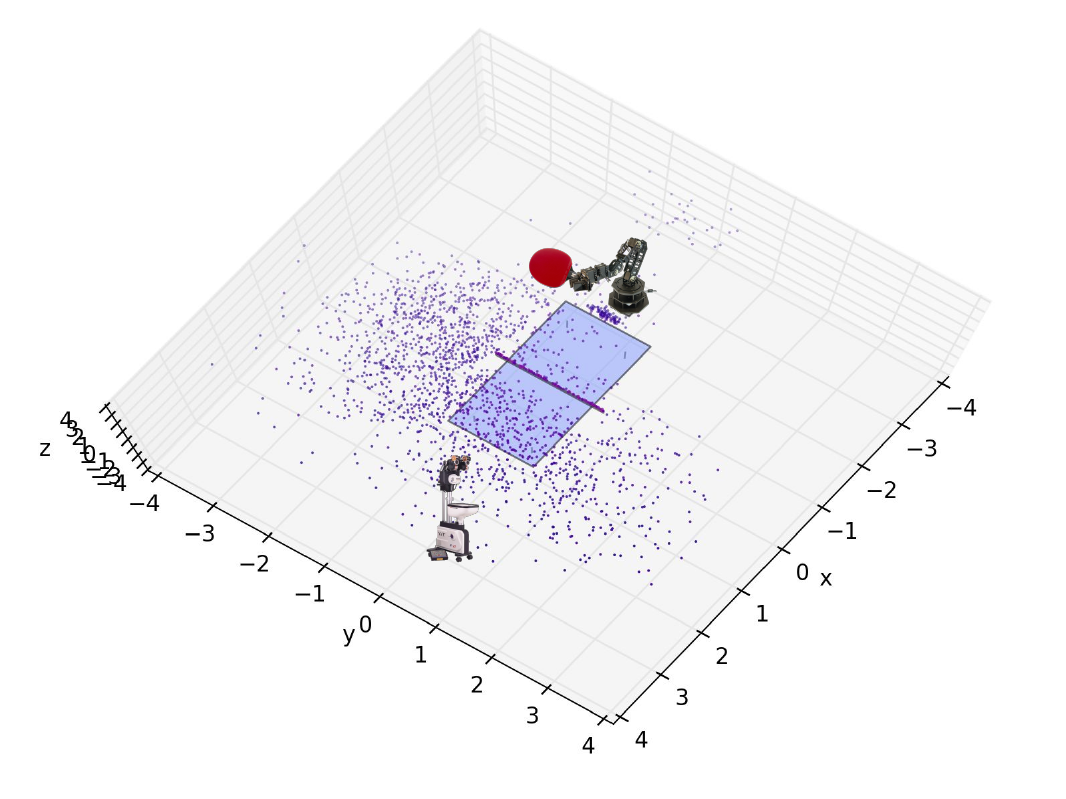}
\mycaption{Distribution of Landing Positions in Landing Trajectories Generated on the Robot.}{The blue dots show the eventual landing location of the trajectories in the training dataset.  For balls that do not hit the table, the dot shows the point where the ball trajectory intersects with the surface plane of the table.  The visualized distribution shows that a good percentage of the returned balls do land on the opponent's side of the table.  The visualization also shows that hitting target right behind the net is difficult for the robot, which is why the random landing targets are chosen to be at least 40\,cm behind the net.  The landing trajectories which are generated by the robot are used to train alternative dynamics models.  These models are used to evaluate whether training the models only on strikes that are known to be feasible on the robot improve performance.  Experiments show that this is not the case, thereby showing that requiring the analytic controller to reproduce striking motions demonstrated by humans does not limit the striking skill's performance.}
\label{fig:landball:coop-landing}
\end{figure}

\reffig{landball:coop-landing} shows the distribution of landing positions resulting from a sample set of landing trajectories generated with the above setup.  Since many of the returned balls do not land successfully, a total of 900,000 landing trajectories are generated to have plenty of landed balls to learn from.  The landing trajectories collected in this process are used to train the forward and inverse landing-prediction models described in \refsec{dyn}.  A land-ball policy using these dynamics models is evaluated using the same evaluation setup outlined in \refsec{landball:eval}.  \reftab{landball:robot} shows the evaluation results.

\begin{table}[htb!]
\centering
  \begin{tabular}{|l|c|c|r|r|}
  \hline 
  \thead{Method} & \thead{Search} & \thead{Data} & \thead{Return Rate} & \thead{Target Error} \\
  \hline 
  \hline 
  Model-Based & - & Robot & 92\% & 0.191 m \\
  Model-Based & CEM & Robot & 94\% & 0.203 m \\
  \hline
  \end{tabular}
\mycaption{Evaluation of Model-Based Land-Ball Policies Trained with Data Generated on the Robot.}{Mean ball return rate and mean landing target error computed over 600 attempts with random targets.  The policy uses dynamics models trained from 900,000 landing trajectories collected on the robot itself.  Comparing the results with \reftab{landball:modelbased} shows that these policies have lower target accuracies than policies trained from the human demonstrations collected in the VR environment.}
\label{tab:landball:robot}
\end{table}

Comparing the mean target errors with the policy from \refsec{landball:modelbased} shows that this policy achieves a lower accuracy for the land-ball task.  The training data for this policy uses only strikes that are guaranteed to be feasible on the robot since they were generated by the robot in the first place.  However, the higher accuracy of the policies trained with the VR data suggest that the analytic controller is able to execute arbitrary strikes demonstrated by the humans as well.  The most likely reason behind the lower accuracy in this policy is that the random strikes generated using fixed joint velocity targets are not general enough to capture all possible useful motions.


\subsection{Model-Free Land-Ball Policy}
\label{sec:landball:modelfree}

The model-based land-ball policies employ the dynamics models that are trained either from human demonstrations or random strikes generated on the robot.  In order to assess the impact of dynamics models on sample-efficiency and accuracy of the land-ball policies, this section describes and evaluates an alternative implementation of the policy which uses model-free reinforcement learning to train the policy from scratch.  The experiments help put the sample-efficiency and accuracy of the model-based land-ball policies into perspective.

\subsubsection{Training}

Unlike the hierarchical controller used in model-based land-ball policies, the model-free land-ball policies are expected to control the robot at the joint level and choose actions at every timestep.  Since the land-ball task is more difficult to learn under these conditions, the model-free agents are also trained and evaluated on a simpler task whose goal is to just successfully return the ball to the opponent's side.  The reward functions used for training the model-free policies are:

\begin{enumerate}
\item \textbf{Return-Ball}: The agent is expected to return the ball over the net successfully so that it bounces on the opponent side of the table.  This task has a binary binary reward $\delta_\text{return}$ indicating success or failure.
\item \textbf{Land-Ball}: The agent is expected to land the ball at a randomly-picked target position (with no specified target velocity).  The reward for this task is specified by $\delta_\text{return}(3 - \| l(b_g) - l(g)\|_2)$, where $l(b_g)$ denotes the observed landing position of the ball at the moment of contact with the table and $l(g)$ denotes the target landing position.  So, if agent the cannot return the ball, it receives a reward of zero.  Otherwise, it receives a positive reward depending on how close it gets to the target.
\end{enumerate}


\subsubsection{Evaluation}

This section describes the evaluation results obtained from learning model-free policies using Proximal Policy Optimization (PPO)~\cite{schulman2017proximal} and Augmented Random Search (ARS)~\cite{mania2018simple} algorithms.

The model-free policies use feed-forward neural networks with a single hidden layer of 20 units. The input observation vector includes the position and velocity of the ball and robot joints.  When trying to send the ball to a particular target location, the observation also includes the coordinates of the target position.  The output of the policy is the six-dimensional vector containing target robot joint velocities.

Learning the task with PPO was not very successful.  When trained on the simpler return-ball task, the PPO agent could achieve a successful ball return rate of 40\%.  Training the PPO agent on the more difficult land-ball task did not produce effective policies.

The ARS agent uses Algorithm 2 by Mania \ea~\cite{mania2018simple}, with normalized observations, step-size $0.05$ scaled by reward standard deviation, perturbation standard deviation $0.05$, and updates limited to the five top-performing perturbations out of the 30 evaluated.  The ARS algorithm applies perturbations to the policy weights and evaluates each perturbation using a rollout.  For the return-ball and land-ball tasks, each rollout corresponds to evaluating the policy on a new random target and a new random launcher ball in the environment.  Since the parameters of the launcher ball and landing target are different from episode to episode, the tasks are highly stochastic.  So, the algorithm is modified to evaluate each perturbation using up to 15 rollouts rather than a single rollout.  This increase in the number of evaluations led to faster training progress and better performance, possibly due to choosing policy update directions that perform well over the randomness in the environment.

\begin{figure}[htb!]
\centering
\includegraphics[width=0.7\linewidth, trim=1cm 0cm 1cm 1cm, clip]{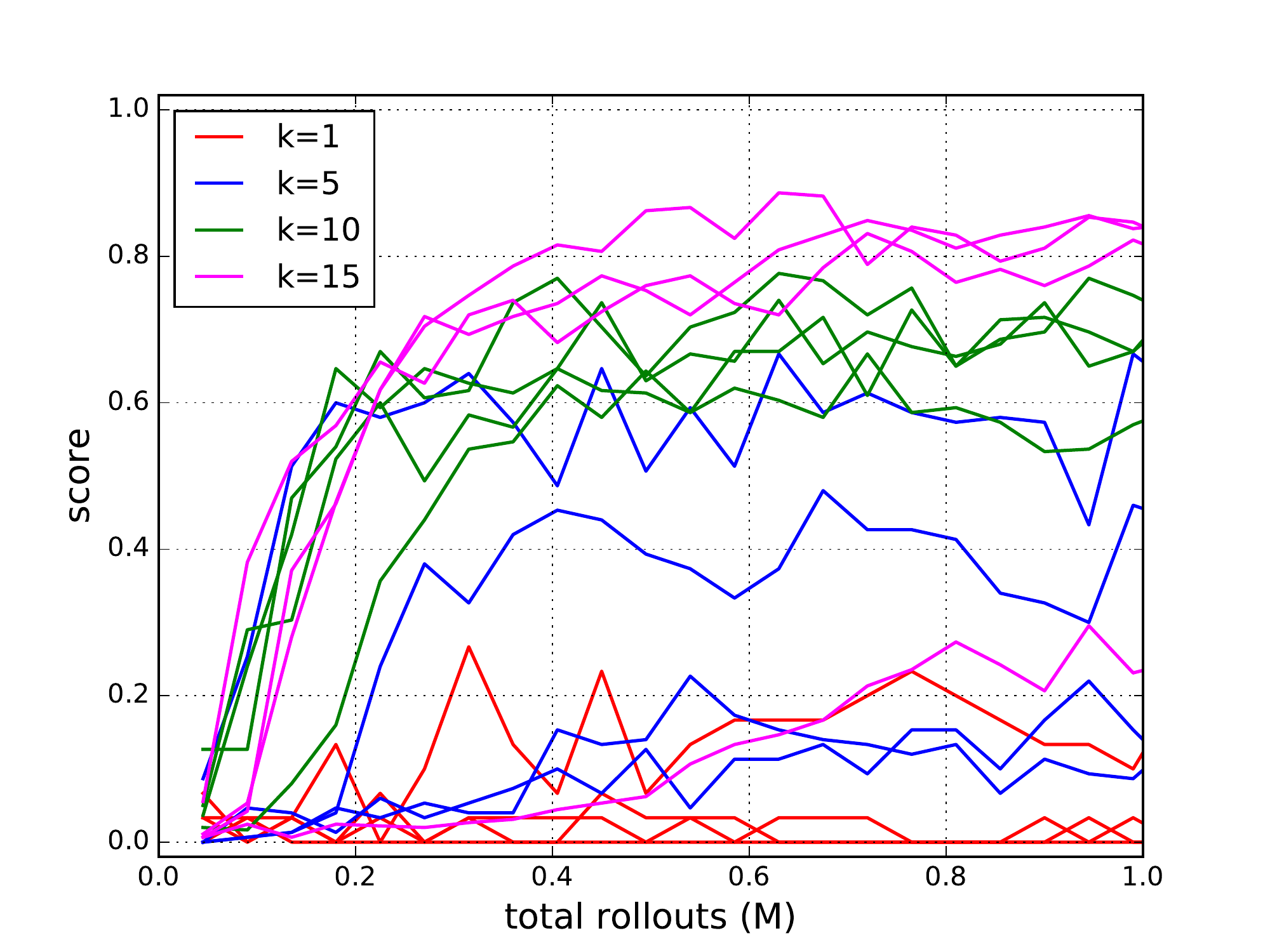}
\mycaption{Training Model-Free Return-Ball Policy with ARS.}{The figure shows training score vs. total number of rollouts on the binary-reward task of successfully returning the ball, for ARS runs with a different number of evaluations $k$ per perturbation (indicated by colors).  Evaluating each perturbation using multiple rollouts improves the training score, as the algorithm chooses update directions that perform well over the randomness in the environment.}
\label{fig:elandball:ars_num_evals}
\end{figure}

\reffig{elandball:ars_num_evals} shows the training rewards received by the ARS agent on the return-ball task under different number of evaluations $k$ per policy perturbation.  The figure shows four runs of the algorithm for each value of $k$.  Even though higher values for $k$ increase the number of evaluations needed, they improve the learning performance.  When each policy perturbation is evaluated multiple times, the stochasticity in the environment does not influence the  direction of policy weight updates as much.

\begin{figure}[htb!]
\centering
\includegraphics[width=0.7\linewidth]{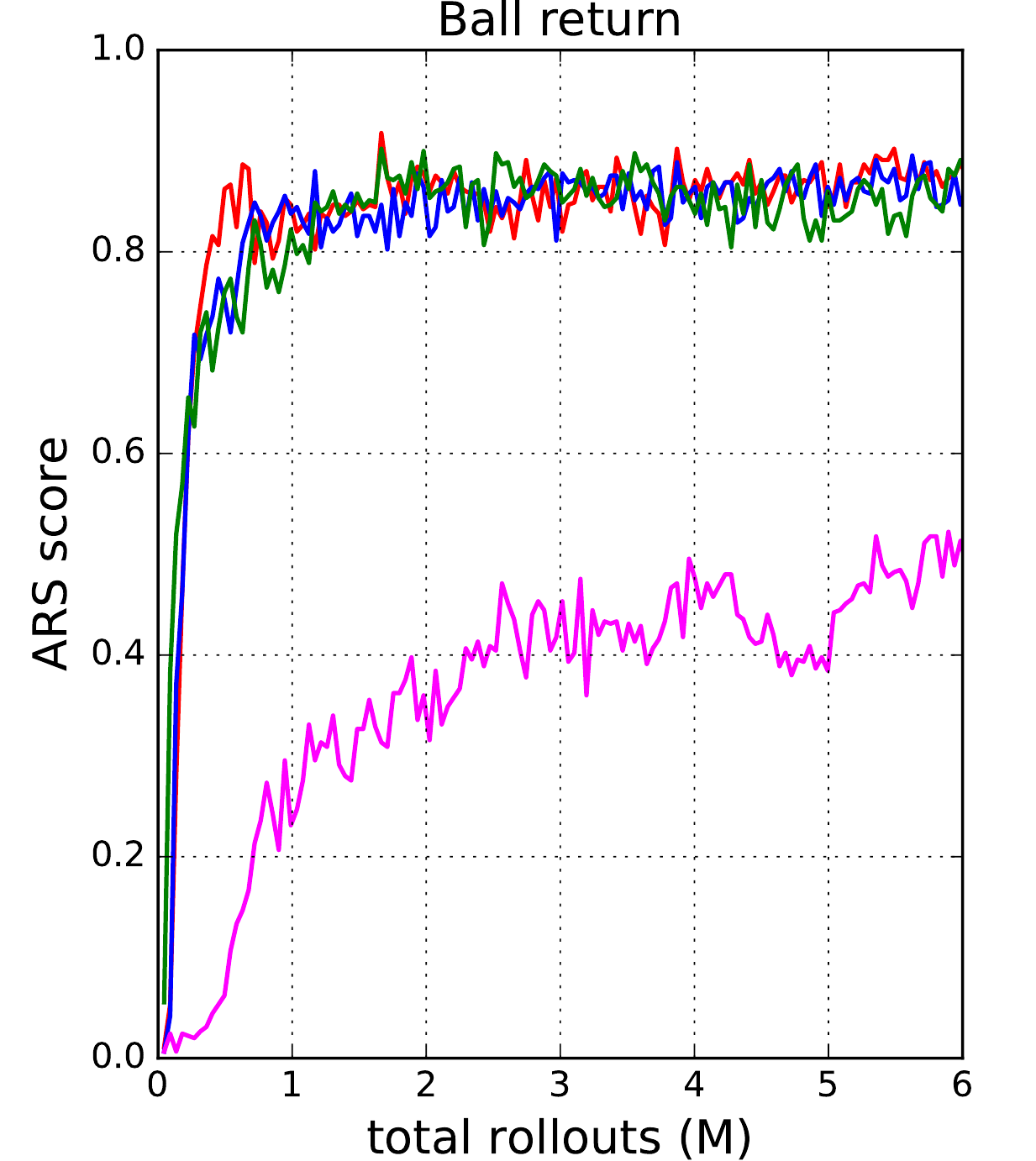}
\mycaption{Evaluation of Model-Free Return-Ball Policy Trained with ARS.}{The four lines show the evaluation scores received from four different runs of ARS on the return-ball task.  All runs use $k = 15$.  Learning the policy with model-free RL requires $\sim$1-2M episodes.  Such a high number of samples prevents using such algorithms on physical robots in the real world.}
\label{fig:elandball:ars_return}
\end{figure}

\reffig{elandball:ars_return} shows the evaluation rewards received by four different runs of the ARS training with $k = 15$.  ARS stopped making progress after $\sim$2M rollouts, and the best policy (out of four runs) succeeded $88.6\%$ of the time.  This experiment shows that ARS is able to discover a better model-free return-ball policy compared to PPO.

\begin{figure}[htb!]
\centering
\includegraphics[width=0.7\linewidth]{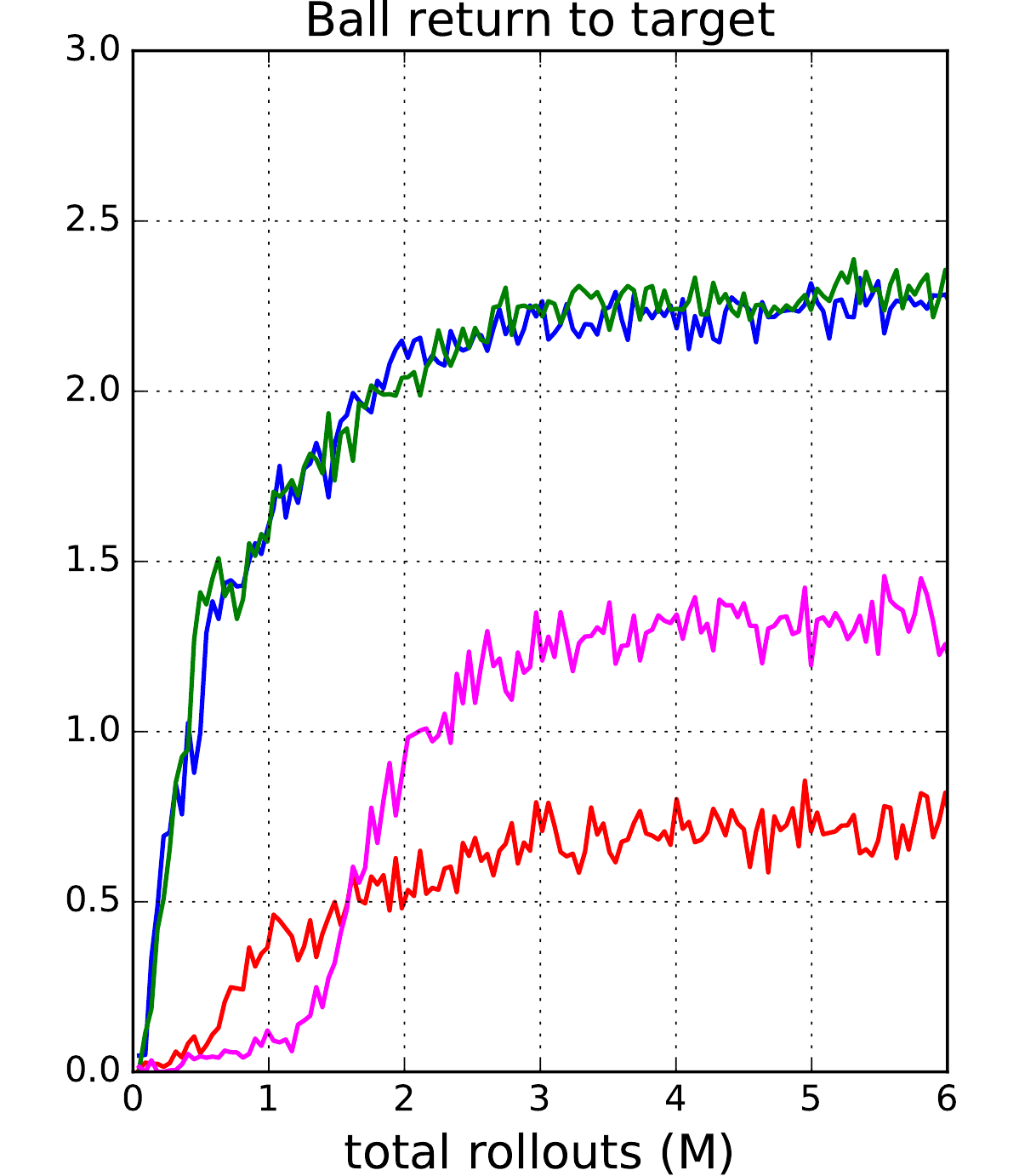}
\mycaption{Training Model-Free Land-Ball Policy Trained with ARS.}{The four lines show the ARS training score vs. total number of rollouts on the task of returning the ball to a particular target location.  All runs use $k = 15$.  Two out of the four runs perform better and achieve a mean reward of between 2.0 and 2.5, which as the evaluation show, corresponds to a landing accuracy of about 40\,cm.  Moreover, the model-free RL algorithm requires $\sim$3M episodes, which makes it impractical for the real world.}
\label{fig:elandball:ars_target}
\end{figure}

\reffig{elandball:ars_target} shows the training rewards received by the ARS agent on the land-ball task.  The agent converges after $\sim$3M rollouts.  Evaluating the best policy at the end of training shows a ball return rate of $88\%$ and an average landing distance of 0.4\,m from the target.  While this policy is fairly inaccurate, it is better than random.  In comparison, the best policy trained on the return-ball task achieves a distance of 0.52\,m from the target on average.

\reftab{landball:modelfree} summarizes the evaluation scores achieved by the model-free agents.  Overall, ARS produced substantially better results than PPO; policies trained using ARS were able to return the ball successfully $\sim$88\% of the time, while those trained using PPO only succeeded $\sim$40\% of the time.  However, the ARS agent used $\sim$3M training episodes to learn the land-ball task.  It is not feasible to run such a large number of training episodes on physical robots in the real world.  Comparing the performance of the model-free policies to the model-based hierarchical policies in  \reftab{landball:modelbased} shows that the model-based approach produces policies with much lower error (12\,cm vs. 40\,cm) and using far fewer samples (140,000 vs 3M).  Moreover, the samples used for model-based policies come from human demonstrations, which are far easier to obtain than samples collected with the robot itself.

\begin{table}[htb!]
\centering
  \begin{tabular}{|l|c|c|r|r|}
  \hline 
  \thead{Method} & \thead{Algorithm} & \thead{Samples} & \thead{Return Rate} & \thead{Target Error} \\
  \hline 
  \hline 
  Model-Free & PPO & 0.1\,M & 40\% & - \\
  Model-Free & ARS & 3\,M & 88.6\% & 0.4 m \\
  \hline
  \end{tabular}
\mycaption{Evaluation of Model-Free Land-Ball Policies.}{Mean ball return rate and mean landing target errors for random targets.  Comparing the results with \reftab{landball:modelbased} shows that this policy has a lower target accuracy.}
\label{tab:landball:modelfree}
\end{table}

\subsection{Model-Based Hit-Ball Policy}

The model-based land-ball policies require three trained dynamics models to operate.  Using the dynamics models make the policy implementations sample-efficient.  On the other hand, if the models are not updated with additional trajectories collected using other policies, the policies that depend on those models would stay limited in their behavior.

\subsubsection{Policy Implementation}

The hit-ball policy is an alternative striking policy that can execute more versatile striking motions.  It uses only one dynamics model (the ball-trajectory prediction model).  Using the ball-trajectory model makes it easy for the policy to ensure contact with the ball.  The position of the paddle is simply set to the prediction position of the ball at the desired time of contact.  The other parameters of the paddle-motion target (paddle orientation, linear and angular velocity) are given as inputs to the hit-ball policy.  So, a higher-level skill like the strategy skill can use the hit-ball policy to execute arbitrary strikes without consulting any additional models to decide a target paddle-motion state.

As described in \refsec{skill:hit-ball}, the hit-ball task is specified as:
\\
\begin{align}
\pi_h(l_x(p_t), N(p_t), v(p_t), \omega(p_t) \mid b_s) = \pi_p(t, p_t \mid p_s).
\end{align}
\\
where $l_x(p_t)$ denotes the desired contact plane.  The policy is instructed to hit the incoming ball when it reaches the plane $x = l_x(p_t)$.

\begin{figure}[htb!]
\centering
\includegraphics[width=0.9\columnwidth]{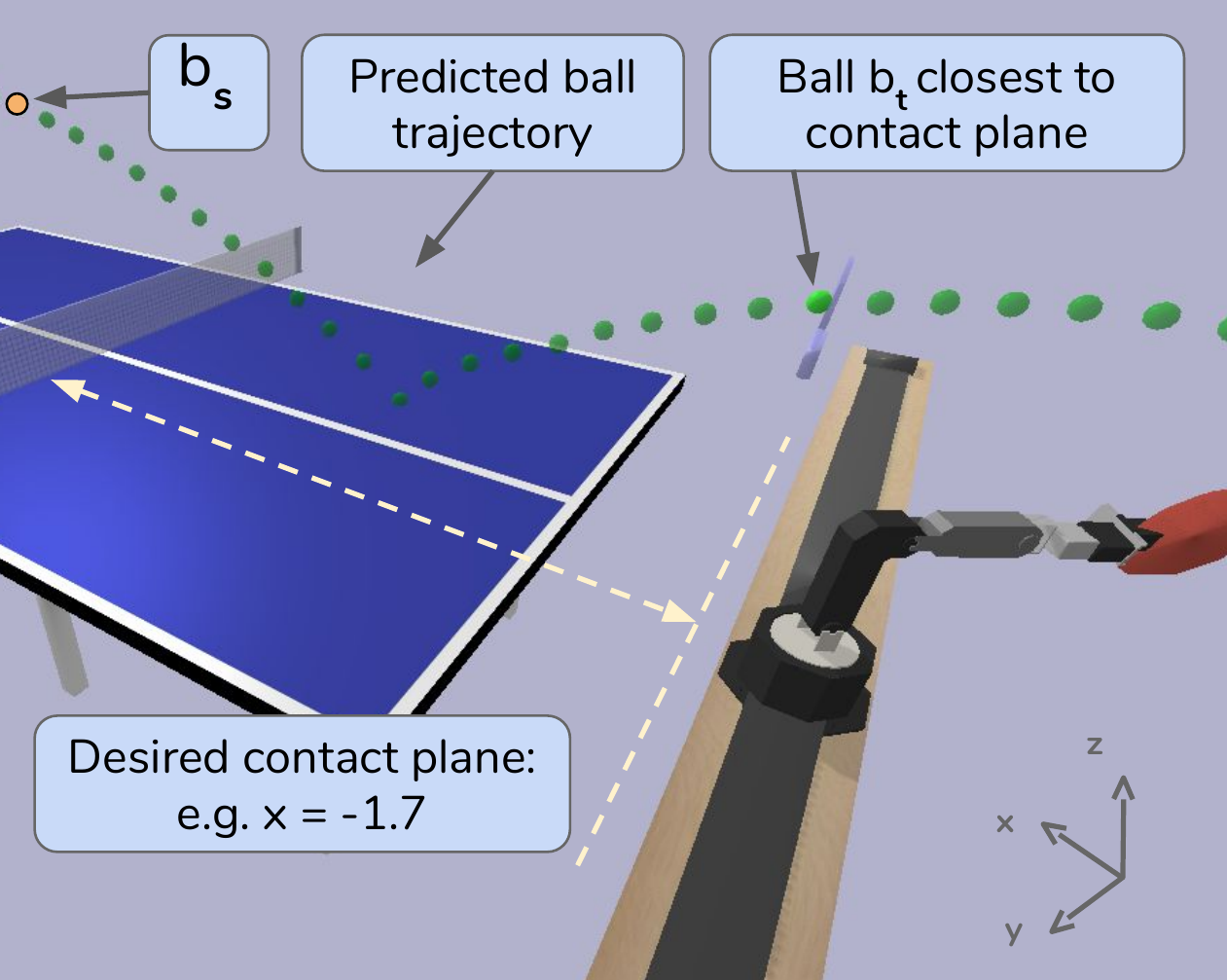}
\mycaption{Model-Based Hit-Ball Policy.}{The action space of the hit-ball skill includes a desired contact plane, which indirectly specifies a target point for striking the ball.  Given the current ball-motion state $b_s$, the hit-ball policy predicts the future ball trajectory.  The point $b_t$ in the predicted trajectory that is closest to the requested contact plane is picked as the striking target.  This point, which is visualized as the light-green ball, specifies both the target position for the paddle $l(p_t)$, and the target time $t$.  The other components of the paddle-motion target, including its normal vector, and linear and angular velocities are given as inputs to the hit-ball policy.  Using the simple ball-trajectory model helps the hit-ball policy make contact with the ball easily.  On the other hand, accepting any orientation or velocity for the paddle allows the hit-ball skill to execute arbitrary paddle strikes.  Thereby, the hit-ball skill is more flexible than the land-ball skill.}
\label{fig:landball:hitball}
\end{figure}

\reffig{landball:hitball} illustrates the implementation of the hit-ball policy.  The ball-trajectory prediction model is used to predict the incoming ball's trajectory based on the ball's initial motion state $b_s$.   The policy then intersect the ball trajectory with the the desired contact plane $x = l_x(p_t)$ and picks the point $b_t$ that in the ball trajectory that is closest to this plane.  This point determines the contact time $t$ and the full position of the paddle $l(p_t)$ at the time of contact.  The other attributes of the paddle's motion state $N(p_t), v(p_t), \omega(p_t)$ are given as inputs to the hit-ball policy.  Together with the computed paddle position, they fully specify the paddle-motion target $p_t$, which is passed as a high-level action to the paddle-control skill.

\subsubsection{Evaluation}

The hit-ball policy can not be easily evaluated in isolation since it depends on a higher-level policy to specify the orientation and velocity of the paddle at the time of contact.  This policy is used in the next section by a strategy agent to play whole table-tennis games.  The hit-ball policy allows the strategy agent to discover strikes that are particularly useful for cooperative or adversarial games, beyond what the model-based land-ball policy produces.

\subsection{Conclusion}

This section described multiple implementations for the land-ball policy.  Experiments showed that:

\begin{itemize}
\item The model-based land-ball policy trained from VR data is very sample-efficient and can hit ball targets with a mean position error of only $12$\,cm.
\item Training the model-based land-ball policy with data collected directly on the robot does not improve its performance.  So the analytic paddle controller is able to execute arbitrary strikes demonstrated by humans.
\item It is difficult to learn the land-ball skill with model-free learning methods that control the robot at joint level.  Such model-free policies require about 3M training samples and achieve a target landing accuracy of about $40$\,cm.
\end{itemize}

Although the land-ball skill itself can capture any game-play behavior, the canonical land-ball policy implemented in this article uses dynamics models trained only from human demonstrations.  So, the implemented land-ball policy only executes paddle strikes that were used by humans.  The trained models might generalize to use blended strikes within the space of demonstrations, but the models cannot output strikes that lie completely outside the space of demonstrations.  For example, if the humans have never hit the ball from below so it goes high up, the trained models cannot produce or evaluate such strikes.  On the other hand, this choice of implementation is not a limitation on the approach.  It is possible to fine-tune the landing models trained from human demonstrations with data collected from other striking policies, as done in the dagger method~\cite{ross2011reduction}.  Such an approach can combine the benefit This approach is left as part of future work.

The next section describes how the striking policies can be used by the top-level strategy skill in the task hierarchy to play cooperative and adversarial table-tennis games.

\section{Learning Strategy with Self-Play}
\label{sec:strategy}

This section describes how the strategy skill is trained.  It then evaluates different variants of the strategy skill that use different striking skills.

\subsection{Approach}

The striking skills discussed in section \refsec{striking} make it possible to execute high-level paddle actions that either send the ball to a desired target location (land-ball skill), or make contact with the ball with a desired paddle orientation and velocity (hit-ball skill).  Similarly, the positioning skill described in \refsec{paddle:positioning} can be used to move the robot from any motion state to a desired pose in minimum time so that they robot can be ready for responding to the next shot from the opponent.  However, the striking and positioning skills are parameterized skills.  They should be given goals to execute.  Playing successful table-tennis games using the striking and positioning skills requires deciding appropriate targets for them during each exchange of the ball with the opponent.  It is possible to operate the underlying skills using hand-coded targets.  However, such a hand-coded implementation would be very limiting in exploiting the diverse behaviors that are possible in the striking and positioning policies.  So, this method uses a model-free RL algorithm to learn diverse game-play strategies.

\subsubsection{Reinforcement Learning}

The task hierarchy discussed in \refsec{skill} addresses this problem by including a top-level strategy skill whose only job is to pick targets for the striking and positioning skills.  An effective strategy policy requires complex reasoning about the opponents, which is difficult to do in a model-based way.  So, the strategy skill is learned with model-free reinforcement learning.  The decomposition of the tasks and the abstractions offered in the skill hierarchy makes it possible to combine this model-free strategy layer with model-based striking skills and the analytic robot controller in a single agent.

In the method implemented in this disseration, the strategy skill is the only skill that needs to be trained on the robot.  Although model-free reinforcement learning is not sample-efficient, the design of the hierarchical policy simplifies learning the strategy layer, thereby making model-free learning applicable to the problem.  Since the strategy skill does not need to be aware of the internal implementation of the underlying skills, its inputs and outputs can stay low-dimensional.  Also, since the striking and positioning skills have long time horizons spanning over an entire ball exchange with the opponent, the strategy skill gets to observe the game at a high-level.  Therefore, it can focus on the high-level game-play.  If the strategy skill picks a bad landing target or a bad waiting position in a game, it receives the negative reward immediately in the next step.  The long time horizon helps eliminate the reward delay, which can make learning this skill more efficient.  Since the skill can be trained after the striking and position policies are developed, all the training episodes used when learning the strategy can focus on exploring the space of strategies, rather than also being concerned with learning striking and positioning policies.

\subsubsection{Cooperative and Adversarial Rewards}

This article considers two types of table-tennis games: cooperative and adversarial.  In a cooperative game, the objective of the two players is to keep the game going as long as possible.  In this mode, the players receive one point each time they land the ball anywhere on the other player's side.  The cooperative reward function also gives the players one point each time they hit the ball with the paddle.  Since the land-ball and hit-ball policies automatically make contact with the ball, they do not benefit that much from this reward shaping mechanism.  However, this section also includes experiments with learning a strategy directly over the paddle-control skill.  Reward shaping is useful for learning such a low-level strategy.  So, it is included for all agents to make comparison of rewards easier.  With reward shaping, a cooperative agent can receive a maximum reward of +4 per step.

In adversarial games, the objective is to win the rally.  Each player tries to land the ball over the opponent's side in a way that the opponent cannot successfully return the ball.  In this mode, the player who wins the rally receives one point.  Players also receive a reward of 0.1 for making contact with the ball or successfully landing the ball anywhere on the opponent's side.

\subsubsection{Self-Play}

\begin{figure}[htb!]
\centering
\includegraphics[width=0.9\columnwidth]{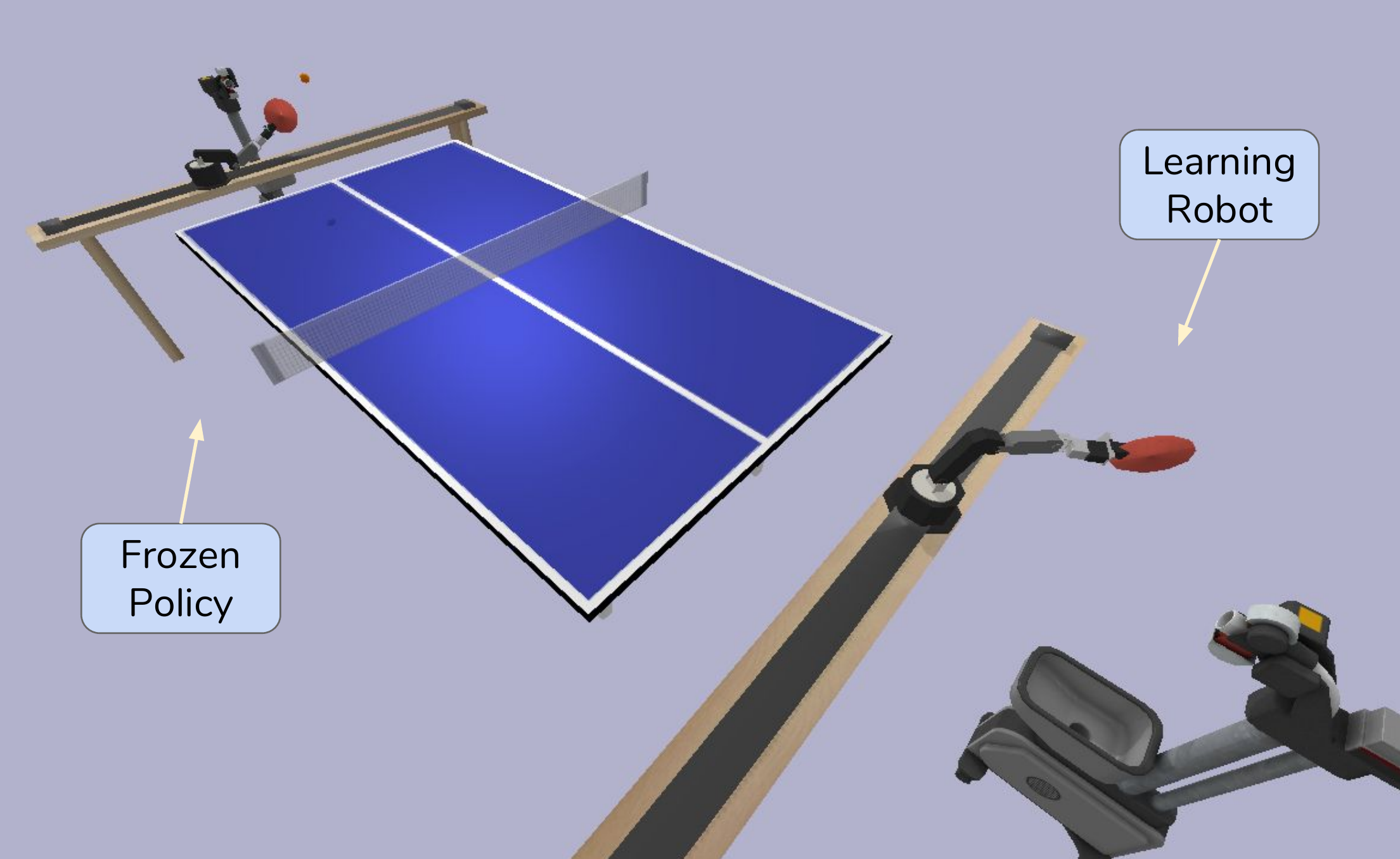}
\mycaption{Self-Play Setup.}{Two robots play cooperative or adversarial games with each other.  The rallies are started randomly by one of the two ball launchers behind the robots.  Only one of the two robots is learning the strategy policy.  During each self-play level the learning robot plays against a its own frozen policy from the end of the previous level.  As more self-play games are played, the strategy policy learns to adapt and respond to its own behavior.}
\label{fig:strategy:self-play}
\end{figure}

Since the strategy skill does not have a specific target, it is best to learn the skill using self-play.  The only objective for the strategy skill is to maximize the reward.  Since the reward received depends also on the behavior of the opponent, the skill needs to adapt to the behavior of the opponent.  This goal is achieved in a self-play setup, which is shown in  \reffig{strategy:self-play}.  Two identical robot assemblies are mounted on the two sides of the table.  Since the agent policies do not support serving the ball, the rally is always started with a ball launcher.  There is a launcher behind each robot and the rally starts randomly from one of the two ball launchers.  When the ball goes out of play, the next ball is launched and a new rally starts.

Of the two robots playing against each other, only one is actively training.  The other robot is playing using a frozen policy.  Self-play learning is done over a number of levels.  During the first level, the learning agent plays against an initial fixed policy.  After the end of the first level, the learner's policy is frozen and transferred over to the opponent.  So, during level $i$, the learner plays against its own frozen policy at the end of level $i - 1$.

\subsubsection{Observations and Actions}

Each step in the environment lasts over an entire ball exchange.  Each ball exchange can include two strikes, one by the learner and another by the opponent.  In the current implementation, the strategy skill receives only the current ball-motion state $b_s$ as the observation.  Preliminary experiments showed that adding other observations like the agent's paddle position $l(p_t)$ or the opponent's paddle position do not improve performance in the current implementation.  It is likely that in a setup where the agent can make more than one decision per exchange including such additional observations would be useful.  As actions, the strategy skill produces targets for the striking skill and the positioning skill.

In the current implementation, the strategy skill specifies a waiting position for the robot only once per exchange.  This choice is made because the agent does not receive observations from the opponent's movements as the opponent approaches the ball to hit it back.  Therefore, the strategy agent does not observe opponent's position when deciding on a target for the positioning skill.  However, the behavior of the opponent and how it approaches to return the ball may contain clues for the other agent as to how to best position itself to be prepared for the return shot.  In an alternative implementation where the positioning skill is invoked in a continuous manner, it would make sense to also include the opponent's position in the observations.


\subsection{Training Setup}

All strategy policies except for the joint-level strategy policy are implemented using the TensorFlow implementation of batched PPO~\cite{hafner2017agents}.  The PPO agents use feed-forward policy and value networks.  Both the policy and value networks have two hidden layers of size 10 each.  The policy network produces a mean and a standard deviation value for each action dimension.  Therefore, the learned policies are stochastic.  The stochastic actions are active for the frozen opponent policy and for evaluation of the learner's policy as well.  In addition to the environment rewards, the algorithm also uses an entropy reward with a coefficient of 0.1 to encourage more randomness in actions.  This stochasticity is useful for preventing the self-play agents from converging to very narrow policies.  The learning rate for the policy network is $10^{-4}$ while the learning rate for the value network is $10^{-3}$.  

Each self-play level consists of 2400 training steps, which can span over any number of episodes.  The method uses PPO with a batch size of 24, where each worker trains for a 100 steps.  Each training and evaluation episode runs for a maximum of 10 steps.  So, if a cooperative game reaches 10 steps, it is terminated even if the rally is still going on.  Terminating the episodes early allow the agent to get more training time over difficult initial conditions where the first launcher ball is more difficult to return.


\subsection{Land-Ball Strategy}

The canonical model-based land-ball policy in this article is trained solely from human demonstrations and does not use any training episodes on the robot.  Given a landing target $g$, the land-ball policy consults the dynamics models to find a paddle strike that can land the ball at $g$.  Even though the land-ball policy is limited to using human paddle strikes, it can still execute landing targets that were never tried in human demonstrations.  The normalization process used in training the dynamics models makes them capable of executing any targets resulting from translation and rotation of observed landing trajectories.  Training a strategy policy over the model-based land-ball skill amounts to learning a better approach to playing the game using the same ball-handling technique demonstrated by human players.

When playing adversarial games, a common strategy is to aim for targets near the right or left edges of the table.  However, in the action space of the land-ball policy, these areas lie on different regions of the action space.  So, even with the entropy terms, a strategy is likely to aim only for the right or the left side of the table.  To address this problem, when deciding land-ball targets, the land-ball strategy produces a target position $l(g)$, a target speed $|v(g)|$, and a probability $p_\text{flip}$ for flipping the target across the $y$ dimension (left-right) over the table.  The flip probability allows the strategy to specify targets near the right and left edges of the table without having to generate landing coordinates explicitly for both right and left sides.

\begin{figure}[htb!]
  \centering
  \includegraphics[width=0.9\columnwidth]{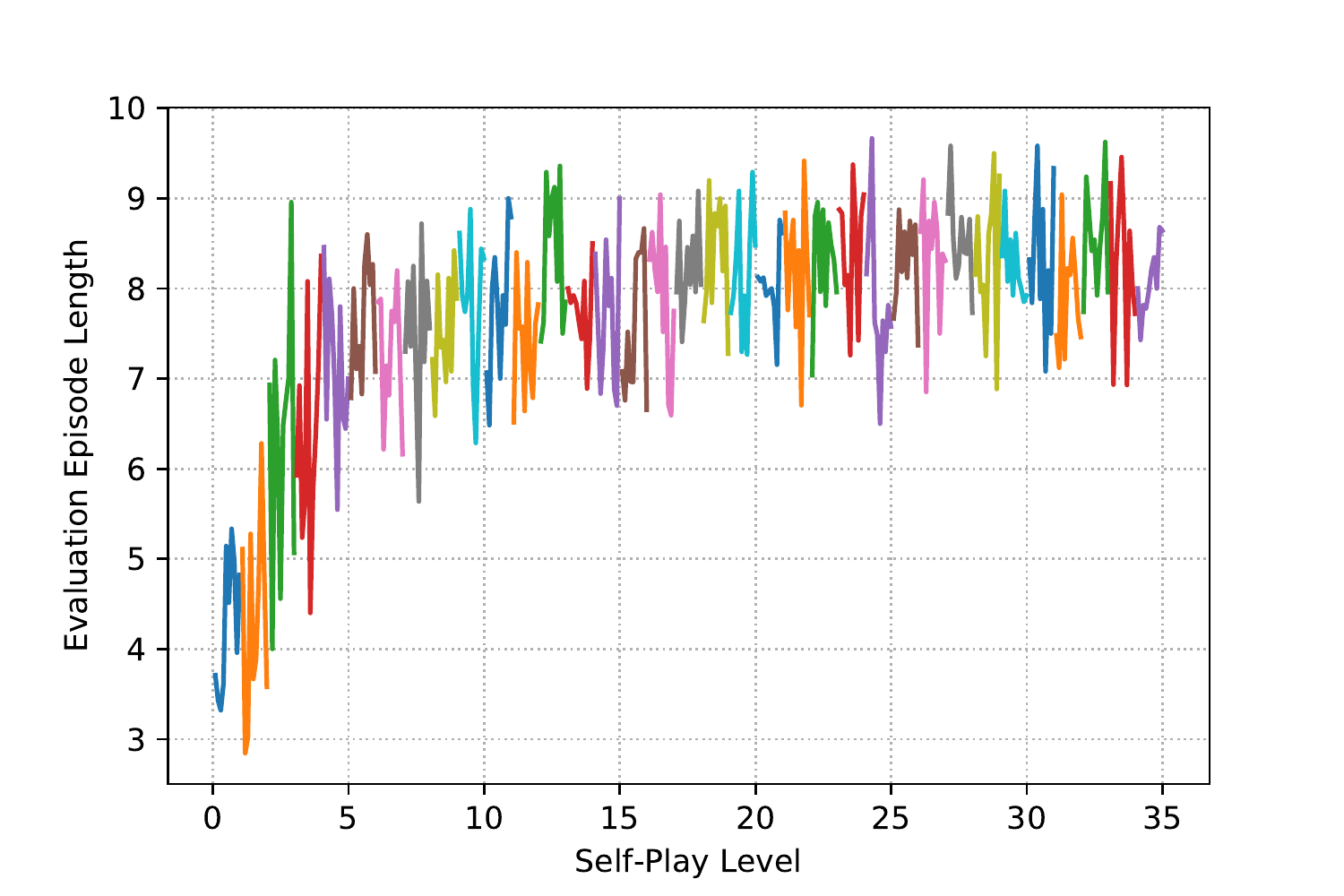}
  \mycaption{Self-Play Learning of Cooperative Land-Ball Strategy.}{Mean evaluation episode length over 35 self-play levels.  Each colored line segment corresponds to a different level.  The vertical axis shows the average length of a cooperative episode in evaluation.  Each point is an average over 240 evaluation episodes.  The maximum episode length is 10.   The cooperative agent converges to an episode length of about eight after about 10 self-play levels (24,000 training strikes.)}
  \label{fig:estrategy:landballcoop}
\end{figure}

\begin{figure}[htb!]
  \centering
  \includegraphics[width=0.9\columnwidth]{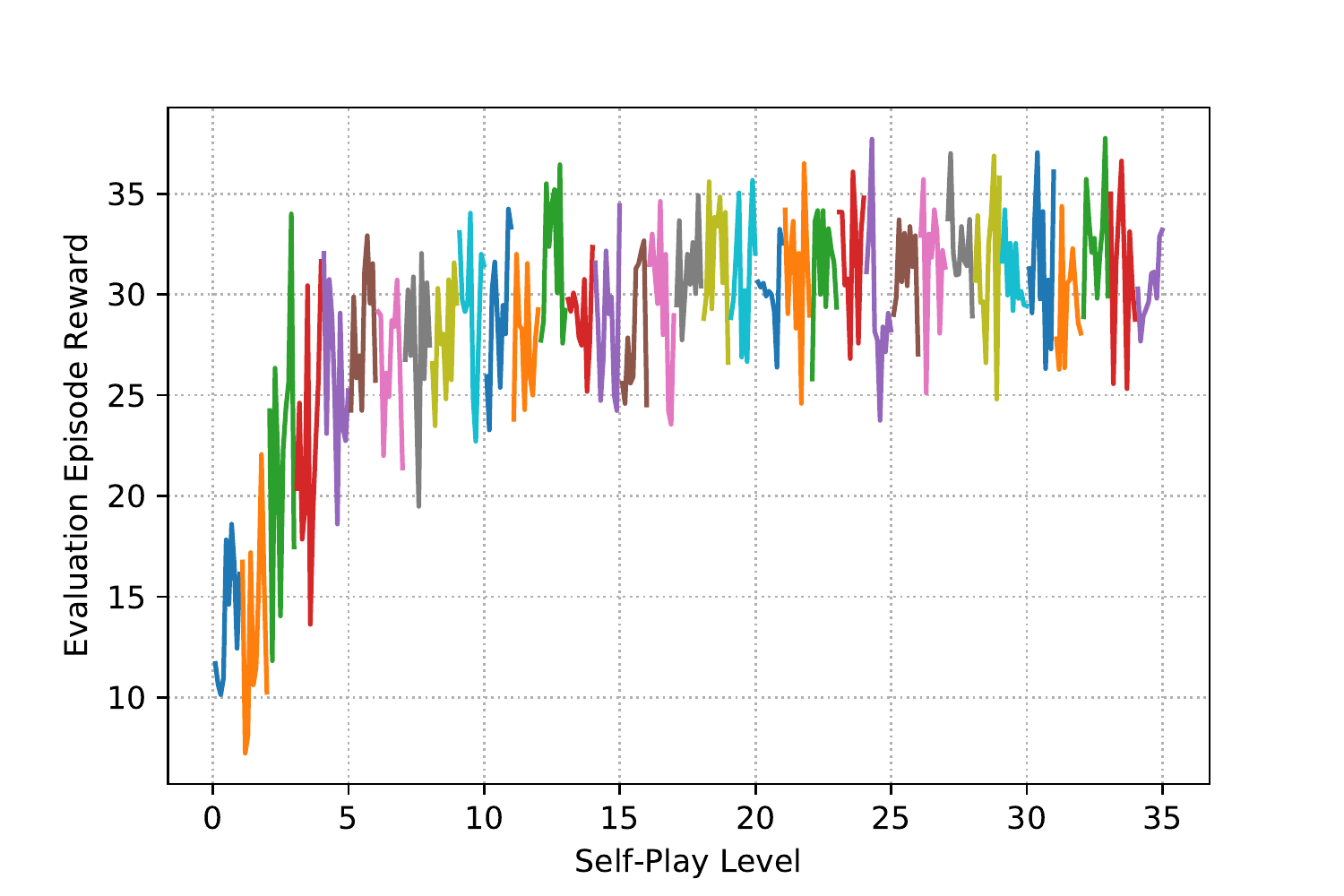}
  \mycaption{Average Evaluation Rewards for Cooperative Land-Ball Strategy.}{Mean evaluation reward over 35 self-play levels.  The average reward equals the episode length multiplied by 4.0.}
  \label{fig:estrategy:landballcoopscore}
\end{figure}

\reffig{estrategy:landballcoop} shows the training progress for the land-ball strategy for cooperative games.  The plot spans over 35 self-play levels and shows the average length of a cooperative episode in evaluation.  The cooperative agent converges to an episode length of eight after about 10 self-play levels (24000 training steps.)  The policy is initialized to have a mean action corresponding to a target near the center of the opponent's side.  As shown in the plot, the strategy skill can sustain a mean episode length of four right at the start of training.  This high performance shows that the model-based land-ball policy is effective.  Since the analytic paddle controller and the model-based land-ball policy do not use the robot for training, this experiment demonstrates zero-shot learning in cooperative table tennis.  \reffig{estrategy:landballcoopscore} shows the average evaluation reward for the cooperative land-ball strategy.  The average reward equals the episode length multiplied by 4.0.

\begin{figure}[htb!]
  \centering
  \includegraphics[width=0.9\columnwidth]{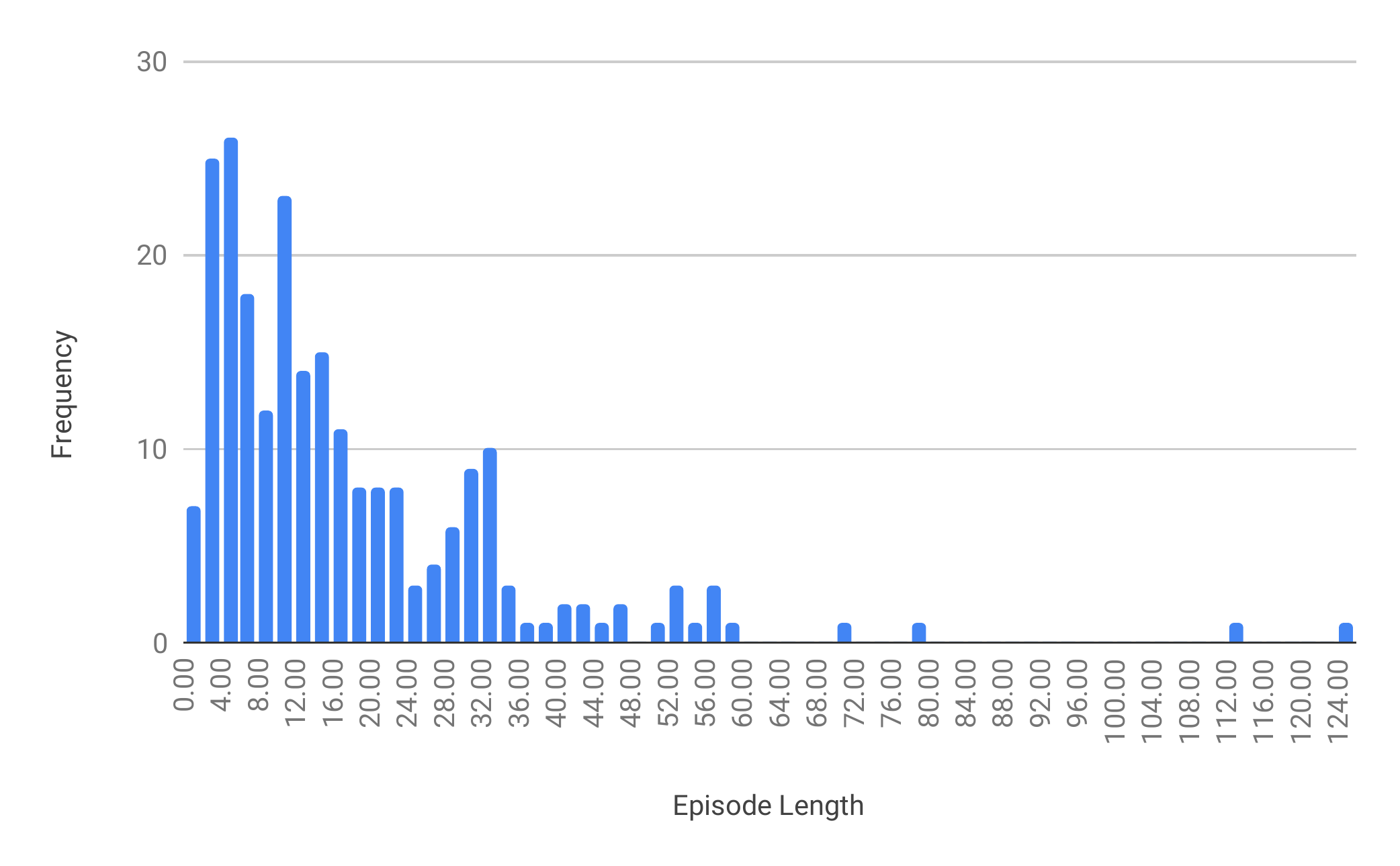}
  \mycaption{Histogram of Episode Lengths for Cooperative Land-Ball Strategy.}{Distribution of episode length over 240 test episodes with an episode length cap of 1000.  The land-ball strategy achieves a mean episode length of 17.4 with a standard deviation of 17.1.}
  \label{fig:estrategy:landballcooptest}
\end{figure}

After about 10 self-play levels, the evaluation reward does not improve much more.  The agent trained at the end of the 35\textsuperscript{th} level is evaluated against itself in 240 cooperative episodes.  In these episodes, the length cap is raised to 1000, so the agents are allowed to continue the rally for up to 1000 exchanges.   \reffig{estrategy:landballcooptest} shows the histogram of episode lengths observed in these evaluation games.  The mean episode length is 17 steps, with the maximum length reaching 124.

\begin{figure}[htb!]
  \centering
  \includegraphics[width=0.9\columnwidth]{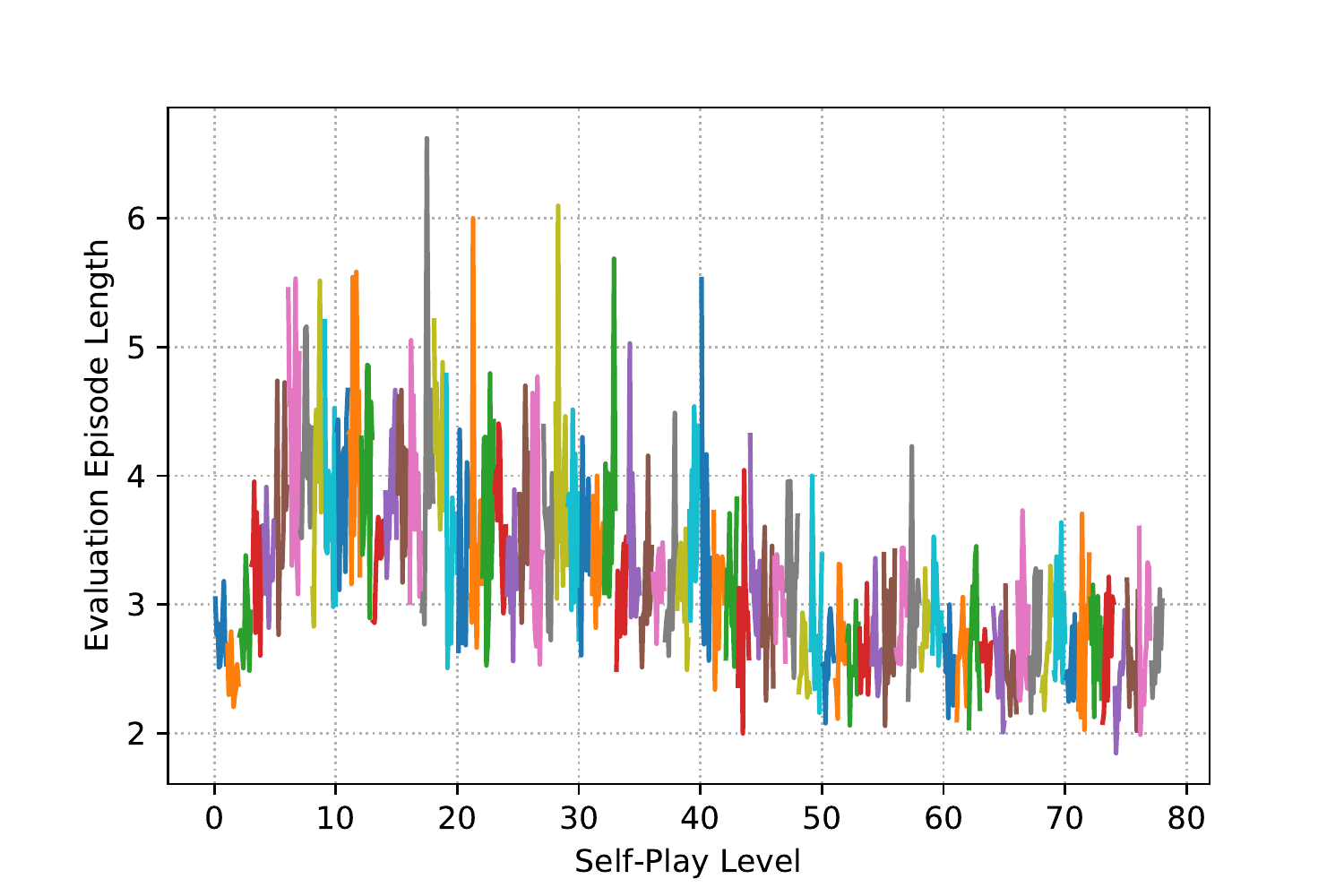}
  \mycaption{Self-Play Learning of Adversarial Land-Ball Strategy.}{Mean evaluation episode length over 77 self-play levels.  Each colored line segment corresponds to a different level.  The vertical axis shows the average length of an adversarial episode in evaluation.  Each point is an average over 240 evaluation episodes.  The maximum episode length is 10.  As the training process continues, the agent learns to win the point more quickly.  The mean episode length reaches four at the start of the training process, and then goes down to about 2.5 after 77 levels.}
  \label{fig:estrategy:landballadv}
\end{figure}

\begin{figure}[htb!]
  \centering
  \includegraphics[width=0.9\columnwidth]{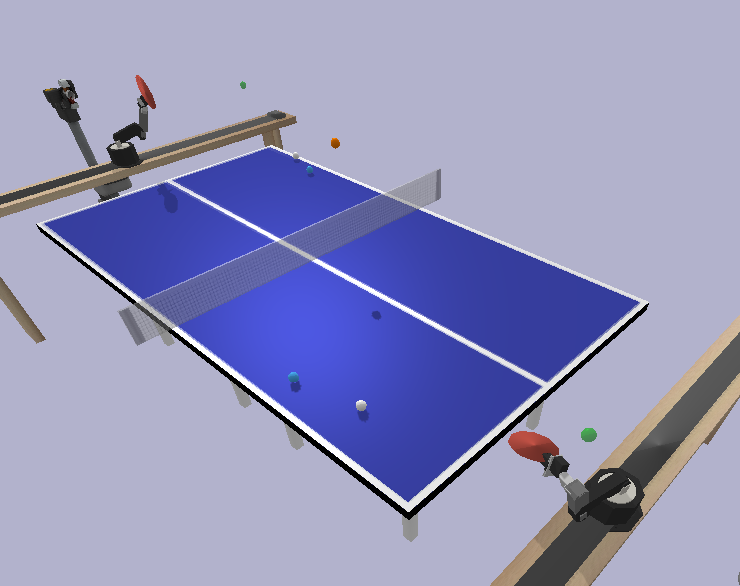}
  \mycaption{Visualization of Adversarial Land-Ball Strategy.}{The agent learns to use fast balls or aim for targets near the edges of the table.  In the example shown both robots have aimed for such targets.  Landing targets are indicated by white marker balls.}
  \label{fig:estrategy:landballadvdemo}
\end{figure}

The land-ball striking skill can also be used to learn adversarial strategies.  \reffig{estrategy:landballadv} shows the training progress for an adversarial land-ball strategy.  After 77 levels, the mean episode length reaches about 2.5, meaning that on overage the episode ends after five strikes from the two robots.  Visualizing the trained strategy shows that the agents try to hit the ball with high velocity or aim closer to the edges of the table.  \reffig{estrategy:landballadvdemo} visualizes an example.

Although the land-ball strategies can discover game-plays that are different from the game-plays employed by humans, the agents are limited to the striking techniques that are present in human demonstrations.  The next section describes how a hit-ball strategy can learn new striking techniques as well.


\subsection{Hit-Ball Strategy}
\label{sec:strategy:hit-ball}

The hit-ball skill helps the agent make contact with the ball by setting the target position of the paddle to be equal to the predicted position of the ball when it crosses a desired contact plane.  The land-ball skill just expects its user to specify the paddle orientation (normal) and linear and angular velocities at the moment of contact.  Learning a hit-ball strategy is more difficult, since hit-ball's action space has more dimensions than that of land-ball.  On the other hand, a hit-ball strategy is free to use arbitrary paddle strikes.

\begin{figure}[htb!]
  \centering
  \includegraphics[width=0.9\columnwidth]{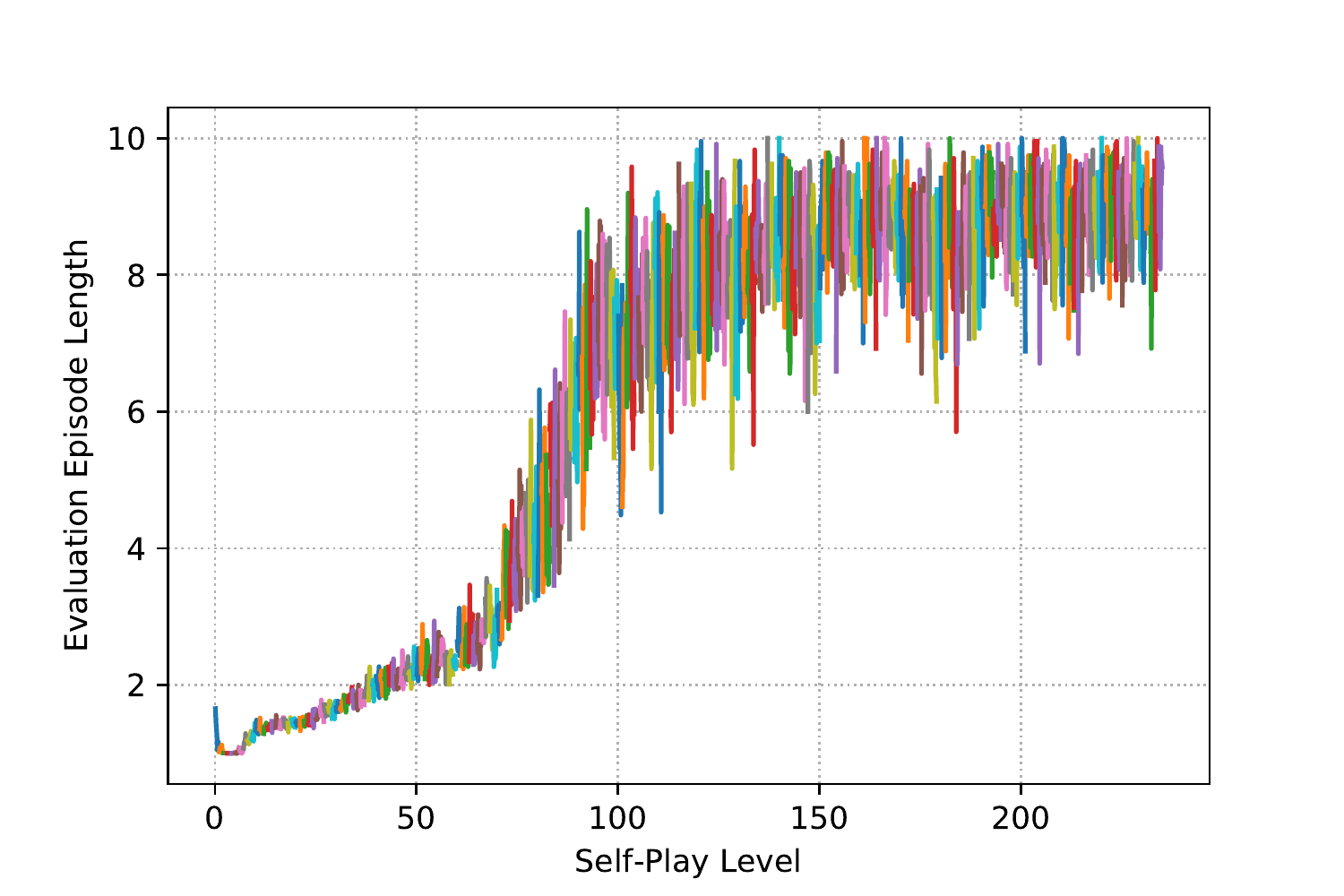}
  \mycaption{Self-Play Learning of Cooperative Hit-Ball Strategy.}{Mean evaluation episode length over 233 self-play levels.  Each colored line segment corresponds to a different level.  The vertical axis shows the average length of a cooperative episode in evaluation.  Each point is an average over 240 evaluation episodes.  The maximum episode length is 10.   The training progress for the hit-ball strategy is much slower than the land-ball strategy, however, given enough training the hit-ball strategy can discover novel strikes that allow it to score better than the land-ball strategy.}
  \label{fig:estrategy:hitballcoop}
\end{figure}

\reffig{estrategy:hitballcoop} shows the training progress for a cooperative hit-ball strategy.  The plot spans over 233 self-play levels.  The hit-ball strategy starts with a mean episode length of one.  After about 75 self-play levels, the mean episode length reaches four, which is the value achieved at the start for the land-ball strategy.  After about 125 levels, the hit-ball strategy achieves a mean episode length of eight, and then it surpasses it to get close to nine on average.  At times, the agent can achieve a the maximum episode length of 10 over 240 evaluation episodes.

\begin{figure}[htb!]
\centering
\includegraphics[width=0.9\columnwidth]{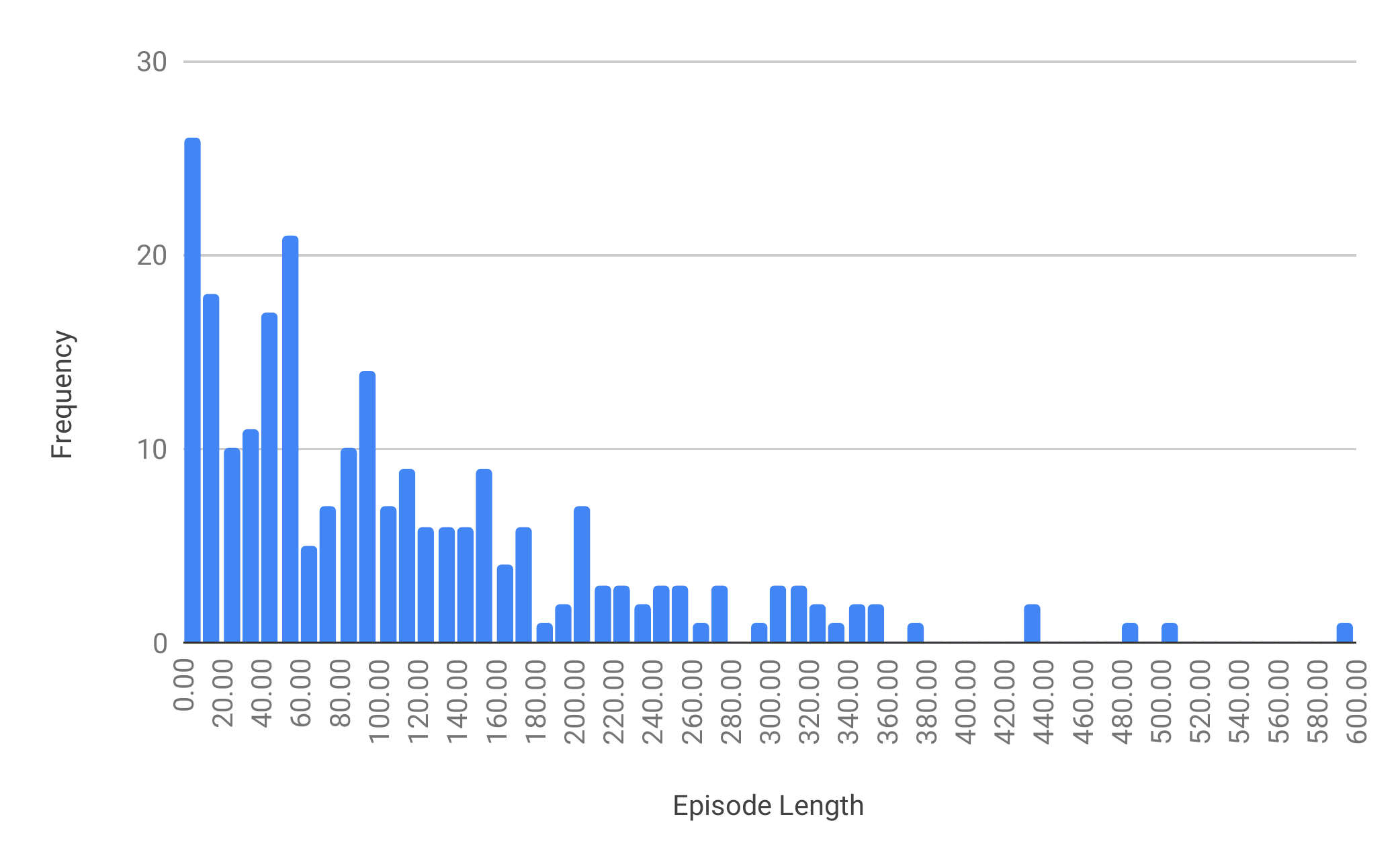}
\mycaption{Histogram of Episode Lengths for Cooperative Hit-Ball Strategy.}{Distribution of episode length over 240 test episodes with an episode length cap of 1000.  The hit-ball strategy achieves a mean episode length of  111.9 with a standard deviation of 105.9.  It significantly outperforms the land-ball strategy.  However, training the hit-ball strategy requires far more samples than the land-ball strategy.}
\label{fig:estrategy:hitballcooptest}
\end{figure}

Learning a cooperative hit-ball strategy is quite slower than learning a land-ball strategy, so it is less sample-efficient.  On the other hand, it can achieve higher rewards.  \reffig{estrategy:hitballcooptest} shows the histogram of episode lengths observed in 240 evaluation games run with the final cooperative hit-ball policy.  The agent achieves a mean episode length of 111.9 with the maximum reaching 600.  Visualization of the policies leads to an interesting observation.  The agents have learned to hit the ball rather slowly and with a paddle motion that causes it to land at about a 45 degree angle on the other player's side.  Such high balls are less likely to hit the net or go out.  Moreover, after bouncing on the table, the ball usually reaches the top of its arc in front of the other robot, making it easier for the other player to return it.

\begin{figure}[htb!]
  \centering
  \includegraphics[width=0.9\columnwidth]{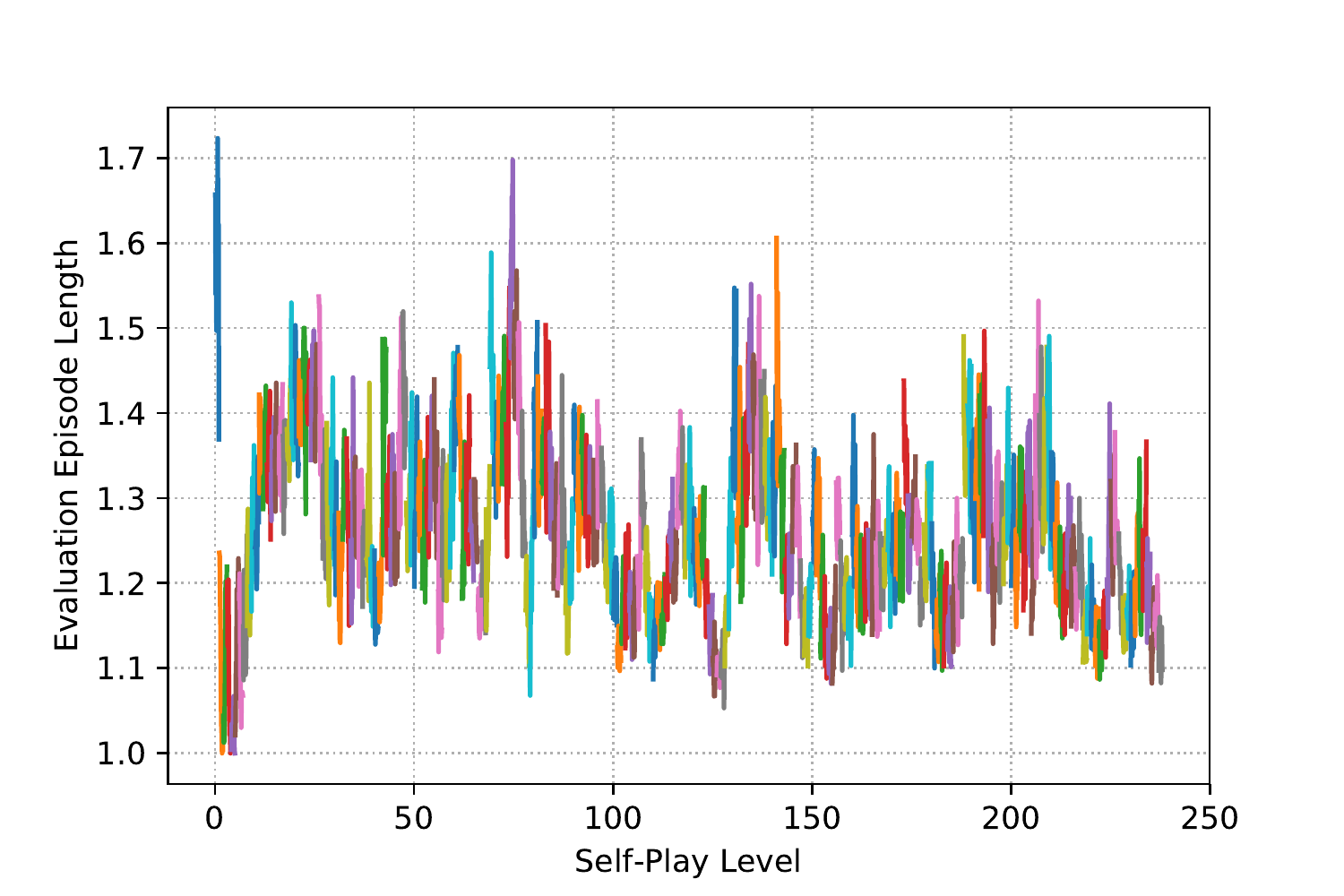}
  \mycaption{Self-Play Learning of Adversarial Hit-Ball Strategy.}{Mean evaluation episode length over 237 self-play levels.  Each colored line segment corresponds to a different level.  The vertical axis shows the average length of an adversarial episode in evaluation.  Each point is an average over 240 evaluation episodes.  The maximum episode length is 10.  The mean episode length very quickly approaches slightly above one, which signifies that the episodes end very quickly.  While at the beginning the episodes end because the robots cannot successfully return the ball, at some point they learn to quickly win the point on the first strike.}
  \label{fig:estrategy:hitballadv}
\end{figure}

\begin{figure}[htb!]
  \centering
  \includegraphics[width=0.9\columnwidth]{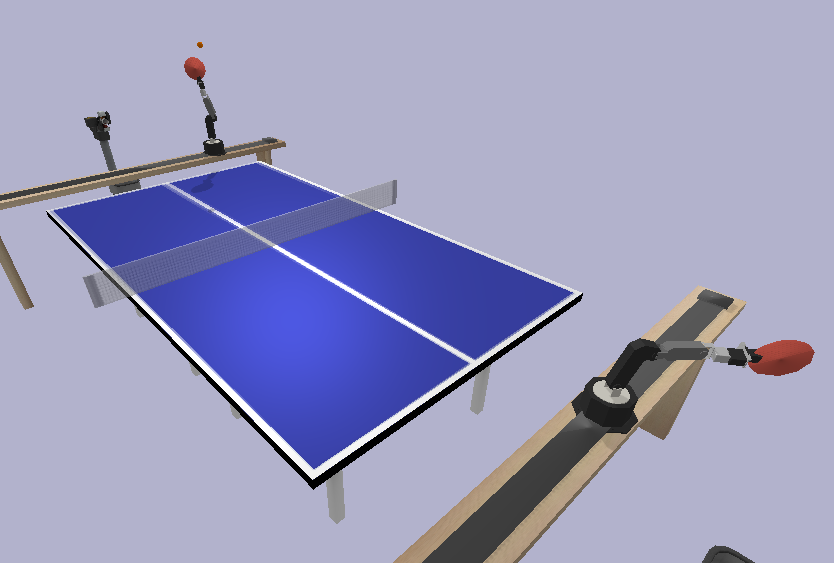}
  \mycaption{Visualization of Adversarial Hit-Ball Strategy.}{The strategy agent learns to win the rally by sending the ball over the opponent.}
  \label{fig:estrategy:hitballadvdemo}
\end{figure}

In contrast, \reffig{estrategy:hitballadv} shows the training progress for an adversarial hit-ball strategy.  Very quickly, the mean episode length drops to one, signifying that the adversarial games do not last long. \reffig{estrategy:hitballadv} shows the mean evaluation rewards received by the learning agent during adversarial training.  The mean episode reward reaches close to one at the beginning, which means the learning agent can easily exploit the opponent with the frozen policy.  With more training, the reward becomes more balanced.  Visualizing the trained strategy shows that the agents have discovered a way to send the ball over the head of the opponent.  In most episodes, the agent who gets to act first, hits the ball with enough vertical force that after bouncing on the table it cross above the reachable space of the other robot.  \reffig{estrategy:hitballadvdemo} visualizes and example of this behavior.

\begin{figure}[htb!]
  \centering
  \includegraphics[width=0.9\columnwidth]{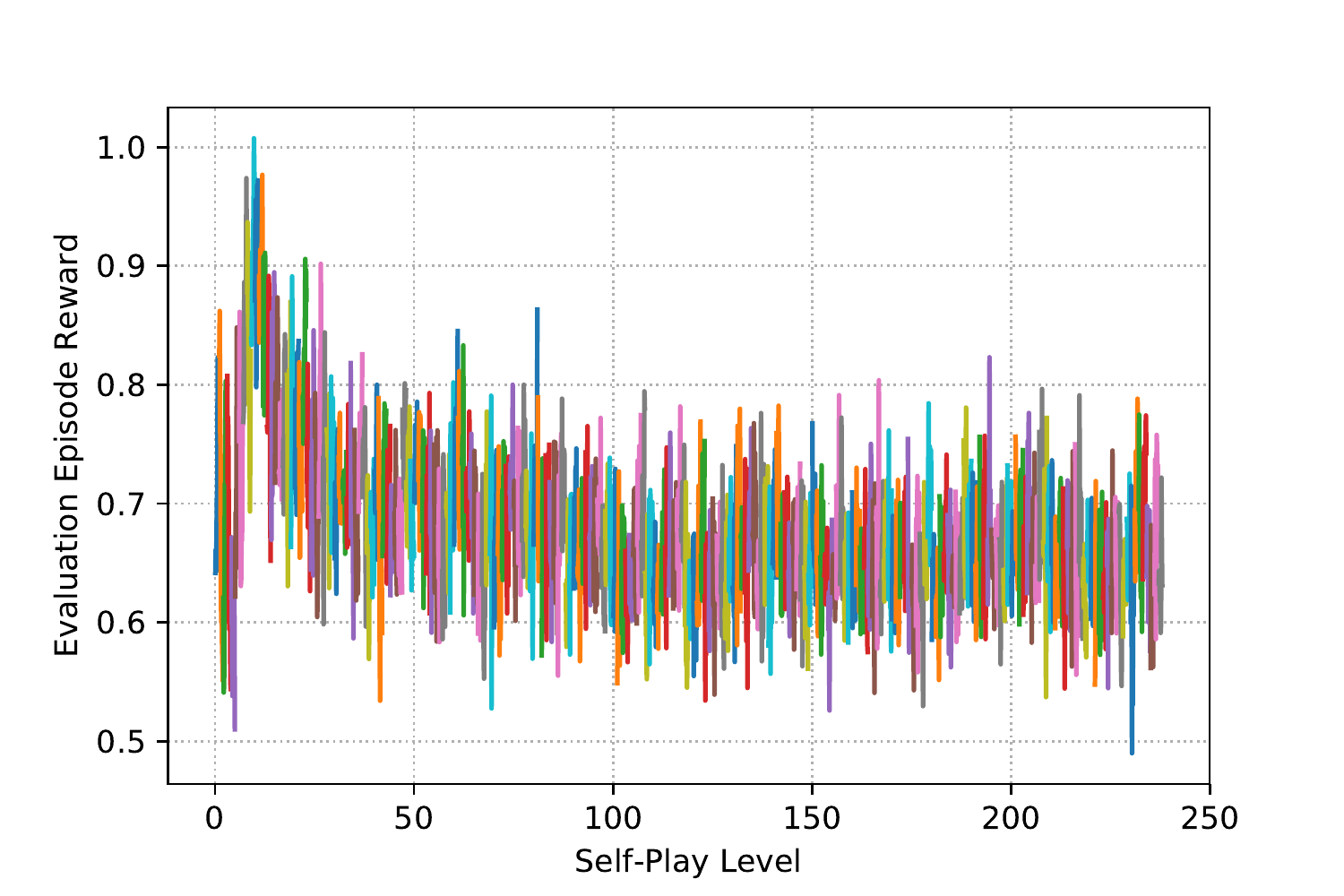}
  \mycaption{Average Evaluation Rewards for Adversarial Land-Ball Strategy.}{Mean evaluation reward over 237 self-play levels.  The mean episode reward reaches close to one at the beginning, which means the learning agent can easily exploit the opponent with the frozen policy.  With more training, the reward becomes more balanced.}
  \label{fig:estrategy:hitballadvscore}
\end{figure}

The experiments show that a hit-ball strategy can discover more effective cooperative and adversarial strategies at the expense of lower sample-efficiency.  The hit-ball skill supports executing arbitrary strikes and is as general as a striking policy can be.  It uses only a ball-trajectory prediction model that is easy to train.  The experiment in the next section evaluates the impact of using the ball-trajectory prediction model by training a strategy over a striking policy that does not use any models.


\subsection{Paddle-Control Strategy}
\label{sec:strategy:paddle-control}

As described in \refsec{skill}, there are three variants of the policy hierarchy with three different striking skills.  The previous two sections covered the land-ball and hit-ball striking skills.  The third is to use the paddle-control skill directly.  The difference between this striking policy and the hit-ball policy is that the agent is expected to specify all attributes of the paddle-motion state, including its target position $l(p_t)$ and the corresponding time $t$.  In other words, without access to the ball-trajectory prediction model in the hit-ball skill, this variant is expected to predict a position and time for incoming ball implicitly and aim for that spot with the paddle.

\begin{figure}[htb!]
\centering
\includegraphics[width=0.9\columnwidth]{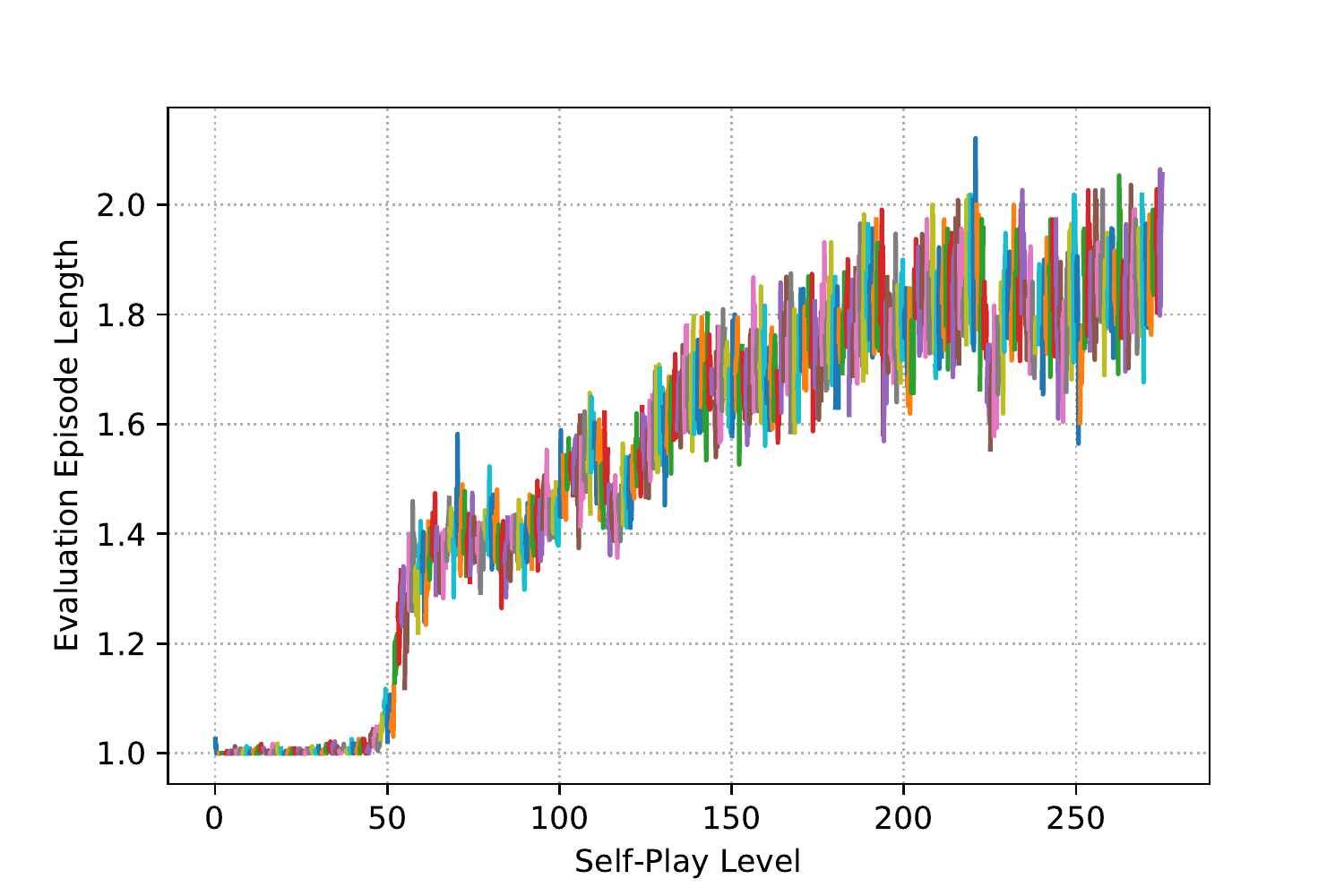}
\mycaption{Self-Play Learning of Cooperative Paddle-Control Strategy.}{Mean evaluation episode length over 274 self-play levels.  Each colored line segment corresponds to a different level.  The vertical axis shows the average length of a cooperative episode in evaluation.  Each point is an average over 240 evaluation episodes.  The maximum episode length is 10.   The training process for the paddle-control strategy is much slower than the land-ball and hit-ball strategies, since the agent needs to figure out how to make contact with the ball in the first place.}
\label{fig:estrategy:modelfreecoop}
\end{figure}

\begin{figure}[htb!]
\centering
\includegraphics[width=0.9\columnwidth]{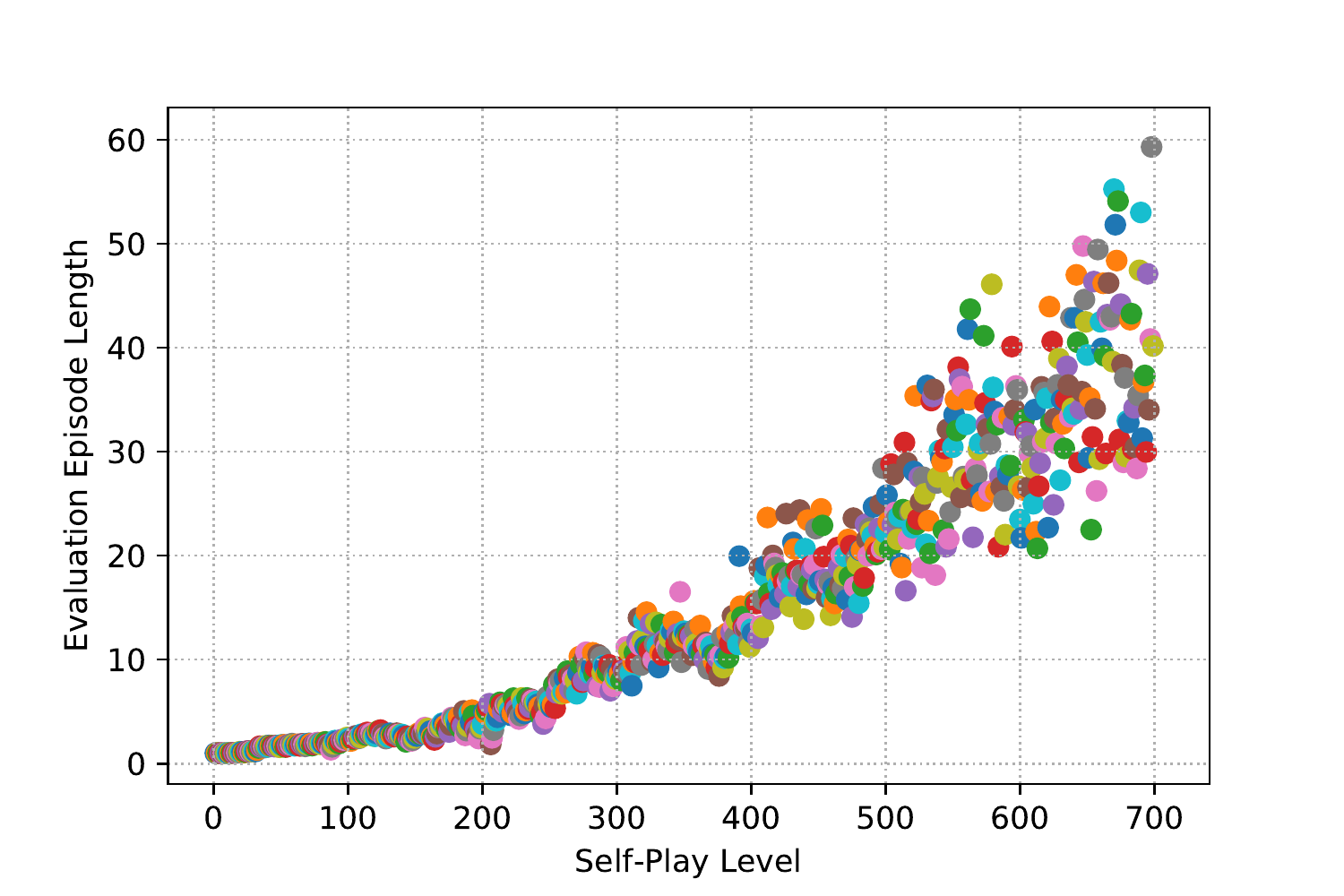}
\mycaption{Self-Play Learning of Cooperative Paddle-Control Strategy with More Training Time.}{Mean evaluation episode length over 698 self-play levels.  The vertical axis shows the average length of a cooperative episode in evaluation.  Each colored dot corresponds to a different self-play level and represents the mean over 2400 episodes.  Note that unlike the previous cooperative plots, here the maximum episode length is 100.  The training process for the paddle-control strategy is much slower than the land-ball and hit-ball strategies, since the agent needs to figure out how to make contact with the ball in the first place.}
\label{fig:estrategy:modelfreecoop100}
\end{figure}

\reffig{estrategy:modelfreecoop} shows the training progress for a cooperative paddle-control strategy.  After about 250 self-play levels, the mean episode length hardly reaches two.  The hit-ball strategy maxes out the episode length after this number of self-play levels.  This experiment was terminated early, but an earlier experiment that ran for many more self-play levels shows that given more time the model-free PPO algorithm is able to find better paddle-control strategies.  \reffig{estrategy:modelfreecoop100} shows the training progress for an experiment with lasted about 700 self-play levels, corresponding to about 1.7\,M training exchanges.  In this experiment, the episode cap was set to 100, instead of 10.  The strategy agent was able to achieve a mean episode length of about 40 and the performance continues to climb.

\reffig{estrategy:modelfreecoop} shows the training progress for an adversarial paddle-control strategy.  The mean episode length hardly rises above one, as was the case with the adversarial hit-ball policy as well.  However, with the hit-ball policy, the episodes ended quickly because after some amount of training either robot could win the rally in one shot.  The paddle-control strategy has a hard time learning to make contact with the ball, so, the side that has to act first has a high chance of losing the rally by failing the return the ball.  Although it is possible that with more training the adversarial paddle-control strategy could improve, there is no advantage to using the paddle-control strategy over the hit-ball strategy, as they both have the same action space.

\begin{figure}[htb!]
\centering
\includegraphics[width=0.9\columnwidth]{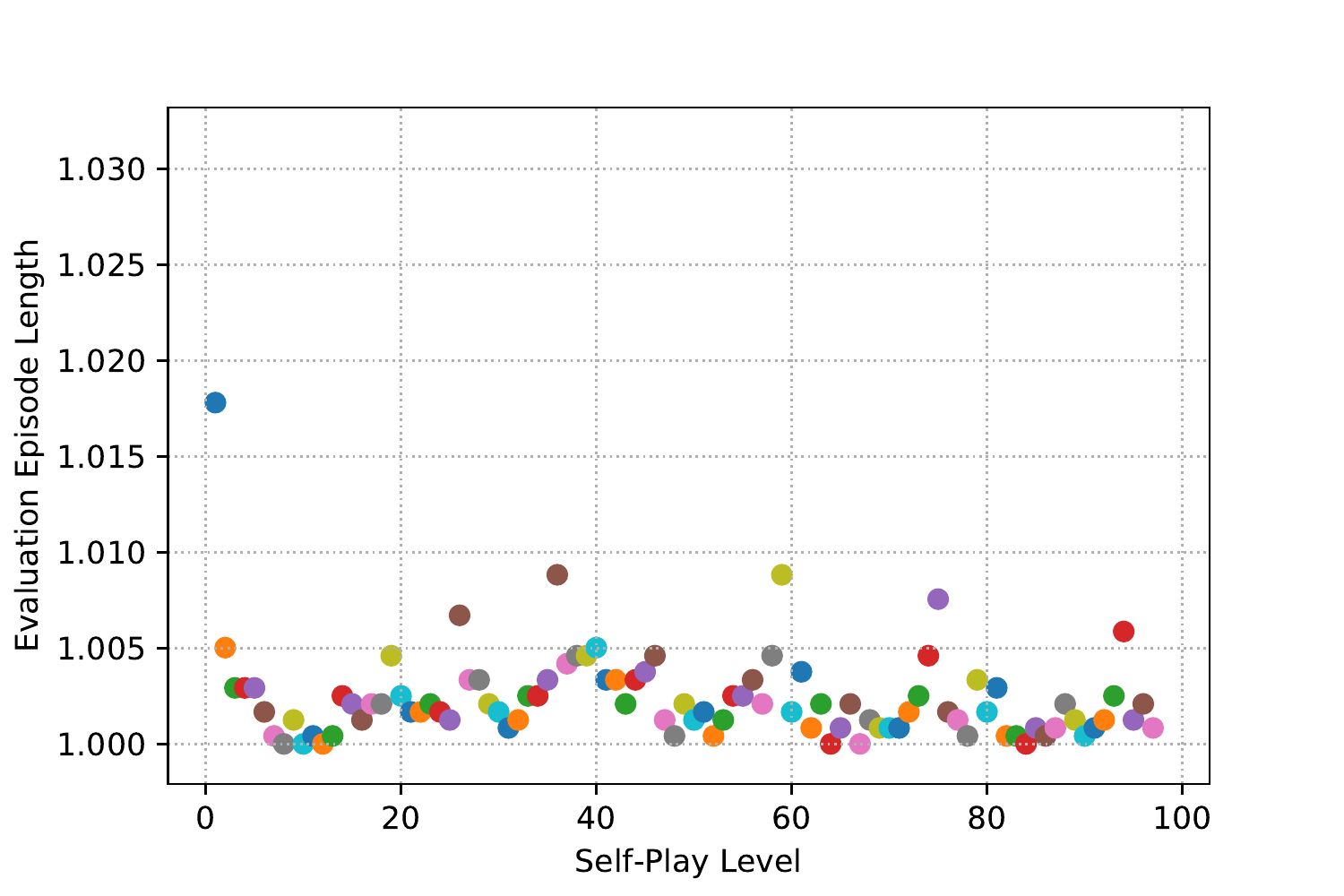}
\mycaption{Self-Play Learning of Adversarial Paddle-Control Strategy.}{Mean evaluation episode length over 96 self-play levels.  The vertical axis shows the average length of a cooperative episode in evaluation.  Each colored dot corresponds to a different self-play level and represents the mean over 2400 episodes.  The mean episode length hardly goes above one, since neither the learning agent nor the opponent with the frozen policy can learn to make contact with the ball and return it successfully with this much training. In these games, the first robot which gets to act usually loses the point.}
\label{fig:estrategy:modelfreeadv}
\end{figure}

The experiments in this section show that the simple dynamics model that predicts the ball's future trajectory can improve the effectiveness and sample-efficiency of cooperative and adversarial strategies dramatically.

\subsection{Joint-Control Strategy}
\label{sec:strategy:joint-control}

Learning a strategy over the joint-control skill requires the agent to specify joint velocity targets on every timestep of the environment.  A typical ball exchange lasts 70-100 timesteps.  A joint-control strategy needs to execute successful strikes just in time as the ball is reaching the robot, and then properly position the robot to be ready for the next strike.  So, in effect, it needs to execute two complementary skills.

\begin{figure}[htb!]
\centering
\includegraphics[width=0.7\linewidth]{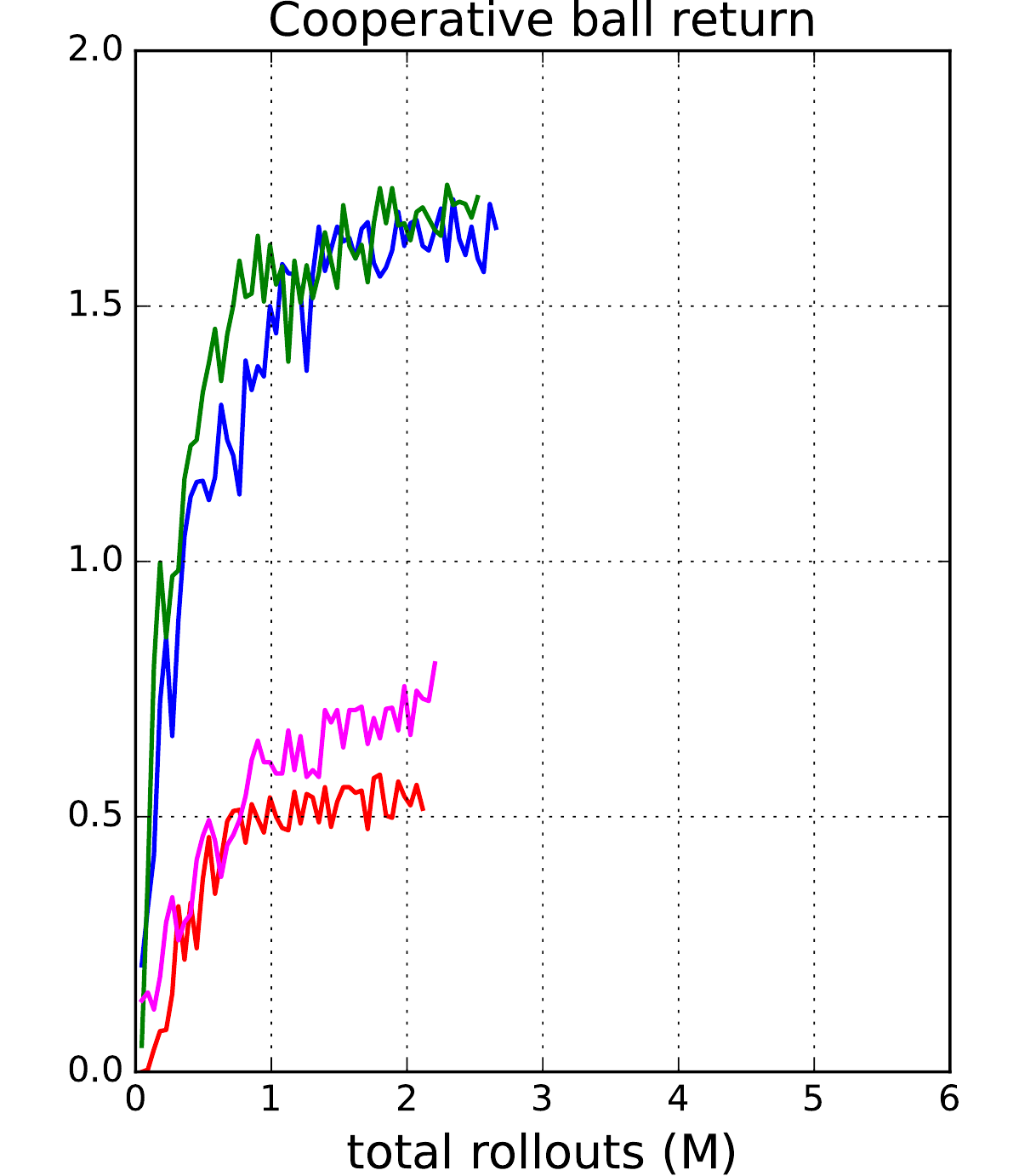}
\mycaption{Self-Play Learning of Model-Free Joint-Control Strategy with ARS.}{Cooperative scores for four randomly-initialized runs.  The vertical axis shows the number of successfully-returned balls per rally.  In all four runs the number of evaluations $k$ per policy perturbation was set to 15.  Training the model-free joint-control strategy is less effective and less sample-efficient than other strategy skills.}
\label{fig:estrategy:ars_coop}
\end{figure}

\reffig{estrategy:ars_coop} shows the results for training an ARS agent to play cooperative games against itself.  The ARS policies converge after about 2M episodes.  The best policy achieved a score of 2.0 (both arms return the ball) 87.1\% of the time, and a score of 1.0 about 0.1\% of the time.  While the learned strategy was cooperative in the sense that the second robot almost always succeeded in returning the ball, this behavior did not extend to longer plays, \ie the first robot could not return the ball after the second robot hit it.  Longer cooperative play could possibly be achieved by allowing longer rollouts at training time, as well as increasing the randomness in the environment initial state (position and velocity of the ball and robots). 

Training the model-free joint-control strategy is more difficult than training the model-free paddle-control strategy discussed in \refsec{strategy:paddle-control}.  While the paddle-control strategy needed many training episodes, it could gradually improve the cooperative rewards and achieve cooperative episode lengths equal to 40\% of the maximum possible after less than 2M training episodes.  The rewards achieved by the model-free joint-control policy are lower by comparison.  This difference can be attributed to the high-level paddle actions that allow the paddle-control strategy to execute effective paddle strikes just by specifying a target motion state and time for the paddle.  The long time horizon of these high-level actions make learning easier by reducing the delay in receiving the reward for the chosen action.


\subsection{Conclusion}

This section discussed training and evaluation of different variants of the strategy skill operating over various striking policies including land-ball, hit-ball, paddle-control, and joint-control.  The hierarchical policy design used in this article allows the table-tennis agent to combine model-based, model-free, and analytic skills in a single policy that can effectively learn cooperative and adversarial games with good sample-efficiency.

Using model-based striking policies greatly increases the sample-efficiency of self-play learning.  On the other hand, allowing the strategy skill to execute arbitrary strikes with the hit-ball skill results in discovering novel strikes suitable for cooperative and adversarial games.  Self-play strategies over completely model-free skills like paddle-control and joint-control require more more training episode and are less effective at the task.

The experiment results suggest that model-based learning is a good way to increase sample-efficiency of learning robotic tasks, especially when the aim is to learn tasks directly in the real world.  It is possible to fine-tune the landing models trained from human demonstrations with data collected from other striking policies, as done in the dagger method~\cite{ross2011reduction}.  Such an approach can combine the sample-efficiency of model-based learning with flexibility of model-free policies.




\section{Discussion and Future Work}
\label{sec:disc}

%

%

%

%


This article presents a method for learning the technical and tactical dimensions of robotic table tennis in a virtual reality environment.  The technical aspects of the game are learned from realistic human demonstrations, and are executed using a robot-agnostic analytic controller that can execute timed motion targets for the paddle including its pose and linear and angular velocities.  The tactical aspect of the game is captured in the strategy skill and trained using model-free RL efficiently.  This section evaluates the results presented in this article and outlines steps for future work.

\subsection{Development Process}
\label{sec:disc:process}

The work reported in the literature often focuses on results from the most successful experiments and the best-performing agents.  However, it is not always clear how much effort goes into training those agents.  In some cases, extensive hyperparameter tuning and experiments with multiple random seeds are needed to train high-performing agents.  Such a development process is not feasible in robotic domains, where each training episode carries a significant cost in terms of development time and wear and tear on the hardware.

In an end-to-end setup, the impact of a change in the implementation can be evaluated only by its impact on the overall performance of the agent.  Often, learning algorithms are able to function moderately well even with bugs, making it difficult to observe the negative impact of a new bug immediately after it has been introduced.  The hierarchical policy used in this article makes the development and debugging of individual skills easier.  Since the skills can be evaluated separately, the failure cases and the root causes for performance issues can be identified and studied.  Moreover, it is possible to give the policies perfect observations and actions to accurately measure their intrinsic errors.

The high-level skills like striking and positioning in the hierarchical policy give the top-level strategy skill access to neutral high-level behaviors that it can exploit in different ways to discover new ways of playing table tennis.  With this design, the application of model-free RL can be localized to the strategy skill.  Since the high-level skills are robust and effective, a greater percentage of the action space of the strategy skill consists of useful actions.  So, during exploration, the RL algorithm is less likely to waste its effort on trying actions that are not useful.

Since the underlying skills are robust and action space for the strategy skill is effective, general-purpose RL algorithms like PPO can discover effective policies using only about 24,000 training episodes.  All hierarchical reinforcement learning experiments in this article were carried out on one or two local workstations, without hyperparameter tuning.  This is a huge improvement over standard practice in model-free RL.


\subsection{Driving Different Robot Assemblies}

Decomposing the environment into a game space and a robot space makes it possible to learn the game dynamics independently of how robot table tennis is to be played.  If the robot assembly is replaced, the game dynamics models can be used with the new robot without retraining.

The analytic paddle controller is also agnostic to the robot setup.  Given the motion constraints for the robot, it can optimally drive any robot assembly to execute motion targets for the paddle using the Reflexxes library.  The analytic controller treats a multi-robot setup (for example, a linear actuator plus an arm) as a single assembly, which allows it to control complex robot setups.

The action spaces for the striking and positioning skills work with paddle positions.  This choice allows the same implementation for these skills to work with different robot assemblies.  However, the strategy policies learned on a particular robot assembly would not transfer to a different robot assembly.

Replacing the robot assembly requires updating the action spaces to reflect the area of the space that is reachable by the robot.  Also, the canonical forehand and backhand poses that are used to initialize the IK search need to be updated (see \refsec{paddle:positioning}).

Thus the approach is quite general and independent of the robot, which makes it possible to transfer it to the hardware without much further work on the controller.


\subsection{Observation Uncertainty}
\label{sec:disc:noise}

The simulator used in this article (PyBullet) has no built-in observation and action noise.  As discussed in \refsec{disc:process} it was beneficial to train the agents with no observation uncertainty to study the best possible performance of the dynamics models and the training policies and gain better insight into any factors that contribute to errors in prediction or behavior.  However, deploying this method in the real-world requires working with observation uncertainty.

Even though the underlying simulation environment does not have any observation noise, the method is designed with observation noise in mind.  The Reflexxes library was chosen in part because of its ability to react to sensory noise.  Reflexxes can plan to reach a target motion state given any current motion state.  So, if the paddle target is changed due to updated observations on the trajectory of the ball, Reflexxes can recompute a trajectory to reach a new target from the current motion state.

The land-ball algorithm discussed in \refsec{landball:modelbased} is an open-loop algorithm.  Working with noisy observations requires implementing closed-loop control policies.  Experiments are needed to evaluate and measure any loss of performance due to observation uncertainty.  Noisy observations would make the task more difficult.  On the other hand, a closed-loop controller can improve performance by continuously updating the predicted trajectory of the incoming ball.  In the current implementation, this  trajectory is predicted as soon as the ball starts traveling in the direction of the robot.  In other words, the striking policy computes -- and never updates -- a prediction for the position and velocity of the ball when it reaches the robot.  This prediction is computed when the ball is on the other side of the table.  Closed-loop policies may obtain more accurate predictions when the ball gets closer to the robot.  This increase in prediction accuracy can improve the landing accuracy.

\begin{figure}[htb!]
\centering
\includegraphics[width=0.9\columnwidth]{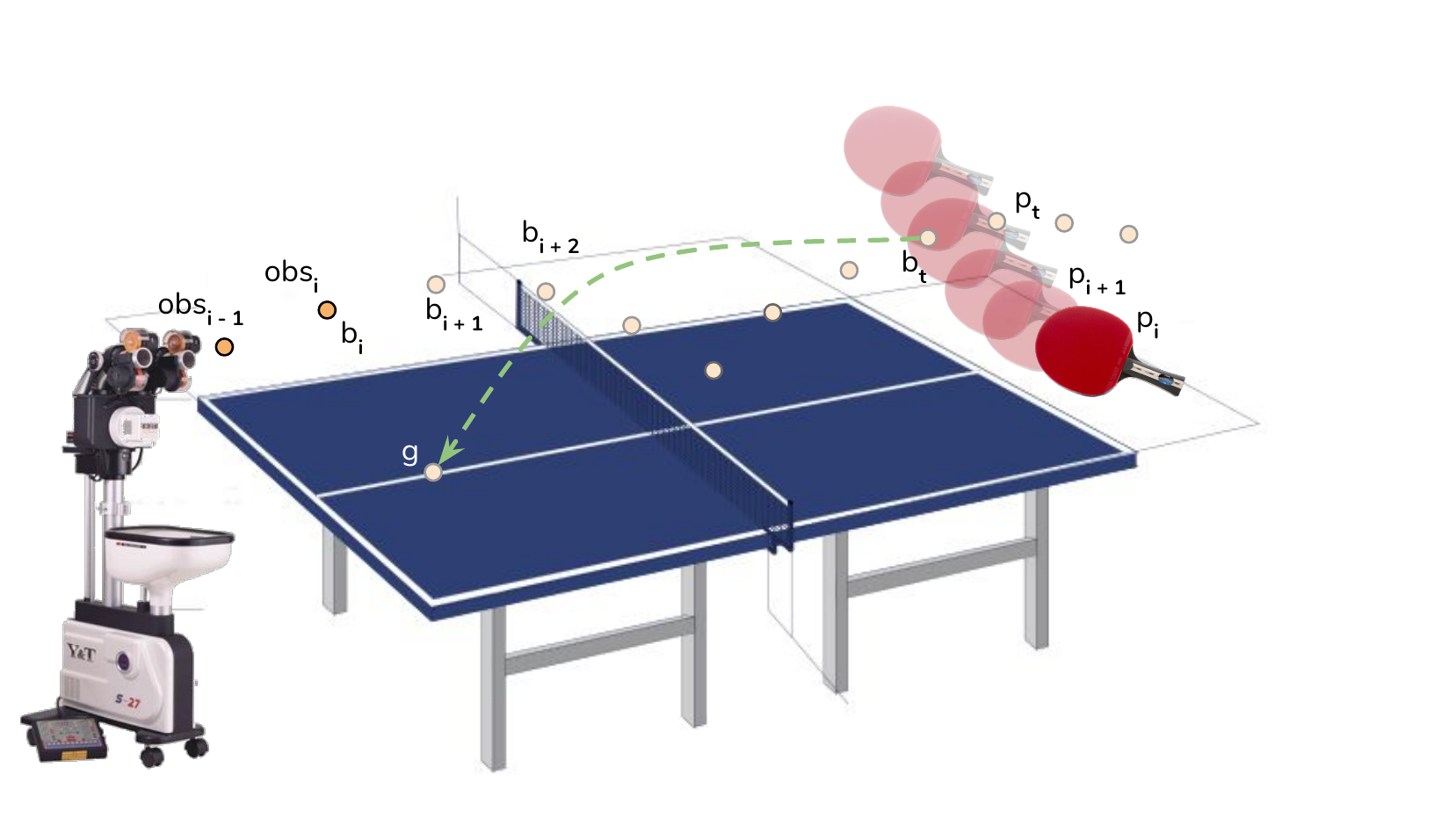}
\mycaption{The Game Environment with Observation Noise.}{At every timestep $i$, the environment exposes an observation obs$_i$ of the state of the ball.  The observations could be the ball position and velocity estimates coming from a ball tracker, or a set of two-dimensional coordinates for the ball blob from multiple cameras.  The current and past observations can be used to estimate the true state of the ball $b_i$.  This setup makes it possible to use the developed approach in environments with observation noise.}
\label{fig:noise:env}
\end{figure}

\reffig{noise:env} shows the environment with observation noise included.  At each timestep $i$, the environment exposes some observations $\text{obs}_i$ of the ball.  The observations could be the position and velocity estimates from a ball-tracking algorithm, or the raw blob locations for the ball from multiple cameras.  The policy needs to estimate the true motion state of the ball from these observations.  Such an estimate can be obtained from a model described in \refsec{disc:noise:model}.  Since the ball state estimate keeps changing after receiving new observations, \refsec{disc:noise:policy} describes how the striking policies like land-ball can be adjusted to update their predictions and targets during the course of a paddle swing.

\subsubsection{Ball-State Estimation Model}
\label{sec:disc:noise:model}

The ball-state estimation model receives a sequence of noisy observations of the ball and estimates the ball's current position and velocity.  It is described by the function

\begin{align}
b_0, b_1, \dots, b_s = S(\text{obs}_0, \text{obs}_1, \dots, \text{obs}_{s}),
\end{align}
\\
where $S$ denotes the ball-state estimation model, $\text{obs}_s$ denotes a noisy position observation on the ball obtained at timestep $s$, and $b_s$ denotes the estimate on the ball's motion state at time $s$.

\begin{figure}[htb!]
  \centering
  \includegraphics[width=0.9\columnwidth]{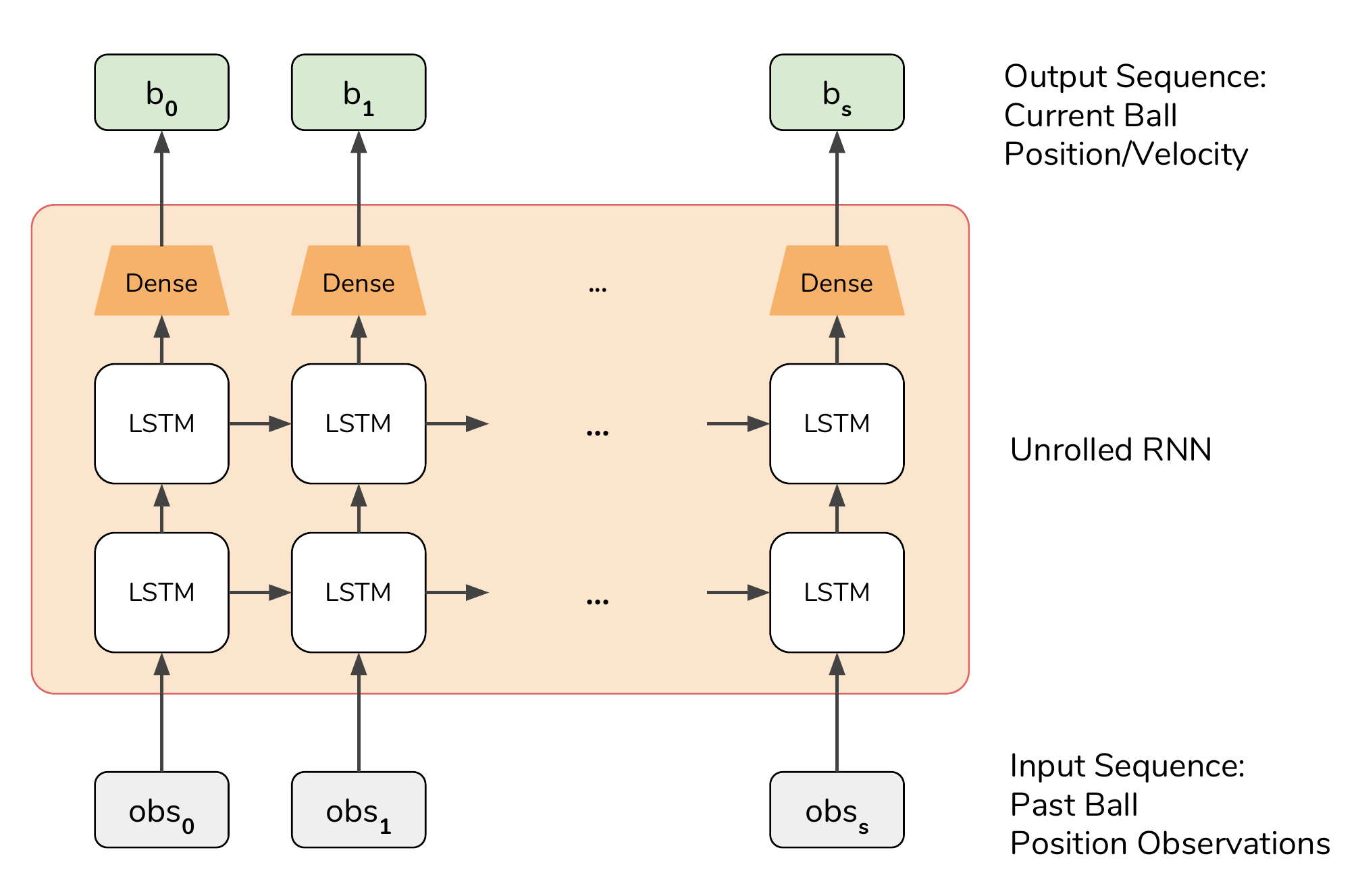}
  \mycaption{Ball-State Estimation Model.}{The current ball-motion state is estimated using a sequence model with LSTM layers followed by a fully-connected layer.  The figure shows the model unrolled through time for better visualization.  This model makes it possible to estimate the current position and velocity of the ball given a sequence of noisy position observations.}
  \label{fig:noise:ball1}
\end{figure}

\reffig{noise:ball1} shows the network architecture for this model.  It is a recurrent model with two LSTM layers followed by a fully-connected layer.  At each timestep, the model is given a new noisy position observation and it produces an updated estimate on the ball's current position and velocity.  Over time, with more observations, this model can produce estimates that get closer to the true state of the ball.

The agent maintains one instance of the model.  As the episode progresses, on each new timestep, the model is updated with a new observation from the current timestep and produces an updated estimate on the current ball-motion state.  The estimate obtained from this model can be used to generate predictions about the future states of the ball using the ball trajectory prediction model as explained in the next section.


\subsubsection{Closed-Loop Land-Ball Skill}
\label{sec:disc:noise:policy}

Algorithm \ref{alg:landball2} shows the updated land-ball algorithm that works in an environment with noisy observations.  Instead of computing a target paddle-motion state $p_t$ only once, this algorithm recomputes the paddle target on every timestep.  The algorithm starts by using the ball observations to update its current estimate of the ball-motion state $b_s$.  It then predicts the ball trajectory $T$, and computes a paddle-motion state $p_t$ to hit the ball toward target $g$.  However, instead of following $p_t$ using an open-loop controller, the algorithm follows $p_t$ just for one timestep.  On the next timestep, $p_t$ is updated based on a new ball observation, and the paddle-control skill is requested to adjust its target to the updated $p_t$.  Since Reflexxes is able to adapt to moving targets, the land-ball algorithm works with observation noise as well.

\begin{algorithm}[htb!]
  \SetKwInOut{Input}{inputs}
  \SetKwInOut{Output}{output}
  \Input{Sequence of noisy ball observations $\{{obs}_0, \text{obs}_1, \dots\}$}
  \Input{Current timestep $s$}
  \Input{Desired landing target $g$}
  $i \gets s$\;
  \Repeat{robot paddle hits the ball or episode ends}{
    read latest ball observation $obs_i$\;
    $b_i \gets S(\{\{\text{obs}_0, \text{obs}_1, \dots, \text{obs}_i\}\})$\;
    $T = b_{i+1}, b_{i+2}, \dots, b_n \gets B(b_i)$\;
    \ForEach{$b_k \in T$ such that $b_k$ is reachable}{%
      $p_k \gets L^{-1}(b_k, g)$\;
      $\hat{p}_k \gets P(p_k, p_s)$\;
      $\hat{g}_k \gets L(\hat{p}_k, b_k)$\;
    }
    $t \gets \argmin_k ||\hat{g}_k - g||$\;
    emit first action $u_i$ from $\pi_p(t, p_t \mid p_s)$\;
    $i \gets i + 1$\;
  }
  \caption{Closed-Loop Land-Ball Skill Algorithm}
  \label{alg:landball2}
\end{algorithm}

Some heuristics are needed to avoid potential failure cases.  For example, when the paddle is close to the target, if Reflexxes is moving the paddle with maximum velocity or acceleration, a small update to the target make render it infeasible.  In such cases, the update should be ignored, opting for a small error, rather than a failure.

This extension takes a major step towards taking this system to real hardware.  Modeling ball spin and using vision to estimate it are other such extensions, which are discussed in subsequent sections.


\subsection{Ball Spin and Magnus Forces}

The PyBullet simulator does not simulate Magnus forces that cause a spinning object to curve its trajectory in the air.  In the current implementation, the ball's spin only affects its motion after contact due to the friction forces acting on the ball.

The method can be extended to factor in the ball's spin.  The ball-trajectory prediction model can be extended to receive an estimate on the ball's spin as input.  As experiments in \refsec{dyn:e} show, ball-trajectory prediction is a simple problem and the model is expected to have the capacity to handle ball spin as well.  To make it easier to predict the future states of a spinning ball, the dynamics models discussed in \refsec{dyn} can be augmented with physics models that capture Newtonian motion, air friction, and Magnus forces.  The parameters in the physics models can be fitted to observed trajectories.  Once fitted, the physics models can be used to establish initial estimates of the future states of the ball.  Then the neural network can predict only a residual over the estimates from the physics models, which is a simpler prediction problem.  The next section discusses how the ball's spin can be estimated in real-world environments.


\subsection{Vision}

The method advocates working with low-dimensional state over raw visual input.  The primary element in the environment that needs to be observed with vision is the ball.  A variety of ball tracking algorithms exist that can compute estimates on the position and velocity of a moving ball.  Detecting the ball in a camera image is a relatively simple computer vision task which does not require expensive computations.  On the other hand, using neural networks to track the ball using raw visual input requires carefully varying the lighting conditions and backgrounds to avoid overfitting the model to the training environment.

The only situation that requires handling raw visual input is estimating the impact of a human opponent's paddle motion on the ball.  Since the opponent paddle is not instrumented, it would be useful to establish an estimate on its motion at the moment of contact with the ball.  Such an estimate can be used to help the ball tracking algorithm predict an expected trajectory for the ball.  It can also be used as an input to the ball-state estimation model shown in \reffig{noise:ball1}.

Establishing a prior on the motion of the ball as a result of contact with the opponent paddle is specially useful for estimating the ball's spin.  While the ball position and velocity can be estimated using a few observations, the ball's spin is harder to estimate.  However, during the game the ball always starts with no spin in the player's hand.  Physics models can be used to establish an estimate on the ball spin as a result of contact with the human or robot paddles.  With these extensions, the method can be deployed in the real world.


\subsection{Hardware Evaluation}
\label{sec:disc:hardware}

Experiments in \refsec{striking} show that only about 7,000 human demonstrations can be used to train a land-ball policy that can hit ball targets with about 20\,cm error.  Similarly, experiments in \refsec{strategy} show that only about 24,000 self-play exchanges on the robot can be used to learn a strategy that can sustain rallies lasting 14-16 hits.    So, these experiment suggest that the method may work in the real world as well.

The virtual reality environment developed in this article has many similarities with a real-world table-tennis environment.  Most notably, the same motion sensors used for tracking the paddle in the VR environment can be used to track human paddles in the real world.  The sensory readings from VR trackers are similar to readings from any other motion tracking system.  The data collection setup used in this article already deals with many problems arising from working with physical sensors.  For example, there is often jitter present in location readings coming from the VR tracker, which makes estimating the true position of the VR paddle difficult.  Also, the VR trackers do not provide any velocity information.  The data collection program estimates the linear and angular velocity from the position readings.  Lastly, HTC VR trackers work at 90\,Hz, while the simulator works at 1000\,Hz.  The difference in frequency requires producing intermediate estimates for the state of the tracker.

The VR environment can be regarded as an implementation between a simulation and the real world.  Although the sensory readings coming from VR trackers are similar to sensory readings that would be available in the real world from a motion tracking system, deploying the method in the real world requires obtaining the ball state from vision using a ball-tracking algorithm.  As outlined in \refsec{disc:noise}, the striking policies need to be updated to work in a closed-loop manner and update their targets based on new ball observations.  Additionally, deploying the method in the real world requires training a residual dynamics model for the paddle-control policy to capture control inaccuracies due to imperfect robot hardware and robot controllers.

The experiments carried out in this article show that a robotic table-tennis agent can be trained using only a few thousand human demonstrations and about 20 thousand  exchanges in robot games.  This high sample-efficiency suggests that the method can work in the real world as well.  It is likely that training the models and policies in the real world requires dealing with additional challenges that would reduce the sample-efficiency or performance of the method.  However, there are also opportunities for increasing the sample-efficiency or performance, \eg by augmenting demonstrated trajectories using height reduction (\refsec{dyn}), augmenting dynamics models with physics models (\refsec{dyn:physics}), using closed-loop controllers (\refsec{disc:noise:policy}), and mixing dynamics models with model-free policies as discussed in the next section.


\subsection{Mixing Dynamics Models with Model-Free Policies}

Dynamics models greatly improve the sample-efficiency of the agent.  Only about 7,000 demonstrated strikes were enough to train a striking policy with a mean target error of about 20\,cm.  Similarly, only about 24,000 self-play exchanges were enough to train a cooperative land-ball strategy sustaining rallies lasting about eight exchanges on average.  These results show that model-based learning requires much fewer samples.  On the other hand, the dynamics models that were trained from human demonstrations were never updated during robot experiments.  Therefore, the land-ball striking policy that uses these dynamics models stays limited to strikes that were demonstrated by humans.

The hit-ball strategy experiments in \refsec{strategy:hit-ball} show that more flexible striking skills can discover striking motions beyond what is observed in demonstrations.  On the other hand, since the hit-ball policy uses only one dynamics model for ball-trajectory prediction, and its action space has more dimensions compared to the entirely model-based land-ball skill, training a strategy over hit-ball is less sample-efficient.

The method can be extended to allow updating the dynamics models with data collected from more flexible policies that are either model-free or use fewer dynamics models.  Doing so would increases the predictive ability of the dynamics models to cover a wider range of behaviors in the environment.

On the other hand, rather than learning the skills from scratch, the model-free policies can be modified to use the predictions from the models as a starting point for their decisions.  Doing so allows the model-free policies to expand the boundary of explored behaviors gradually by trying out variations slightly outside the space of behaviors captured in the models.  Such guided exploration would make of the model-free policies more sample-efficient.


\subsection{Conclusion}

This article demonstrated sample-efficient learning of technical and tactical aspects of robotic table tennis.  Since the experiments were carried out in a VR environment with real-world motion sensors, the results suggest that the method can be used to learn the task in the real world without relying on transfer of models or policies from simulation.  The ability of Reflexxes to adjust the trajectory target would dynamically allow the method to work with noisy ball position and velocity estimates coming from a ball-tracker in the real world.

\section{Related Work}
\label{sec:related-work}


\subsection{Robot Table Tennis}

There has been a broad body of work on robot table tennis~\cite{andersson1988robot, angel2005robotenis, miyazaki2006learning}.  More recent approaches by Muelling \ea~\cite{mulling2011biomimetic, muelling2013modeling, mulling2013learning, muelling2014learning, wang2017anticipatory} are similar to the approach in this article in that they use human demonstrations to seed the robot's behavior and they decompose the task into high-level and low-level skills.

Inspired by studies of humans playing table tennis, Muelling \ea~\cite{mulling2011biomimetic} break up the game into four different stages -- awaiting, preparation, hitting, and finishing.  In the hitting stage, they use an inverse dynamics model to compute the parameters of the robot at time of contact to achieve a desired target location, similar to the land-ball skill presented in this article.  Muelling \ea also decompose the task of playing table tennis into subtasks, which is similar to this work.  They implement the low-level control of the robot by learning a set of hitting movements from human-guided kinesthetic demonstrations on the robot~\cite{mulling2013learning}.  These movements are compiled into a library of motor primitives, which are then expanded by mixing existing movements to create new motor primitives.  They then employ inverse reinforcement learning (IRL) over human demonstrations to learn a high-level strategy for the full game~\cite{muelling2014learning}.

Rather than using kinesthetic teaching, the approach in this article captures human strikes using instrumented paddles in free-form human games.  This choice avoids limiting the human's movements to what is possible with kinesthetic teaching.  Instead of representing hitting movements at the joint level, the method in this article abstracts strikes by the paddle-motion state that is in effect at the time of contact.  This high-level representation makes it possible to solve the motor control problem using the general-purpose Reflexxes~\cite{kroger2011opening} trajectory planning algorithm.  Since the paddle-motion state is not coupled to any specific robot, this approach can execute the learned strikes on any robot without a need for new demonstrations.

Muelling \ea~\cite{muelling2014learning} use human demonstrations to learn a reward function for the game, which represents a high-level game-play strategy.  The method in this article uses human demonstrations only to learn the game dynamics, which gives the policy the freedom to discover more diverse game-playing strategies beyond what is demonstrated by humans.  Using self-play in robot vs. robot games makes it possible to to freely explore and refine game-play strategies.  Also, the method proposed in this work is hierarchical and assembles both high-level and low-level skills into one overall system, which can be refined end-to-end if needed.  This is while Muelling \ea~\cite{muelling2013modeling} train the different components of the hierarchy as separate models.


\subsection{Model-Based and Model-Free Learning}

The approach in this article decomposes the problem into parts that are model-based, such as dynamics models of the ball motion, and of contact, and parts that are model-free, such as learning a high-level strategy. This decomposition breaks down the problem in a way that is consistent with the strengths of these different approaches. The model-based parts are trained with supervised models and it is easy to generate data for them.  Meanwhile, using model-free techniques for training the strategy allows much better exploration without being very expensive, since only a few dimensions are optimized in the strategy.  The work by Muelling \ea~\cite{muelling2013modeling, mulling2011biomimetic} follows a similar decomposition overall, but high-level strategies are not trained with model-free approaches.  Instead, the strategy is either hard-coded~\cite{muelling2013modeling} or learned from demonstrations using inverse reinforcement learning techniques~\cite{mulling2011biomimetic}.  In this article, the high sample-efficiency makes it possible to employ model-free techniques for higher-level skills, which in turn leads to discovering novel strikes and game-play strategies.


\subsection{Hierarchical Reinforcement Learning}

Hierarchical reinforcement learning is a long-established discipline.  From the original options framework~\cite{sutton1999between, precup2000temporal} to the more recent approaches, like FeuDAL networks~\cite{dayan1993feudal, vezhnevets2017feudal, nachum2018data}, there is a broad body of work on training layered policies where each skill sets goals for other skills below it.  These approaches share the same underlying goal -- namely, to learn policies with a reduced number of timesteps by using models that operate at different rates.  In FeuDAL networks the policy consists of two task levels that are trained together, such that the higher-level task receives rewards over the actual objective in the environment, while the lower-level task receives rewards for achieving targets set by the higher-level task.  A common motivation for hierarchical reinforcement learning methods is automated discovery of useful subtasks.  However, in this article the main motivation is developing a specific task hierarchy that is most suitable for learning robotic tasks with extremely high sample-efficiency.  In this work, the targets set by the higher-level skills for the lower-level skills have specific semantics; they either signify a desired outcome in the environment, such as a ball target location, or indicate a desired state for the robot, such as a target paddle-motion state at the time of contact.  Such intermediate targets are effective at reducing the dimensionality of the control problem while allowing for incorporating human demonstrations in a robot-agnostic controller.


\subsection{Self-Play Learning}

Self-play strategies have been applied extensively in fully-observable games such as Go and chess~\cite{silver2017mastering}.  Lately, these have also been applied to partially-observed games such as StarCraft~\cite{vinyals2017starcraft}.  Self-play has not previously been applied to continuous-control tasks such as robotic table tennis, mainly because existing self-play techniques require a prohibitive number of training episodes, which is not feasible for robotic domains.  The hierarchical policy and the action representations developed in this work greatly reduce the dimensionality of the higher-level skills like the strategy skill, thereby reducing the number of self-play episodes needed.  Effective application of self-play learning makes it possible to learn more diverse striking motions and cooperative and adversarial table-tennis strategies than would be possible through learning from demonstrations or inverse reinforcement learning.


\subsection{Underlying Methods}

The following libraries and algorithms are used in developing the method in this work.

\subsubsection{Reflexxes}

\begin{figure}[H]
\centering
\includegraphics[width=0.8\linewidth]{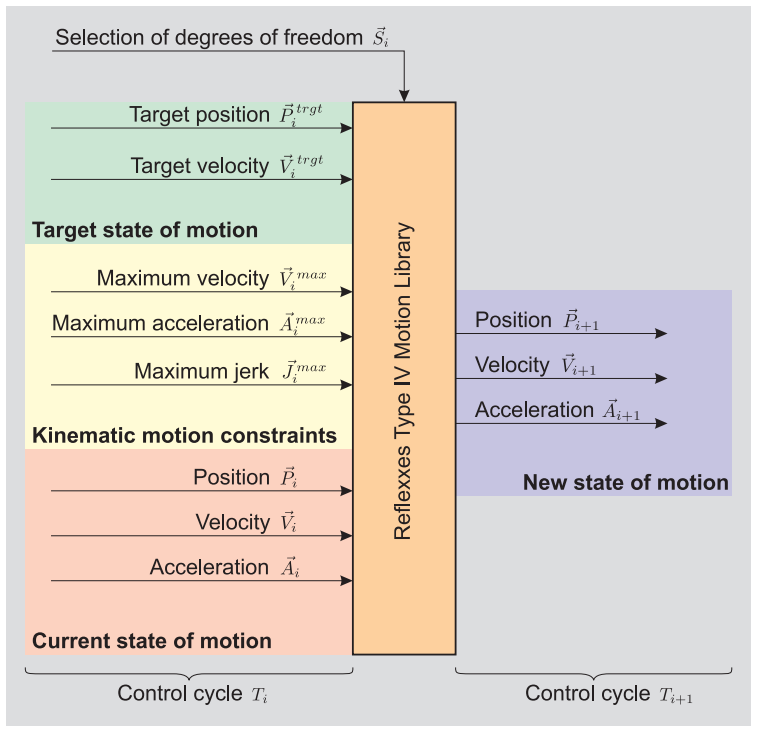}
\mycaption{The Interface to Reflexxes Motion Libraries.}{At each control cycle, Reflexxes receives the current state of motion (including position, velocity, and acceleration), the target state of motion (including position and velocity), and kinematic motions constraints (including velocity, acceleration, and jerk).  Given these inputs, Reflexxes produces the new state of motion (including position, velocity, and acceleration) for the next control cycle.  The new state of motion is computed using a fast deterministic algorithm, and is guaranteed to be on a time-optimal trajectory toward the target state of motion.  Image source: Kr{\"o}ger\cite{kroger2011opening}.}
\label{fig:related-work:reflexxes}
\end{figure}

Reflexxes~\cite{kroger2011opening} is an online trajectory planning algorithm that is originally designed to allow robots to react to unforeseen events.  It is capable of computing robot motion trajectories from arbitrary initial states of motion to reach a target state of motion while satisfying given motion constraints.  \reffig{related-work:reflexxes} by Kr{\"o}ger~\cite{kroger2011opening} illustrates the interface to Reflexxes.  Reflexxes is able to produce time-optimal trajectories that take the robot from a given current state of motion to a desired target state of motion without violating the set of given motion constraints.

Reflexxes has been used~\cite{mahjourian2016neuroevolutionary} to smooth out the end-effector targets continuously produced by a neural network so that robot movements do not violate velocity, acceleration, and jerk limits.  In this work, the motion targets for Reflexxes also act as a temporally-abstract action representation, which helps reduce the dimensionality of the control problem.


\subsubsection{Proximal Policy Optimization (PPO)}

Proximal policy optimization~\cite{schulman2017proximal} is a policy gradient method for model-free reinforcement learning which optimizes a surrogate objective function with respect to the parameters $\theta$ of a policy $\pi_\theta(a | s)$ using stochastic gradient ascent. The objective optimized by PPO is
\\
\begin{equation}
\label{eq:ppo}
L(\theta) = \widehat{\mathbb{E}} \left[ \frac{\pi_\theta(a_t|s_t)}{\pi_{\theta_{old}}(a_t | s_t)} - \beta {\rm KL}( \pi_{\theta_{old}}(\cdot|s_t), \pi_\theta(\cdot|s_t) \right]
\end{equation}
\\
where $\widehat A_t$ is an estimator of the advantage function at time $t$, and $\widehat{\mathbb{E}}$ indicates the empirical average over a batch of samples. The simplest version of PPO optimizes \refeq{ppo} with $\beta = 0$ and clips the policy ratio $\frac{\pi_\theta(a_t|s_t)}{\pi_{\theta_{old}}(a_t | s_t)}$ so as to prevent numerical issues. A different version of PPO adaptively updates $\beta$. 

In comparison to related policy gradient methods, the objective optimized by PPO is similar to that of trust region policy optimization (TRPO). The main difference is TRPO imposes a hard constraint on the KL divergence between successive policies and finds solutions using the conjugate gradient method. Empirically, both TRPO and PPO have been found to lead to more stable policy updates than the standard policy gradient methods. 


\subsubsection{Augmented Random Search (ARS)}

Random search methods for model-free RL directly optimize the policy by searching over its parameters. The simplest version of random search computes a finite-difference approximation of the gradient along a direction chosen uniformly at random on a sphere centered around the current parameters. Let $r_{\theta}$ be a sample return corresponding to rolling out a trajectory with policy parameters $\theta$, and let $\delta_k$ be a random perturbation. Basic random search approximates the gradient as
\begin{equation}
    \widehat{g} = \frac{1}{N} \sum_{k=1}^N (r_{\theta + \delta_k} - r_{\theta - \delta_k}) \delta_k
\end{equation}

The Augmented Random Search algorithm proposed by~\cite{mania2018simple} incorporates several heuristics into the basic search: (1) state vectors are whitened based on an online estimate of state mean and covariance, (2) gradient is estimated based on top $b$ directions with highest return rather than all directions, and (3) the learning rate in each update is scaled by the standard deviation of the obtained rewards.  In addition to being simple to describe and implement, ARS has been shown to have competitive performance with other popular model-free algorithms on MuJoCo~\cite{todorov2012mujoco} benchmark environments. 




\section{Conclusion}
\label{sec:conclusion}

This section enumerates the contributions in this work and concludes the article.


\subsection{Contributions}

The list below enumerates the contributions in this work.

\begin{itemize}
\item Integrating the simulator with a virtual reality environment permits capturing realistic paddle strikes demonstrated by human players.  Such strikes contain versatile paddle motions that are not comparable to simple trajectories constructed from smooth splines.
\item The hierarchical policy permits mixing model-free, model-based, and analytic policies in a single agent.  Decomposing the task into subtasks reduces the number of inputs and outputs for individual subtasks, which makes learning them easier.
\item Decomposing the environment into a game space and a robot space permits learning the game dynamics independently of how robot table tennis is to be played.  The environment decomposition also permits transferring the game dynamics models to new robots without a need for retraining them.
\item Using timed motion targets (including pose and velocity) as high-level actions permits encoding complex robot motions with a simple action representation that is suitable for learning algorithms.
\item The robot-agnostic analytic paddle controller can drive any robot assembly to execute paddle strikes specified by timed motion targets for the paddle.
\item Normalizing and subsampling trajectories recorded from human demonstrations permits training dynamics models with few samples.  Using only about 7000 trajectories, a land-ball policy is trained, which can hit targets with about 20\,cm error on average.
\item The analytic controller and the model-based land-ball policy can drive a robot assembly to play table-tennis rallies lasting 6-8 strikes on average, without any training on the robot.
\item Localizing the application of model-free RL to the strategy skill simplifies the reinforcement learning and exploration problems.  Most RL experiments for learning the strategy policies were carried out on two local workstations and without any hyperparameter tuning.
\item Learning the strategy skill with self-play permits discovering novel cooperative and adversarial game-plays using the same paddle strikes demonstrated by humans.  After only about 24000 ball exchanges in self-play, a game-play strategy is learned that can sustain rallies lasting 14-16 hits on average.
\item Applying self-play to more flexible skills like hit-ball results in discovery of novel strikes beyond what was used by humans.  These strikes are customized for cooperative or adversarial games.
\item Successfully training a functioning table-tennis agent using in the order of tens of thousands human and robot samples demonstrates that the method is sample-efficient enough that it may be deployed in the real world to learn table tennis on physical robots without relying on transfer of models and policies from simulation.
\end{itemize}


\subsection{Conclusion}

The intelligent household robots of the coming decades need to be able to learn by observing humans.  They also need to be able to figure out how to complete a task using their knowledge of how the world and its objects work.  If they have to experiment with a task, they should be able to learn a lot from a few tries.

This article takes a step in that direction by demonstrating the possibility of learning a complex robotic task in a sample-efficient way.  The hierarchical policy design allows for incorporating knowledge of the world by observing humans completing the same task, without restricting the agent to the behavior demonstrated by humans.  Employing simple yet highly-expressive action representations and analytic controllers in the underlying skills gives the higher-level skills the freedom to explore the space of high-level behaviors efficiently, leading to accelerated discovery of novel behaviors that are perceived as \emph{intelligent} by human observers.